%% file: main.tex
\documentclass[runningheads]{assets/style/llncs}
\usepackage[T1]{fontenc}
\usepackage{graphicx}
\usepackage{booktabs}
\usepackage[misc]{ifsym}
\newcommand{\corr}{(\Letter)}
\usepackage{mwe}

\usepackage[utf8]{inputenc} 
\usepackage[T1]{fontenc}    
\usepackage{hyperref}       
\usepackage{url}            
\usepackage{booktabs}       
\usepackage{nicefrac}       
\usepackage{microtype}      

\usepackage{csquotes}
\usepackage{xcolor}
\usepackage{rotating}
\usepackage{assets/my_settings}
\input{assets/math_commands}

\usepackage{wrapfig,lipsum,booktabs}

\begin{document}

\title{Adaptive Sparsity Level during Training for Efficient Time Series Forecasting with Transformers}
\titlerunning{Adaptive Sparsity Level during Training}

\author{Zahra Atashgahi\inst{1} \corr  \and
Mykola Pechenizkiy\inst{2}\and
Raymond Veldhuis\inst{1} \and
Decebal Constantin Mocanu\inst{3, 2}}

\authorrunning{Z. Atashgahi et al.}

\institute{Faculty of Electrical Engineering, Mathematics and Computer Science, University of Twente \email{\{z.atashgahi, r.n.j.veldhuis\}@utwente.nl}
\and
Department of Mathematics and Computer Science, Eindhoven University of Technology \email{m.pechenizkiy@tue.nl}
\and
Department of Computer Science, University of Luxembourg
\email{decebal.mocanu@uni.lu}}
\tocauthor{Zahra Atashgahi, Mykola Pechenizkiy, Raymond Veldhuis, Decebal Constantin Mocanu}
\toctitle{Adaptive Sparsity Level during Training for Efficient Time Series Forecasting with Transformers}
\maketitle

\input{sections/0Abstract}
\input{sections/1Introduction}

\input{sections/2background}

\input{sections/3exploring_sparsity}

\input{sections/4method}

\input{sections/5experiments}

\input{sections/6discussion}

\section{Conclusions}
\looseness=-1
In this paper, we aim to decrease the computational and memory costs of training and deploying DNNs for time series forecasting rather than proposing a new forecasting model and beating the state-of-the-art. Particularly, we focus on transformers while showing the generality of PALS on an MLP-based model (Appendix \ref{app:dlinear}). We first showed that pruning networks for time series forecasting can be challenging in terms of determining the proper sparsity level for various datasets, prediction lengths, and models. Therefore, we proposed PALS, a novel method to obtain sparse neural networks, that exploits loss heuristics to automatically find the best trade-off between loss and sparsity in one round of training. PALS leverages the effective strategies of "Shrink" from during-training pruning and "Stable" from DST. Additionally, we introduce a novel strategy called the "Expand" mechanism. The latter allows PALS to automatically optimize the sparsity level during training, eliminating the need for prior determination. Remarkably, PALS could outperform dense training in $12/14$ cases out of 30 cases ($5$ transformer models, $6$ datasets) in terms of MSE/MAE loss, while reducing $65\%$ parameters count and $63\%$ FLOPs on average. \textbf{Limitations and future work.} Due to the lack of proper hardware to support sparse matrices for on-GPU processing, PALS cannot currently take advantage of its theoretical training and inference speed-up and memory reduction in a real-world implementation. Building a truly sparse transformer demands a substantial investment of both effort and a profound understanding of hardware, an area that is beyond the current scope of our research and human resources (Please refer to Appendix \ref{app:ssec:sparse_implementation} for more details). With the ever-increasing body of work on sparse neural networks, we hope that in the near future, the community paves the way to optimally train sparse neural networks on GPU. An open direction to this research can be to start with a highly sparse neural network (as opposed to starting from a dense network used in PALS) and gradually expand the network to be even more efficient during training. 

\bibliographystyle{splncs04}
\bibliography{res}

\input{Supplementary/7appendix}

\end{document}

%% file: assets/math_commands.tex
\def\eqref#1{equation~\ref{#1}}

\def\1{\bm{1}}

\def\vx{{\bm{x}}}

\def\mA{{\bm{A}}}

\def\mG{{\bm{G}}}

\def\mW{{\bm{W}}}
\def\mX{{\bm{X}}}

\DeclareMathAlphabet{\mathsfit}{\encodingdefault}{\sfdefault}{m}{sl}
\SetMathAlphabet{\mathsfit}{bold}{\encodingdefault}{\sfdefault}{bx}{n}

\newcommand{\Ls}{\mathcal{L}}
\newcommand{\R}{\mathbb{R}}



%% file: sections/0Abstract.tex
\begin{abstract}
\looseness=-1
Efficient time series forecasting has become critical for real-world applications, particularly with deep neural networks (DNNs). Efficiency in DNNs can be achieved through sparse connectivity and reducing the model size. However, finding the sparsity level automatically during training remains challenging due to the heterogeneity in the loss-sparsity tradeoffs across the datasets. In this paper, we propose \enquote{\textbf{P}runing with \textbf{A}daptive \textbf{S}parsity \textbf{L}evel} (\textbf{PALS}), to automatically seek a decent balance between loss and sparsity, all without the need for a predefined sparsity level. PALS draws inspiration from sparse training and during-training methods. It introduces the novel "expand" mechanism in training sparse neural networks, allowing the model to dynamically shrink, expand, or remain stable to find a proper sparsity level. In this paper, we focus on achieving efficiency in transformers known for their excellent time series forecasting performance but high computational cost. Nevertheless, PALS can be applied directly to any DNN. To this aim, we demonstrate its effectiveness also on the DLinear model. Experimental results on six benchmark datasets and five state-of-the-art (SOTA) transformer variants show that PALS substantially reduces model size while maintaining comparable performance to the dense model. More interestingly, PALS even outperforms the dense model, in \textcolor{blue}{12} and \textcolor{blue}{14} cases out of 30 cases in terms of MSE and MAE loss, respectively, while reducing \textcolor{blue}{65\%} parameter count and \textcolor{blue}{63\%} FLOPs on average. Our code and supplementary material are available on Github\footnote{\tiny \url{https://github.com/zahraatashgahi/PALS}}\let\thefootnote\relax\footnotetext{\scriptsize Accepted for publication at ECML PKDD 2024}.

\end{abstract}

%% file: sections/1Introduction.tex
\section{Introduction}

\looseness=-1
The capabilities of transformers \cite{vaswani2017attention} for learning long-range dependencies \cite{wolf2020transformers,dosovitskiy2020image,subakan2021attention} make them an ideal model for time series processing \cite{wen2022transformers}. Several transformer variants have been proposed for the task of time series forecasting, which is crucial for real-world applications, e.g., weather forecasting, energy management, and financial analysis, and have proven to significantly increase the prediction capacity in long time series forecasting (LTSF) \cite{liu2022non}. In addition, attention-based models are inherently an approach for increasing the interpretability for time series analysis in critical applications \cite{lim2021time}. Moreover, recent transformer time series forecasting models (e.g., \cite{wu2021autoformer,zhou2022fedformer,liu2022non}) perform generally well in other time series analysis tasks, including, classification, anomaly detection, and imputation \cite{wu2022timesnet}.

\looseness=-1
Despite the outstanding performance of transformers, these models are computationally expensive due to their large model sizes as shown in \cite{strubell2020energy} for natural language processing. 
With the ever-increasing collection of large time series and the need to forecast millions of them, the requirement to develop computationally efficient forecasting models is becoming significantly critical \cite{talagala2018meta,hyndman2016fast,rakthanmanon2012searching}. For industry-scale time series data, which are often high-dimensional and long-length, deploying transformers requires automatically discovering memory- and computationally-efficient architectures that are scalable and practical for real-world applications \cite{wen2022transformers}. While there have been some efforts to reduce the computational complexity of transformers in time series forecasting \cite{zhou2022fedformer,zhou2021informer}, these models have in order of millions of parameters, that can be too large for resource-limited applications, e.g., mobile phones. The over-parameterization of these networks causes high training and inference costs, and their deployment in low-resource environments (e.g., lack of GPUs) would be infeasible. To address these issues, we raise the research question: \textit{How can we reduce the computational and memory overheads of training and deploying transformers for time series forecasting without compromising the model performance?}

\begin{figure*}[!t]
\centering
\vspace{-3mm}
\includegraphics[width=\textwidth]{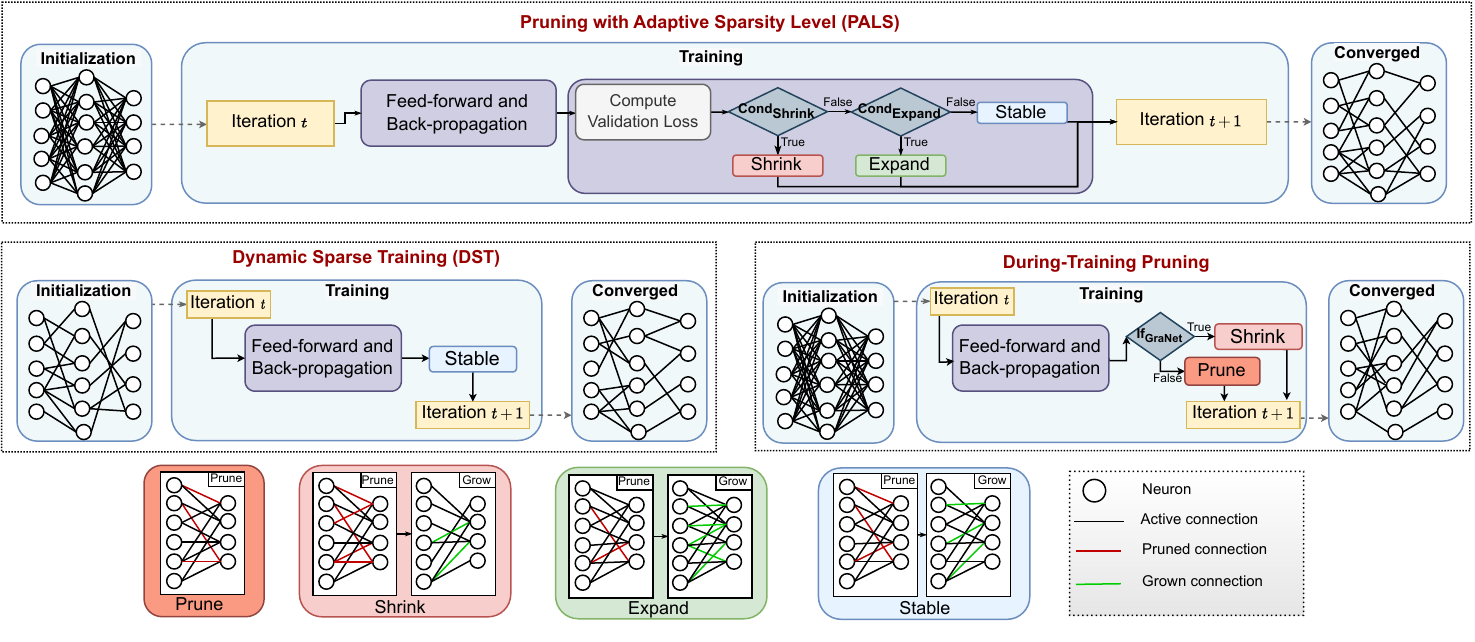}
\vspace{-3mm}
\caption{\looseness=-1 Schematic overview of the proposed method, \textbf{PALS} (Algorithm \ref{alg:PALS}), Dynamic Sparse Training (DST) \cite{mocanu2018scalable,evci2020rigging}, During-training pruning (Gradual Magnitude Pruning (GMP) \cite{zhu2017prune}, and GraNet \cite{liu2021sparse}). While DST and during-training pruning use a fixed sparsity schedule to achieve a pre-determined sparsity level at the end of the training, PALS updates the sparse connectivity of the network at each $\Delta t$ iterations during training, by deciding whether to "Shrink" (decrease density) or "Expand" (increase density) the network or remain "Stable" (same density), to automatically find a proper sparsity level.} \label{fig:dstgra}
\vspace{-5mm}
\end{figure*}

\looseness=-1
Seeking sparsity through sparse connectivity is a widely-used technique to address the over-parameterization of deep learning models \cite{hoefler2021sparsity}. Early approaches for deriving a sparse sub-network prune a trained dense model \cite{han2015learning}, known as \textit{post-training} pruning. While these methods can match the performance of the dense network as shown by the Lottery Ticket Hypothesis (LTH) \cite{frankle2018lottery}, they are computationally expensive during training due to the training of the dense network. \textit{During-training} pruning aims to maintain training efficiency by gradually pruning a dense network during training \cite{liu2021sparse}. Sparse training \cite{mocanu2018scalable} pushed the limits further by starting with a sparse network from scratch and optimizing the topology during training. However, as we study in Section \ref{sec:sparsity_effect}, the main challenge when using any of these techniques for time series forecasting is to find the proper sparsity level automatically.

\looseness=-1
In this paper, we aim to move beyond optimizing a single objective (e.g. minimizing loss) and investigate sparsity in DNNs for time series prediction in order to find a good trade-off between computational efficiency and performance automatically. Our contributions are:
\textbf{(1)} We analyze the effect of sparsity (using unstructured pruning) in SOTA transformers for time series prediction \cite{liu2022non,zhou2022fedformer,wu2021autoformer,zhou2021informer}, and vanilla transformer \cite{vaswani2017attention}. We show they can be pruned up to $80\%$ of their connections in most cases, without significant loss in performance.
\textbf{(2)} We propose an algorithm, called \enquote{\textbf{P}runing with \textbf{A}daptive \textbf{S}parsity \textbf{L}evel} (\textbf{PALS}) that finds a decent loss-sparsity trade-off by dynamically tuning the sparsity level during training using the loss heuristics and deciding at each connectivity update step weather to \emph{Shrink} or \emph{Expand} the network, or keep it \emph{Stable}. PALS (Figure \ref{fig:dstgra}) creates a bridge between during-training pruning and dynamic sparse training research areas by inheriting and enhancing some of their most successful mechanisms, while - up to our best knowledge - introducing for the first time into play also the Expand mechanism. Consequently, PALS does not require a desired pre-defined sparsity level which is necessary for most pruning or sparse training algorithms. 
\textbf{(3)} We evaluate the performance of PALS in terms of the loss, the parameter count, and FLOPs on six widely-used benchmarks for time series prediction and show that PALS can substantially sparsify the models and reduce parameter count and FLOPs. Surprisingly, PALS can even outperform the dense model on average, in \textcolor{blue}{12} and \textcolor{blue}{14} cases out of 30 cases in terms of Mean Squared Error (MSE) and Mean Absolute Error (MAE) loss, respectively (Table \ref{tab:summary_results}).

%% file: sections/2background.tex
\section{Background}

\subsection{Sparse Neural Networks}
\looseness=-1
Sparse neural networks (SNNs) use sparse connectivity among layers to reduce the computational complexity of DNNs while maintaining a close performance to the dense counterpart in terms of prediction accuracy. SNNs can be achieved using dense-to-sparse or sparse-to-sparse approaches \cite{mocanu2021sparse}. 

\looseness=-1
\paragraph{Dense-to-sparse} methods prune a dense network; based on the pruning phase, they are categorized into three classes: \textit{post-training} \cite{han2015learning,frankle2018lottery}, \textit{Before-training} \cite{lee2018snip}, and \textit{during-training} \cite{zhu2017prune,louizos2018learning,liu2021sparse} pruning. Post-training pruning suffers from high computational costs during training and before-training approaches usually fall behind the performance of the dense counter-part network. In contrast, during-training approaches, maintain close or even better performance to the dense network while being efficient through the training process. A standard during-training pruning is  Gradual Magnitude Pruning (GMP) \cite{zhu2017prune} which gradually drops unimportant weights based on the magnitude during the training process. GraNet \cite{liu2021sparse} is another during-training algorithm that gradually shrinks (decreasing density) a network to reach a pre-determined sparsity level. It prunes the weights (as performed in GMP) while allowing for connection regeneration (as seen in Dynamic Sparse Training (DST) which will be explained in the following). As the number of grown weights is less than the pruned ones, the network is shrunk and the density is decreased. For details regarding GraNet, refer to Appendix \ref{app:sparsity_effect_granet}.

\looseness=-1
\paragraph{Sparse-to-sparse} methods start with a random sparse network from scratch and the number of parameters is usually fixed during training and can be determined based on the available computational budget. The sparse topology can remain fixed (static) \cite{mocanu2016topological} or dynamically optimized during training (a.k.a Dynamic Sparse Training (DST)) \cite{mocanu2018scalable,evci2020rigging,jayakumar2020top,liu2021we,yuan2021mest,liu2021topological}. At each topology update iteration, a fraction of unimportant weights are dropped (usually based on magnitude), and the same number of weights are grown. The growth criteria can be random, as in Sparse Evolutionary Training (SET) \cite{mocanu2018scalable}, or gradient, as in Rigged Lottery (RigL) \cite{evci2020rigging}. 

\begin{wraptable}{r}{7cm}
\begin{scriptsize}
    \vspace{-6mm}
    \caption{Comparison of related work.}\label{tab:comparison}
    \vspace{-5mm}
    \begin{center}
    \scalebox{0.7}{ 
    \begin{tabular}{@{\hskip 0.04in}c@{\hskip 0.04in}|@{\hskip 0.04in}c@{\hskip 0.04in}c@{\hskip 0.04in}c@{\hskip 0.04in}c@{\hskip 0.04in}c@{\hskip 0.04in}c@{\hskip 0.04in}}\toprule 
    \bt Method         & \bt Shrink &\bt Stable &\bt Expand & \bt \makecell{Adaptive\\Sparsity Schedule} & \bt \makecell{Automatic tune\\of sparsity level}\\\midrule
    \bt RigL & \textcolor{gray}{\xmark}   & \cmark    & \textcolor{gray}{\xmark}  & \textcolor{gray}{\xmark}   &\textcolor{gray}{\xmark}  \\\hline
    \bt GMP &\cmark & \textcolor{gray}{\xmark}  & \textcolor{gray}{\xmark}  &\textcolor{gray}{\xmark}  &\textcolor{gray}{\xmark}  \\\hline
    \bt GraNet   &\cmark     & \cmark & \textcolor{gray}{\xmark}  &\textcolor{gray}{\xmark}  &\textcolor{gray}{\xmark}  \\\hline
    \bt PALS (ours)    &\cmark    &\cmark  &\cmark & \cmark  &\cmark\\\bottomrule       
    \end{tabular}}
    \end{center}
\end{scriptsize}
\vspace{-8mm}
\end{wraptable}

\looseness=-1
In this work, we take advantage of the successful mechanism of \enquote{\textit{Shrink}} from during-training pruning (e.g., GraNet \cite{liu2021sparse}) and \enquote{\textit{Stable}} from DST (e.g., RigL \cite{evci2020rigging}) and propose for the first time the \enquote{\textit{Expand}} mechanism, to design a method to automatically optimize the sparsity level during training without requiring to determine it beforehand. Each of these mechanisms is explained in Section \ref{sec:proposed_method}. In Table \ref{tab:comparison}, we present a summarized comparison with the closest related work in the literature. Figure \ref{fig:dstgra} presents a comprehensive embedding of our proposed method in the literature. Unlike these methods, which update the network using fixed schedules to reach a pre-determined sparsity level, PALS proposes an adaptive approach. It automatically determines whether to shrink or expand the network or remain stable, in order to tune the sparsity level and find a good trade-off between loss and sparsity.

\looseness=-1
Only a few works investigated SNNs for time series analysis \cite{9814603}. \cite{xiao2022dynamic} investigates sparsity in convolutional neural networks (CNNs) for the time series classification and shows their proposed method has superior prediction accuracy while reducing computational costs. \cite{kieu2019outlier} exploit sparse recurrent neural networks (RNNs) for outlier detection. \cite{liu2021selfish} and \cite{pmlr-v151-furuya22a} explore sparsity in RNNs for sequence learning.

\paragraph{Sparsity in Transformers.} 
\looseness=-1
Several works have sought sparsity in transformers \cite{ganesh2021compressing,prasanna2020bert}. These approaches can be categorized into structured (blocked) \cite{michel2019sixteen} or unstructured (fine-grained) pruning \cite{NEURIPS2020_b6af2c97}. As discussed in \cite{hoefler2021sparsity}, structured sparsity for transformers is able to only discover models with very low sparsity levels; therefore, we focus on unstructured pruning. \cite{pmlr-v119-li20m} analyses pruning transformers for language modeling tasks and shows that large transformers are robust to compression. \cite{chen2021chasing} dynamically extract and train sparse sub-networks from Vision Transformers (ViT) \cite{dosovitskiy2020image} while maintaining a fixed small parameter budget, and they could even improve the accuracy of the ViT in some cases. \cite{dietrich2022towards} investigates DST for BERT language modeling tasks and shows Pareto improvement over the dense model in terms of FLOPs. However, these works mostly focus on vision and NLP tasks. To the best of our knowledge, no work has investigated sparse connectivity in transformers for time series analysis that faces domain-specific challenges as we will elaborate in Section \ref{sec:sparsity_effect}. Please note that there is a line of research focusing on \textit{sparse attention} \cite{tay2022efficient} aiming to develop an efficient self-attention mechanism that is orthogonal to our focus in this work (sparsity and pruning) \cite{hoefler2021sparsity}.

\subsection{Time Series Forecasting}\label{ssec:related_forecasting}
\looseness=-1
Initial studies for time series forecasting exploit classical tools such as ARIMA \cite{box2015time}. While traditional methods mostly rely on domain expertise or assume temporal dependencies follow specific patterns, machine learning techniques learn the temporal dependencies in a data-driven manner \cite{lim2021time,wang2022latent,li2022generative}. In recent years, various deep learning models, including RNNs \cite{qin2017dual,salinas2020deepar}, multi-layer perceptrons (MLP) \cite{zeng2022transformers,zhang2022less}, CNNs \cite{lai2018modeling}, and Temporal convolution networks \cite{franceschi2019unsupervised} are utilized to perform time series forecasting \cite{oreshkin2019n,challu2022n,jin2022domain}.

\looseness=-1
Transformers have been extensively used to perform time series forecasting due to their strong ability for sequence modeling. A class of models aims at improving the self-attention mechanism and addresses the computational complexity of vanilla transformers such as LogTrans \cite{NEURIPS2019_6775a063}, Informer \cite{zhou2021informer}, Reformer \cite{Kitaev2020Reformer}. Another category of methods seeks to modify the model to capture the inherent properties of the time series:  Autoformer \cite{wu2021autoformer} introduces a seasonal trend decomposition with an auto-correlation block as the attention module. NSTransformer \cite{liu2022non} proposes to add two modules including series stationarization and de-stationary attention in the transformer architecture. FEDformer \cite{zhou2022fedformer} proposes to combine transformers with a seasonal-trend decomposition method to capture global and detailed behaviour of the time series. The research into designing transformers for time series forecasting is ongoing, and many other transformer variants have been proposed, such as Crossformer \cite{zhang2023crossformer}, ETSformer \cite{woo2022etsformer}, Pyraformer \cite{liu2021pyraformer}. 

\subsection{Problem Formulation and Notations} \label{ssec:roblem_formulation}
\looseness=-1
Let $\vx_t \in \R^m$ denote the observation of a multivariate time series $\mX$ with $m$ variables at time step $t$. Given a look-back window $\mX_{t-L:t}=[\vx_{t-L},...,\vx_{t-1}]$ of size $L$, time series forecasting task aims to predict time series over a horizon $H$ as $\widetilde{\mX}_{t:t+H}=[\widetilde{\vx}_{t},...,\widetilde{\vx}_{t+H-1}]$ where $\widetilde{\vx}_{t}$ is the prediction at time step $t$. To achieve this, we need to train a function $f(\mX_{t-L:t}, \theta)$ (e.g. a transformer network) that can predict future values over horizon $H$. 

\looseness=-1
In this paper, we aim to reduce the model size by pruning the unimportant parameters from $\theta$ such that we find the sparse model $f(\mX_{t-L:t}, \theta_s)$ where $\norm{\theta_s}_0 	\ll \norm{\theta}_0$. $D = \frac{\norm{\theta_s}_0}{\norm{\theta}_0}$ is called the density level of the model $f$ and $S=1-D$ is called as the sparsity level. The aim is to minimize the reconstruction loss between the prediction $\Ls(f(\mX_{t-L:t}, \theta_s), \mX_{t:t+H})$ while finding a proper sparsity level $S$ automatically. We use Mean Squared Error (MSE) as the loss function such that:
\begin{equation}
\Ls(\widetilde{\mX}_{t:t+H}, \mX_{t:t+H}) = \frac{1}{H}\Sigma_{i=0}^{H-1} (\widetilde{\vx}_{t+i} - \vx_{t+i})^2.
\vspace{-4mm}
\end{equation}


%% file: sections/3exploring_sparsity.tex
\section{Analyzing Sparsity Effect in Transformers for Time Series Forecasting}\label{sec:sparsity_effect}
\looseness=-1
In this section, we explore sparsity in several time series forecasting transformers. In short, we apply GraNet \cite{liu2021sparse} to prune each model and measure their performance over various sparsity levels.

\begin{figure}[!b]
    \centering
    \includegraphics[width=\textwidth]{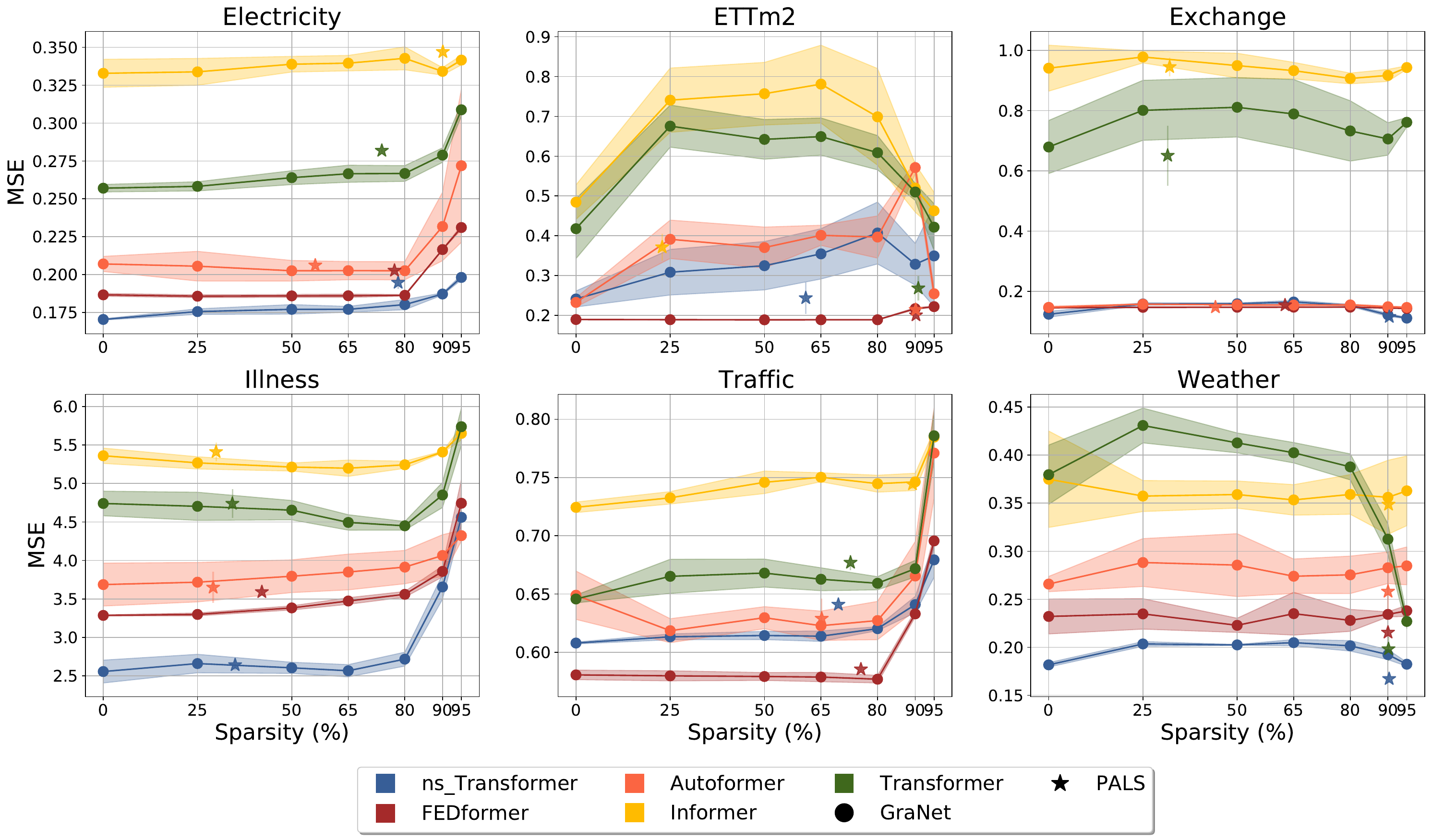}
    \caption{Sparsity effect on the performance of various transformer models for time series forecasting on benchmark datasets in terms of MSE loss (prediction length $=96$, except $24$ for the Illness dataset). Each model is sparsified using GraNet \cite{liu2021sparse} to sparsity levels ($\%$) $\in \{25, 50, 65, 80, 90, 95\}$ and PALS. $Sparsity = 0$ indicates the original dense model.}
    \label{fig:sparsity_effect_granet}
\end{figure}

\paragraph{Experimental Settings.}
\looseness=-1
We perform this experiment on six benchmark datasets, presented in Table \ref{tab:datasets}. We adapt GraNet \cite{liu2021sparse}, a during-training pruning algorithm developed for CNNs, to sparsify transformer models for time series forecasting. GraNet gradually shrinks a network  (here, we start from a dense network) during the training to reach a pre-determined sparsity level, while allowing for connection regeneration inspired by DST. GraNet is described in Appendix \ref{app:sparsity_effect_granet}. For more details regarding the experimental settings, please refer to Section \ref{sec:settings}. For each sparsity level ($\%$) in $\{25, 50, 65, 80, 90, 95\}$, we measure the prediction performance of each transformer model in terms of MSE loss. The results for prediction length $=96$ (except $24$ for the Illness dataset) are presented in Figure \ref{fig:sparsity_effect_granet}. The results for other prediction lengths are presented in Figure \ref{fig:granet_sparsity_effect_all} in Appendix \ref{app:sparsity_effect_granet}.

\paragraph{Sparsity Effect.}
\looseness=-1
We present the results for pruning various transformers in Figure \ref{fig:sparsity_effect_granet}; most models can be pruned up to $80\%$ or higher sparsity levels without significantly affecting performance. Moreover, a counter-intuitive observation is that in some cases, sparsity does not necessarily lead to worse performance than the dense counterpart, and it can even improve the performance. For example, while on the Electricity, Illness, and Traffic datasets, the behavior is as usually expected (higher sparsity leads to lower performance), on the three other datasets, higher sparsity might even lead to better performance (lower loss) than the dense model. In addition, the sparsity effect is different among various models, particularly on the latter group of datasets, including the ETTm2, Exchange, and Weather datasets. We discuss the potential reasons for different behavior among datasets in Appendix \ref{ssec:model_size_effect}. Last but not least, by looking at Figure \ref{fig:granet_sparsity_effect_all} in Appendix \ref{app:sparsity_effect_granet}, the prediction length can also be a contributing factor to the sparsity-loss trade-off.


\paragraph{Challenge.}
\looseness=-1
Based on the above observations, we can conclude that the sparsity effect is not homogeneous across various time series datasets, forecasting models, and prediction lengths for time series forecasting. Our findings in these experiments are not aligned with the statements in \cite{hoefler2021sparsity} for CNNs (vision) and Transformers (NLP), where for a given task and technique, increasing the sparsity level results in decreasing the prediction performance. However, we observe in Figures \ref{fig:sparsity_effect_granet} and \ref{fig:granet_sparsity_effect_all} that increasing the sparsity level does not necessarily lead to decreased performance and it might even significantly improve the performance (e.g. for the vanilla transformer on the Weather dataset). Therefore, it is challenging to decide how much we can push the sparsity level and what is the decent sparsity level without having prior knowledge of the data, model, and experimental settings. While GraNet is the closest in spirit to our proposed method, it cannot automatically tune the sparsity level since it needs the initial and the final sparsity level as its hyperparameters. In this paper, we aim to address this challenge by proposing an algorithm that can automatically tune the sparsity level during training.

\begin{figure}[!t]
\vspace{-3mm}
\centering
\scalebox{1}{
 
    \begin{minipage}{\linewidth}
        \input{assets/PALS_algorithm.tex}
    \end{minipage}%
}
\vspace{-5mm}
\end{figure}

%% file: assets/PALS_algorithm.tex
\centering
\setlength{\textfloatsep}{2pt}
\begin{algorithm}[H]
    \caption{PALS}\label{alg:PALS}
    \algsetup{linenosize=\small }
    \footnotesize
    \begin{algorithmic}[1]
       
       \STATE \textbf{Input}: Time series $\mX \in\R^{T\times m}$, number of training iterations $t_{max}$, Sequence length $L$, Prediction length $H$, model dimension $d_{model}$, pruning rate $\zeta$, mask update frequency $\Delta t$, Initial density $D_{init}$, pruning rate factor $\gamma>1$ and loss freedom factor $\lambda>1$, sparsity bound $S_{min}$ and $S_{max}$.
       \STATE\textbf{Initialization}: Initialize the transformer model with density level $D_{init}$, $S = 1 - D_{init}$, $L_{best} = \inf$.
        \STATE\textbf{Training}:
        \FOR{\texttt{$t \in \{1,\dots, \#t_{max}\}$}}
        \STATE \textbf{I. Standard feed-forward and back-propagation.} The network is trained on $batch_t$ of samples.
        \STATE \textbf{II. Update sparsity mask}
            \begin{ALC@g}
            \IF{$(t \mod \Delta t)=0$} 
                \STATE Compute Validation Loss $L_{valid}^{t}$
                \IF{($S < S_{min}$) \textbf{or} ($ L_{valid}^{t} <= \lambda * L_{best}$ \textbf{and} $S<S_{max}$)} 
                    \STATE  \textit{update\_mask (}$ \zeta_{prune}=\gamma*\zeta, \: \zeta_{grow}=\zeta$\textit{)}
                \ELSIF{$ L_{valid}^{t} > \lambda * L_{best}$ \textbf{and} $S>S_{best}$}
                    \STATE \textit{update\_mask (}$ \zeta_{prune}=\zeta, \: \zeta_{grow}=\gamma*\zeta$\textit{)}
                \ELSE \STATE \textit{update\_mask (}$ \zeta_{prune}=\zeta, \: \zeta_{grow}=\zeta$\textit{)}
                \ENDIF
                \IF{$L_{valid}^{t} < L_{best} $} 
                    \STATE  $L_{best} = L_{valid}^{t}, \: S_{best} = S$
                \ENDIF
                \STATE Set $S$ to the sparsity level of the network.
            \ENDIF
            \end{ALC@g}
                
                    
        \ENDFOR
            
    \end{algorithmic}

\end{algorithm}

%% file: sections/4method.tex

\section{Proposed Methodology: PALS}\label{sec:proposed_method}
\looseness=-1
This section presents our proposed method for automatically finding a proper sparsity level of a DNN, called \enquote{\textbf{P}runing with \textbf{A}daptive \textbf{S}parsity \textbf{L}evel} (\textbf{PALS}) (Algorithm \ref{alg:PALS}). While our main focus in this paper is to sparsify transformer models, PALS is not specifically designed for transformers and can be applied directly to other artificial neural network architectures (See Appendix \ref{app:dlinear} for experiments on training with PALS the DLinear \cite{zeng2022transformers}) model.

\looseness=-1
\paragraph{Motivation and Broad Outline.} As we discussed in Section \ref{sec:sparsity_effect}, the main challenge when seeking sparsity for time series forecasting is to find a good sparsity level automatically. Therefore, PALS aims to tune the sparsity level during training without requiring prior information about models or datasets. PALS is in essence inspired by the DST framework \cite{mocanu2018scalable} and gradual magnitude pruning (GMP) \cite{zhu2017prune,liu2021sparse}. While DST and GMP use fixed sparsification policies (fixed sparsity level (\emph{Stable} in Figure \ref{fig:dstgra}) and constantly prune the network until the desired sparsity level is reached (\emph{Shrink} in Figure \ref{fig:dstgra}), respectively) and require the final sparsity level before training, PALS exploits heuristic information from the network at each iteration to automatically determine whether to \emph{increase}, \emph{decrease}, or \emph{keep} the sparsity level at each connectivity update step. While existing growing methods \cite{han2017dsd,ma2022effective} grow a network or a layer of it to dense connectivity, to the best of our knowledge, this is the first work that allows the network to \emph{expand} by increasing the density during training without requiring dense connectivity, and allows for automatic shrink or expand. If the training starts from a dense neural network ($D_{init}=1$) PALS can be seen as a dense-to-sparse method, while if $D_{init}<1$ then PALS is a sparse-to-sparse method.

\textbf{Training.}
\looseness=-1
 The training of PALS (Algorithm \ref{alg:PALS}) starts with initializing a network with density level $D_{init} = 1 - S_{init}$. Then, the training procedure of PALS consists of two steps:

\textbf{1. Standard feed-forward and back-propagation.} Network's parameters are updated each training iteration $t$ using a batch of samples.

\looseness=-1
\textbf{2. Update Sparse Connectivity.} The novelty of the method lies in updating the sparse connectivity. At every $\Delta t$ iteration, the connectivity is updated in two steps. (2-1) The validation loss at step $t$ is calculated as $L_{valid}^{t}$. (2-2) The sparsity mask is updated ($update\_mask$ in Algorithm \ref{alg:PALS}) by first pruning $\zeta_{prune}$ of weights with the lowest magnitude:
\begin{equation}
    \widetilde{\mW_l} = Update(\mW_l, top(|\mW_l|, 1-\zeta_{prune})),
\end{equation}
where $\mW_l$ is the $l^{\text{th}}$ weight matrix of the network, $Update(\mA,idx)$ keeps only the indices $idx$ of the matrix $\mA$, $top(\mA, \zeta)$ returns the indices of a fraction $\zeta$ of the largest elements of $\mA$. Then, we grow $\zeta_{grow}$ of the weights with the highest gradients:
\begin{equation}
    \mW_l = \widetilde{\mW_l} + top(|\mG_{l, i\notin \widetilde{\mW_l}}|, \zeta_{grow})
\end{equation}
where $\mG_{l, i\notin \widetilde{\mW_l}}$ is the gradient of zero weights in layer $l$. These new connections are initialized with zero values. This process is repeated for each layer in the model. Based on the values of $\zeta_{prune}$ and $\zeta_{grow}$, PALS determines whether to decrease (\emph{shrink}), increase (\emph{expand}), or keep (\emph{stable}) the network:

\quad$\mathbf{S_t > S_{t-1}}$ \textbf{(Shrink).} If the loss does not go beyond $\lambda * L_{best}$, we decrease the overall number of parameters such that $\zeta_{prune}=\gamma*\zeta, \: \zeta_{grow}=\zeta$. The loss freedom coefficient, $\lambda >1$, is a hyperparameter of the network that determines how much the loss value can deviate from the best validation loss achieved so far $L_{best}$ during training. The lower $\lambda$ is, the more strict PALS becomes at allowing the network to go to the \textit{shrink} phase, finally resulting in a lower sparsity network. $\gamma > 1$ is the pruning factor coefficient, which determines how much to prune or grow more in the shrink and expand phases, respectively. We analyze the sensitivity of PALS to $\lambda$ and $\gamma$ in Section \ref{ssec:hyperparameter_sensitivity}. In addition, we define a boundary for sparsity determined by $S_{min}$ and $S_{max}$ which can be determined by the user based on the available resources. If the sparsity level does not meet the minimum sparsity level $S_{min}$, we prune the network more than we grow. If the network sparsity goes beyond $S_{max}$, we do not increase sparsity. 

\quad$\mathbf{S_t < S_{t-1}}$ \textbf{(Expand).}  If $S > S_{best}$ ($S_{best}$ is the sparsity level corresponding to $L_{best}$) and the loss goes higher than $\lambda * L_{best}$, it means that the earlier pruning step(s) were not beneficial to decreasing the loss (improving forecasting quality in the time series forecasting) and the network requires a higher capacity to recover a good performance. Therefore, we expand the network and grow more connections than the pruned ones at this step: $\zeta_{prune}=\zeta, \: \zeta_{grow}=\gamma*\zeta$.
        
\quad$\mathbf{S_t = S_{t-1}}$ \textbf{(Stable).} If none of the above cases happened, we only update a fraction $\zeta$ of the network's parameters without changing the sparsity level:  $\zeta_{prune}=\zeta, \: \zeta_{grow}=\zeta$.

For a better understanding of how the sparsity level evolves during the training process of PALS, please refer to Appendix \ref{app-ssec:sparsity_during_training}.

%% file: sections/5experiments.tex

\section{Experiments and Results }

\subsection{Experimental Settings}\label{sec:settings}
\paragraph{Datasets.}
The experiments are performed on six widely-used benchmark datasets for time series forecasting. The datasets are summarized in Table \ref{tab:datasets} and described in Appendix \ref{app:settings}. These datasets have different characteristics including stationary and non-stationary with/without obvious periodicity. Each database in each experiment is divided into three sets: train, validation, and test set. The data from the test set is only used for the final evaluation of all methods. The validation data is used to choose the best model during training and early stopping for all models including dense and sparse. Therefore, all methods use the loss signal from validation data to tune their model and select the model with the lowest validation loss, and they have all seen an equal amount of data during training

\vspace{-2mm}
\paragraph{Models.}
\looseness=-1
We consider five SOTA transformer models for time series forecasting, including Non-Stationary Transformer (NSTransformer) \cite{liu2022non}, FEDformer \cite{zhou2022fedformer}, Autoformer \cite{wu2021autoformer}, Informer \cite{zhou2021informer}, and vanilla transformer \cite{vaswani2017attention}. Please refer to Section \ref{ssec:related_forecasting} for more details. 

\vspace{-2mm}
\paragraph{Evaluation metrics.}
\looseness=-1
We evaluate the methods in two aspects: 1) Quality of the prediction in terms of MSE and MAE, and 2) Computational complexity in terms of parameter count and FLOPs (Floating-point operations). We report the theoretical FLOPs to be independent of the used hardware, as it is done in the unstructured pruning literature \cite{liu2021sparse,evci2020rigging}. A lower value for these metrics indicates higher prediction quality and lower computational complexity, respectively. We measure the performance of each model for various prediction lengths $H\in\{96, 192, 336, 720\}$ (except $H\in\{24, 36, 48, 60\}$ for the Illness dataset).

\vspace{-2mm}
\paragraph{Implementation.}
\looseness=-1
Experiments are implemented in PyTorch. The start of implementation is the NSTransformer \footnote{\tiny \url{https://github.com/thuml/Nonstationary_Transformers}} and GraNet\footnote{\tiny\url{https://github.com/VITA-Group/GraNet}}. We repeat each experiment for three random seeds and report the average of the runs. In the experiments, $D_{init}$ was set to $1$, thus PALS can be seen as a during-training pruning method. We have run the experiments on \textit{Intel Xeon Platinum 8360Y CPU} and one \textit{NVIDIA A100 GPU}. We will discuss the hyperparameters' settings in Appendix \ref{app:settings}.

\subsection{Results}\label{sec:results}

\input{assets/tables/summary_comparison_models}
\textbf{Multivariate Time Series Prediction.}
The results in terms of MSE and parameter count for the considered datasets and models are presented in Table \ref{tab:results_all} in Appendix \ref{app:comp_results}. In most cases considered, PALS decreases the model size by more than $50\%$ without a significant increase in loss. More interestingly, in most cases on the ETTm2, Exchange, and Weather datasets PALS even achieves lower MSE than the dense counterpart. 

\looseness=-1

\looseness=-1
To summarize the results of Table \ref{tab:results_all} (Appendix \ref{app:comp_results}) and have a general overview of the performance of PALS on each model and dataset, we present the average MSE and MAE, and parameters count in addition to the difference between the dense and the sparse model using PALS (in percentage) in Table \ref{tab:summary_results}. Additionally, we include the inference FLOPs count (total FLOPs for all test samples). PALS even outperforms the dense model in \textcolor{blue}{12} and \textcolor{blue}{14} cases out of 30 cases in terms of MSE and MAE loss, respectively, while reducing \textcolor{blue}{65\%} parameter count and \textcolor{blue}{63\%} FLOPs on average. We summarize the training FLOPs in Appendix \ref{app:train_flops}.

\looseness=-1
Based on the experiments conducted in Section \ref{sec:sparsity_effect} and the description of datasets provided in Appendix \ref{app:datasets},  we observed significant variations in the sparsity-loss trade-off across different datasets and models. The beauty of our proposed method consists in the fact that it does not have to consider any of these differences. We did not make any finetuning for PALS to account for these differences, and it does everything automatically. Of course, finetuning PALS per dataset and model specificity would improve its final performance, but it would reduce the generality of our proposed work and we prefer not to do it.

\textbf{Univariate Time Series Prediction.}
The results of univariate prediction (using a single variable) on the ETTm2 and Exchange datasets are presented in Table \ref{tab:results_all_univariate} and summarized in Table \ref{tab:summary_results|_univariate} in Appendix \ref{app:univariate}. In short, PALS outperforms the dense counterpart model on average, in \textcolor{blue}{7} and \textcolor{blue}{8} cases out of 12 cases in terms of MSE and MAE loss, respectively.

%% file: assets/tables/summary_comparison_models.tex
\begin{sidewaystable}

\caption{Summary of the results on the benchmark Datasets in Table \ref{tab:results_all}. For each experiment on a transformer model and dataset, the average MSE, MAE, number of parameters ($\times 10^{6}$) and the inference FLOPs count ($\times 10^{12}$) for various prediction lengths are reported before and after applying PALS. The difference between these results is shown in \% where the \textcolor{blue}{blue} color means improvement of PALS compared to the corresponding dense model in terms of MSE or MAE.}\label{tab:summary_results}
\centering
\scalebox{0.55}{
\begin{tabular}{c| c@{\hskip 0.02in}c@{\hskip 0.02in}c@{\hskip 0.02in}c|c@{\hskip 0.02in}c@{\hskip 0.02in}c@{\hskip 0.02in}c|c@{\hskip 0.02in}c@{\hskip 0.02in}c@{\hskip 0.02in}c|c@{\hskip 0.02in}c@{\hskip 0.02in}c@{\hskip 0.02in}c|c@{\hskip 0.02in}c@{\hskip 0.02in}c@{\hskip 0.02in}c|c@{\hskip 0.02in}c@{\hskip 0.02in}c@{\hskip 0.02in}c@{\hskip 0.02in}c}
\toprule
\bt Model  &\multicolumn{4}{c}{Electricity} & \multicolumn{4}{c}{ETTm2}  & \multicolumn{4}{c}{Exchange} & \multicolumn{4}{c}{Illness}  & \multicolumn{4}{c}{Traffic}  & \multicolumn{4}{c}{Weather}\\

& MSE &MAE &\#Params &\#FLOPs & MSE &MAE &\#Params &\#FLOPs& MSE &MAE &\#Params &\#FLOPs& MSE &MAE  &\#Params &\#FLOPs & MSE &MAE &\#Params &\#FLOPs & MSE &MAE  &\#Params &\#FLOPs \\
\midrule

NSTransformer& \textbf{0.19}& \textbf{0.30}& 12.0& 9.25& 0.49& 0.43& 10.6& 19.82& 0.54& 0.49& 10.6& 1.89& \textbf{2.14}& \textbf{0.92}& 10.5& 0.05& 0.63& \textbf{0.34}& 14.2& 6.61& 0.29& 0.31& 10.7& 18.10 \\
+PALS& 0.21& 0.32& 2.2& 1.81& 0.38& 0.39& 2.5& 3.70& \textbf{0.49}& \textbf{0.47}& \textbf{5.4}& \textbf{1.07}& 2.33& 0.97& 7.4& 0.04& 0.67& 0.37& 4.3& 2.05& \textbf{0.26}& \textbf{0.29}& 1.0& 1.77 \\
Difference& 10.8\% $\uparrow$& 7.3\% $\uparrow$& \textcolor{black}{81.5\% $\downarrow$}& \textcolor{black}{80.5\% $\downarrow$}& \textcolor{blue}{24.0\% $\downarrow$}& \textcolor{blue}{11.2\% $\downarrow$}& \textcolor{black}{76.7\% $\downarrow$}& \textcolor{black}{81.3\% $\downarrow$}& \textcolor{blue}{9.3\% $\downarrow$}& \textcolor{blue}{3.6\% $\downarrow$}& \textcolor{black}{48.5\% $\downarrow$}& \textcolor{black}{43.3\% $\downarrow$}& 9.1\% $\uparrow$& 5.1\% $\uparrow$& \textcolor{black}{30.0\% $\downarrow$}& \textcolor{black}{30.2\% $\downarrow$}& 5.2\% $\uparrow$& 9.1\% $\uparrow$& \textcolor{black}{70.1\% $\downarrow$}& \textcolor{black}{69.0\% $\downarrow$}& \textcolor{blue}{10.2\% $\downarrow$}& \textcolor{blue}{6.9\% $\downarrow$}& \textcolor{black}{90.3\% $\downarrow$}& \textcolor{black}{90.2\% $\downarrow$} \\
\midrule

FEDformer& 0.21& 0.32& 19.5& 9.30& \textbf{0.30}& 0.35& 17.9& 19.82& 0.50& 0.49& 17.9& 1.89& 2.84& 1.14& 13.7& 0.05& \textbf{0.61}& 0.38& 22.3& 6.71& 0.32& 0.37& 17.9& 18.11 \\
+PALS& 0.23& 0.34& 3.0& 1.35& 0.30& \textbf{0.35}& 1.8& 1.96& 0.51& 0.50& 10.5& 1.15& 3.05& 1.19& 8.3& 0.03& 0.62& 0.38& 5.6& 1.83& 0.31& 0.36& 1.8& 1.81 \\
Difference& 9.0\% $\uparrow$& 4.9\% $\uparrow$& \textcolor{black}{84.7\% $\downarrow$}& \textcolor{black}{85.5\% $\downarrow$}& 1.5\% $\uparrow$& \textcolor{blue}{0.5\% $\downarrow$}& \textcolor{black}{90.2\% $\downarrow$}& \textcolor{black}{90.1\% $\downarrow$}& 2.1\% $\uparrow$& 1.1\% $\uparrow$& \textcolor{black}{41.2\% $\downarrow$}& \textcolor{black}{38.9\% $\downarrow$}& 7.2\% $\uparrow$& 5.0\% $\uparrow$& \textcolor{black}{39.5\% $\downarrow$}& \textcolor{black}{39.6\% $\downarrow$}& 1.0\% $\uparrow$& 1.1\% $\uparrow$& \textcolor{black}{74.6\% $\downarrow$}& \textcolor{black}{72.7\% $\downarrow$}& \textcolor{blue}{2.8\% $\downarrow$}& \textcolor{blue}{3.2\% $\downarrow$}& \textcolor{black}{90.0\% $\downarrow$}& \textcolor{black}{90.0\% $\downarrow$} \\
\midrule

Autoformer& 0.24& 0.34& 12.1& 9.30& 0.33& 0.37& 10.5& 19.82& 0.58& 0.53& 10.5& 1.89& 3.08& 1.18& 10.5& 0.05& 0.64& 0.40& 14.9& 6.71& 0.34& 0.38& 10.6& 18.11 \\
+PALS& 0.26& 0.36& 2.7& 1.71& 0.31& 0.35& \textbf{1.0}& \textbf{1.93}& 0.62& 0.55& 7.1& 1.30& 3.19& 1.22& \textbf{6.7}& 0.03& 0.65& 0.41& 4.5& 1.94& 0.34& 0.38& 1.3& 2.52 \\
Difference& 9.3\% $\uparrow$& 4.6\% $\uparrow$& \textcolor{black}{77.7\% $\downarrow$}& \textcolor{black}{81.6\% $\downarrow$}& \textcolor{blue}{8.1\% $\downarrow$}& \textcolor{blue}{5.6\% $\downarrow$}& \textcolor{black}{90.3\% $\downarrow$}& \textcolor{black}{90.3\% $\downarrow$}& 5.5\% $\uparrow$& 4.4\% $\uparrow$& \textcolor{black}{32.7\% $\downarrow$}& \textcolor{black}{31.0\% $\downarrow$}& 3.5\% $\uparrow$& 3.2\% $\uparrow$& \textcolor{black}{36.6\% $\downarrow$}& \textcolor{black}{36.5\% $\downarrow$}& 1.9\% $\uparrow$& 2.1\% $\uparrow$& \textcolor{black}{69.5\% $\downarrow$}& \textcolor{black}{71.0\% $\downarrow$}& 0.1\% $\uparrow$& \textcolor{blue}{1.3\% $\downarrow$}& \textcolor{black}{87.7\% $\downarrow$}& \textcolor{black}{86.1\% $\downarrow$} \\
\midrule

Informer& 0.36& 0.43& 12.5& 8.51& 1.53& 0.88& 11.3& 18.15& 1.59& 1.00& 11.3& 1.71& 5.27& 1.58& 11.3& 0.05& 0.81& 0.46& 14.4& 6.14& 0.62& 0.55& 11.4& 16.58 \\
+PALS& 0.42& 0.48& \textbf{1.4}& \textbf{0.94}& 1.39& 0.83& 5.3& 8.45& 1.53& 0.98& 8.6& 1.33& 5.23& 1.57& 7.4& \textbf{0.03}& 0.94& 0.53& \textbf{2.3}& \textbf{1.19}& 0.69& 0.56& 4.3& 8.35 \\
Difference& 18.9\% $\uparrow$& 11.3\% $\uparrow$& \textcolor{black}{88.6\% $\downarrow$}& \textcolor{black}{88.9\% $\downarrow$}& \textcolor{blue}{9.0\% $\downarrow$}& \textcolor{blue}{5.8\% $\downarrow$}& \textcolor{black}{53.4\% $\downarrow$}& \textcolor{black}{53.4\% $\downarrow$}& \textcolor{blue}{3.9\% $\downarrow$}& \textcolor{blue}{1.6\% $\downarrow$}& \textcolor{black}{24.2\% $\downarrow$}& \textcolor{black}{22.4\% $\downarrow$}& \textcolor{blue}{0.8\% $\downarrow$}& \textcolor{blue}{1.0\% $\downarrow$}& \textcolor{black}{34.9\% $\downarrow$}& \textcolor{black}{34.8\% $\downarrow$}& 15.5\% $\uparrow$& 15.3\% $\uparrow$& \textcolor{black}{83.9\% $\downarrow$}& \textcolor{black}{80.6\% $\downarrow$}& 12.1\% $\uparrow$& 2.0\% $\uparrow$& \textcolor{black}{61.9\% $\downarrow$}& \textcolor{black}{49.7\% $\downarrow$} \\
\midrule

Transformer& 0.28& 0.38& 11.7& 9.24& 1.48& 0.86& 10.5& 19.81& 1.61& 0.97& 10.5& 1.89& 4.94& 1.49& 10.5& 0.05& 0.67& 0.36& 13.6& 6.61& 0.64& 0.56& 10.6& 18.10 \\
+PALS& 0.31& 0.40& 2.5& 2.24& 1.08& 0.75& 3.2& 8.17& 1.41& 0.91& 6.6& 1.16& 4.91& 1.48& 7.7& 0.04& 0.69& 0.38& 3.8& 1.86& 0.32& 0.38& \textbf{1.0}& \textbf{1.76} \\
Difference& 10.5\% $\uparrow$& 5.7\% $\uparrow$& \textcolor{black}{78.3\% $\downarrow$}& \textcolor{black}{75.8\% $\downarrow$}& \textcolor{blue}{26.7\% $\downarrow$}& \textcolor{blue}{13.3\% $\downarrow$}& \textcolor{black}{69.9\% $\downarrow$}& \textcolor{black}{58.8\% $\downarrow$}& \textcolor{blue}{12.5\% $\downarrow$}& \textcolor{blue}{6.1\% $\downarrow$}& \textcolor{black}{37.5\% $\downarrow$}& \textcolor{black}{38.4\% $\downarrow$}& \textcolor{blue}{0.7\% $\downarrow$}& \textcolor{blue}{0.9\% $\downarrow$}& \textcolor{black}{27.2\% $\downarrow$}& \textcolor{black}{27.3\% $\downarrow$}& 3.3\% $\uparrow$& 5.4\% $\uparrow$& \textcolor{black}{71.7\% $\downarrow$}& \textcolor{black}{71.8\% $\downarrow$}& \textcolor{blue}{49.6\% $\downarrow$}& \textcolor{blue}{32.5\% $\downarrow$}& \textcolor{black}{90.2\% $\downarrow$}& \textcolor{black}{90.3\% $\downarrow$} \\
\midrule

Difference\textsubscript{avg} 
&11.7\% $\uparrow$ &6.8\% $\uparrow$ & \textcolor{black}{82.1\% $\downarrow$}& \textcolor{black}{82.5\% $\downarrow$}
& \textcolor{blue}{13.3\% $\downarrow$}& \textcolor{blue}{7.3\% $\downarrow$}& \textcolor{black}{76.1\% $\downarrow$}& \textcolor{black}{74.8\% $\downarrow$}
& \textcolor{blue}{3.6\% $\downarrow$}& \textcolor{blue}{1.2\% $\downarrow$}& \textcolor{black}{36.8\% $\downarrow$}& \textcolor{black}{34.8\% $\downarrow$}
&3.7\% $\uparrow$ &2.3\% $\uparrow$ & \textcolor{black}{33.6\% $\downarrow$}& \textcolor{black}{33.7\% $\downarrow$}
&5.4\% $\uparrow$ &6.6\% $\uparrow$ & \textcolor{black}{74.0\% $\downarrow$}& \textcolor{black}{73.0\% $\downarrow$}
& \textcolor{blue}{10.1\% $\downarrow$}& \textcolor{blue}{9.0\% $\downarrow$}& \textcolor{black}{84.0\% $\downarrow$}& \textcolor{black}{81.3\% $\downarrow$}
\\

\bottomrule

\end{tabular}}
\end{sidewaystable}

%% file: sections/6discussion.tex
\section{Discussion}
\looseness=-1
In this section, we study the performance of PALS in comparison with other pruning and DST algorithms (\ref{ssec:comp_pruning}) and the hyperparameter sensitivity of PALS (\ref{ssec:hyperparameter_sensitivity}). Additionally in the Appendix, we analyze the performance of PALS in terms of model size (\ref{ssec:model_size_effect}), prediction quality by visualizing the predictions (\ref{app:prediction_visualization}), pruning DLinear \cite{zeng2022transformers} (\ref{app:dlinear}), and computational efficiency from various aspects (\ref{app:efficiency}).

\subsection{Performance Comparison with Pruning and Sparse Training Algorithms}\label{ssec:comp_pruning}
\input{Supplementary/7-app-pruning}

\subsection{Hyperparameter Sensitivity}\label{ssec:hyperparameter_sensitivity}
\looseness=-1
In this section, we discuss the sensitivity of PALS to its hyperparameters including pruning rate factor $\gamma$ and loss freedom factor $\lambda$. We have changed their values in $\{1.05, 1.1, 1.2\}$ and measured the performance of PALS (with NSTransformer) in terms of MSE and parameter count on six benchmark datasets. The results are presented in Table \ref{tab:hyperparameters} in Appendix \ref{app:hyperparameters}. 

\looseness=-1
As shown in Table \ref{tab:hyperparameters}, PALS is not very sensitive to its hyperparameters and the results in each row are close in terms of loss in most cases considered. However, by increasing $\gamma$ and $\lambda$ PALS tends to find a sparser model. A small $\lambda$ results in paying more attention to the loss value, while a large value gives more freedom to PALS to explore a sparse sub-network that might sometimes result in a higher loss value. A small $\gamma$ limits the amount of additional grow/prune in the expand/shrink phase, while a large $\gamma$ gives more flexibility to the algorithm for exploring various sparsity levels. In short, a small value for each of these hyperparameters makes PALS more strict and allows for small changes in sparse connectivity, while a large value increases the exploration rate which potentially results in higher sparsity and/or reduced loss.

%% file: Supplementary/7-app-pruning.tex
\looseness=-1
We compare PALS with a standard during-training pruning approach (GMP \cite{zhu2017prune}), GraNet \cite{liu2021sparse}, and a well-known DST method (RigL \cite{evci2020rigging}). These are the closest methods in the literature in terms of including gradual pruning and gradient-based weight regrowth. 

\looseness=-1
While PALS derives a proper sparsity level automatically, other pruning approaches require the sparsity level as an input of the algorithm. Therefore, to compare PALS with existing pruning algorithms, the sparsity level should be optimized for them. We apply GraNet, RigL, and GMP to NSTransformer for prediction lengths of $H \in \{96, 192, 336, 720\}$ (except for the Illness dataset for which $H \in \{24, 36, 48, 60\}$). For each of these methods (GraNet, RigL, and GMP), the sparsity level is optimized among values of $\{25, 50, 65, 80, 90, 95\}$. This means that for one run of PALS, we run the other methods $6$ times. The model with the lowest validation loss is used to report the test loss. Table \ref{tab:comp_with_prune} summarizes the average loss ($l$), sparsity level ($S$), and training epochs ($e$) (due to early stopping the algorithms might not require the full training) over different prediction lengths.


\begin{table}[!b]
\begin{scriptsize}
\caption{Comparison with other during-training pruning methods (GMP, GraNet) and a DST method (RigL) when sparsifying NSTransformer. The results are average over four prediction lengths. \textcolor{blue}{}}\label{tab:comp_with_prune}
\begin{center}
\scalebox{0.9}{
\begin{threeparttable}
\begin{tabular}{c@{\hskip 0.06in}c@{\hskip 0.06in}c@{\hskip 0.06in}c@{\hskip 0.06in}|@{\hskip 0.06in}c@{\hskip 0.06in}c@{\hskip 0.06in}c@{\hskip 0.06in}|@{\hskip 0.06in}c@{\hskip 0.06in}c@{\hskip 0.06in}c@{\hskip 0.06in}|@{\hskip 0.06in}c@{\hskip 0.06in}c@{\hskip 0.06in}c}
    \toprule
 & \multicolumn{3}{c}{\textbf{PALS}} & \multicolumn{3}{c}{\textbf{GraNet\tnote{*}  }} & \multicolumn{3}{c}{\textbf{RigL\tnote{*}}} & \multicolumn{3}{c}{\textbf{GMP\tnote{*}}} \\\midrule
 
 Dataset& \multicolumn{1}{c}{\textbf{l}} & \multicolumn{1}{c}{\textbf{S}} & \multicolumn{1}{c}{\textbf{e}} & \multicolumn{1}{c}{\textbf{l}} & \multicolumn{1}{c}{\textbf{S}} & \multicolumn{1}{c}{\textbf{e}} & \multicolumn{1}{c}{\textbf{l}} & \multicolumn{1}{c}{\textbf{S}} & \multicolumn{1}{c}{\textbf{e}} & \multicolumn{1}{c}{\textbf{l}} & \multicolumn{1}{c}{\textbf{S}} & \multicolumn{1}{c}{\textbf{e}} \\\midrule

\textbf{Electricity} & 0.21 & 80.5\% & 8.83 & 0.20 & 31.2\% & 9.75 & 0.20 & 31.2\% & 9.12 & 0.20 & 47.5\% & 9.62 \\
\textbf{ETTm2} & 0.38 & 76.7\% & 4.58 & 0.60 & 95.0\% & 9.00 & 0.49 & 77.5\% & 4.33 & 0.60 & 56.2\% & 9.12 \\
\textbf{Exchange} & 0.49 & 48.5\% & 5.83 & 0.47 & 95.0\% & 9.42 & 0.44 & 90.0\% & 4.25 & 0.45 & 95.0\% & 9.50 \\
\textbf{Illness} & 2.33 & 30.0\% & 7.97 & 2.32 & 25.0\% & 9.58 & 2.37 & 25.0\% & 9.58 & 2.22 & 31.2\% & 9.92 \\
\textbf{Traffic} & 0.67 & 70.1\% & 8.83 & 0.64 & 41.2\% & 9.50 & 0.64 & 25.0\% & 8.17 & 0.64 & 50.0\% & 9.79 \\
\textbf{Weather} & 0.26 & 90.3\% & 7.00 & 0.28 & 95.0\% & 9.08 & 0.27 & 41.2\% & 4.08 & 0.29 & 95.0\% & 9.08 \\\midrule
\textbf{Average} & \textbf{0.72} & \textbf{66.0\%} & 7.17 & 0.75 & 63.73\% & 9.38 & \multicolumn{1}{c}{0.74} & \multicolumn{1}{c}{48.3\%} & \multicolumn{1}{c}{\textbf{6.60}} & \multicolumn{1}{c}{0.73} & \multicolumn{1}{c}{62.4\%} & \multicolumn{1}{c}{9.47}\\\midrule

\end{tabular}
\begin{tablenotes}\footnotesize
\item[*] \textbf{Optimized sparsity level (\%) in \{25, 65, 50, 80, 90, 95\}. GraNet, RigL, and GMP, each require 6 runs to optimize the sparsity level while PALS needs only one run.} 
\end{tablenotes}
\end{threeparttable}}
\end{center}
\end{scriptsize}
\end{table}

\looseness=-1
The closest competitor of PALS is GraNet. In Table \ref{tab:comp_with_prune}, for the Electricity dataset, PALS achieves a sparsity level of $80.5\%$ with a loss of $0.21$, while GraNet achieves a sparsity level of only $31.2\%$ with a slightly lower loss of $0.20$. Similarly, for the ETTm2 dataset, PALS achieves a sparsity level of $76.7\%$ with a loss of $0.38$, while GraNet achieves a higher sparsity level of $95.0\%$ but with a much higher loss of $0.60$. On the other datasets, they perform relatively close to each other. 

By looking at the results of all methods in Table \ref{tab:comp_with_prune}, PALS has the highest average sparsity value (66.0\%) compared to GraNet (63.73\%), RigL (48.3\%), and GMP (62.4\%). While RigL requires fewer training epochs ($\sim 6.6$ epochs) compared to PALS ($\sim  7.2$ epochs), it finds lower sparsity networks and has a higher average loss (RigL: 0.74 compared to PALS: 0.72). GraNet and GMP use fixed pruning schedules, and as a result, they need almost full training time ($\sim 9.5$ epochs). The only extra computational requirement of PALS compared to GraNet is an additional step that involves determining the number of weights to prune and grow. This is negligible when considering the overall computation necessary for training the models. On the other hand, as PALS does not require the full training epochs in contrast to GraNet, it needs much lower computational costs. We additionally compared the convergence speed of PALS with the dense model in Appendix \ref{app:convergence_speed}.

In short, PALS has the lowest average loss and highest sparsity values compared to other algorithms, suggesting that PALS could build efficient and accurate sparse neural networks for time series forecasting. 

%% file: Supplementary/7appendix.tex
\newpage
\appendix
\onecolumn

\section{Experimental Settings}\label{app:settings}
\input{Supplementary/7-app-settings}

\section{Analyzing Sparsity Effect in Transformers for Time Series Forecasting}\label{app:sparsity_effect_granet}

This section presents the results for the sparsity effect in various time series forecasting transformers. Specifically, we employ GraNet \cite{liu2021sparse} to prune each transformer model, and then evaluate their effectiveness at different sparsity levels.
\begin{figure}[!ht]
\centering  
\subfigure[$H=192$ ($H=36$ for the Illness dataset)]{\label{fig:granet_sparsity_effect_all_pl_0}\includegraphics[width=0.6\columnwidth]{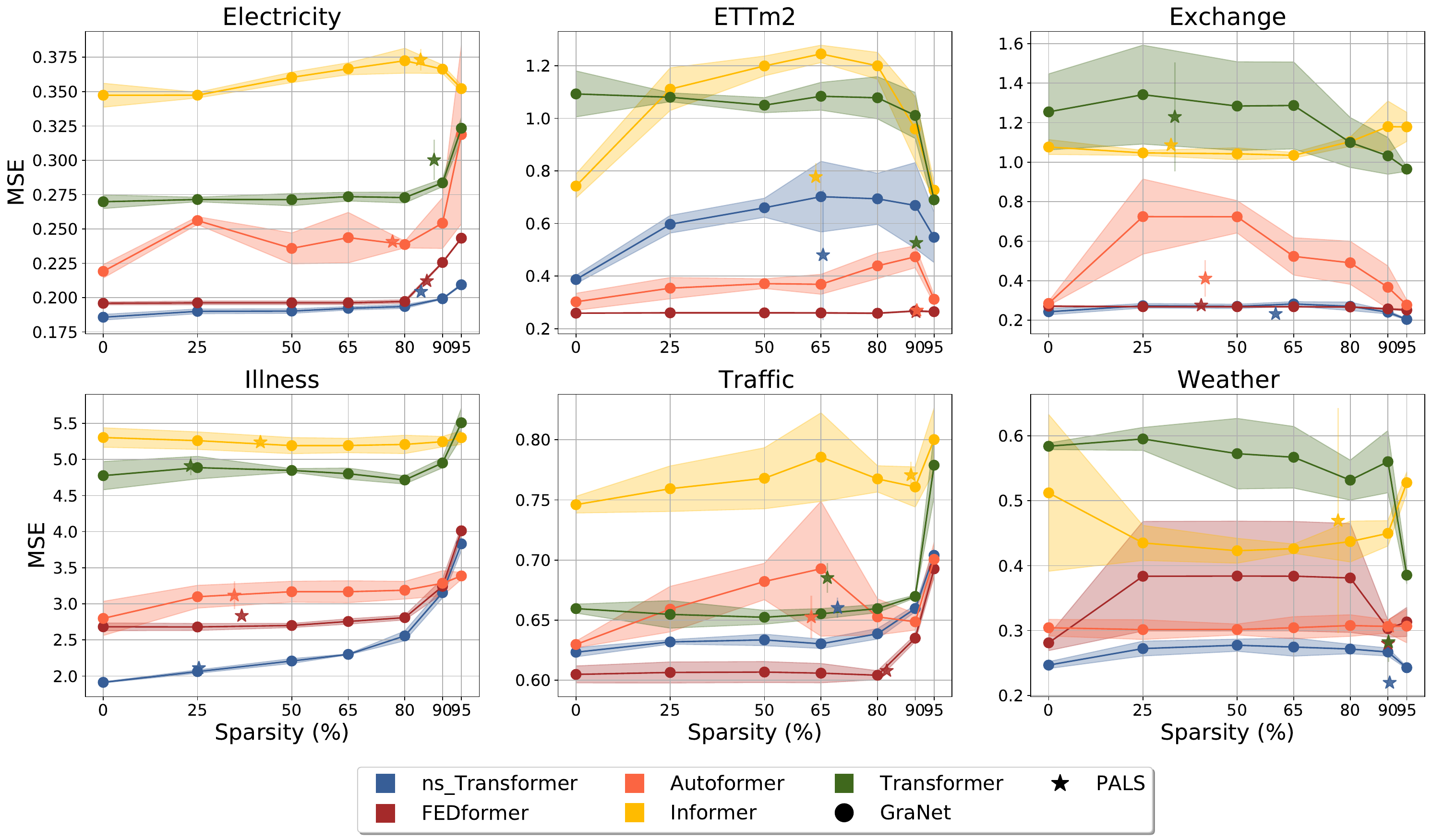}}
\subfigure[$H=336$ ($H=48$ for the Illness dataset)]{\label{fig:granet_sparsity_effect_all_pl_2}\includegraphics[width=0.6\columnwidth]{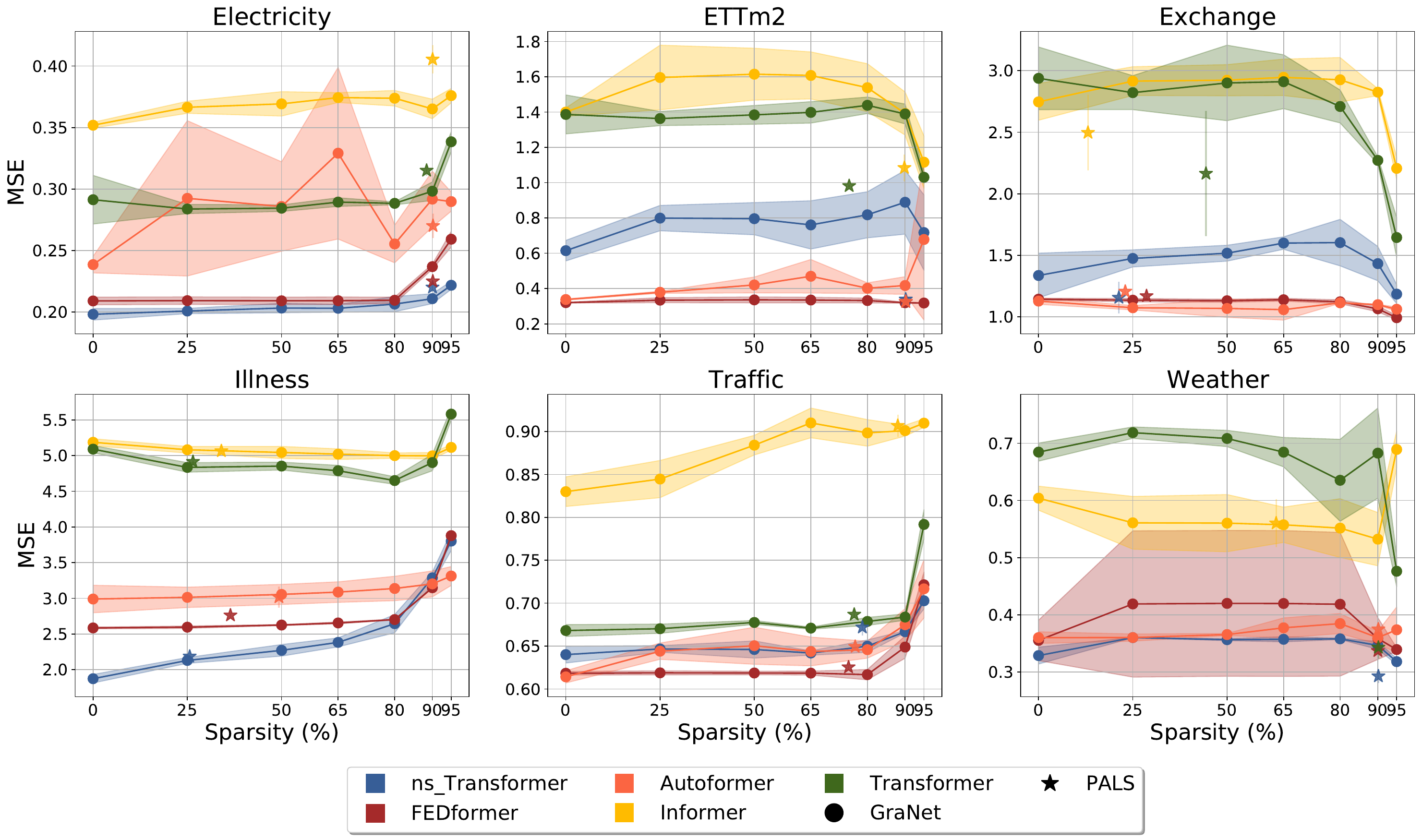}}
\subfigure[$H=720$ ($H=60$ for the Illness dataset)]{\label{fig:granet_sparsity_effect_all_pl_3}\includegraphics[width=0.6\columnwidth]{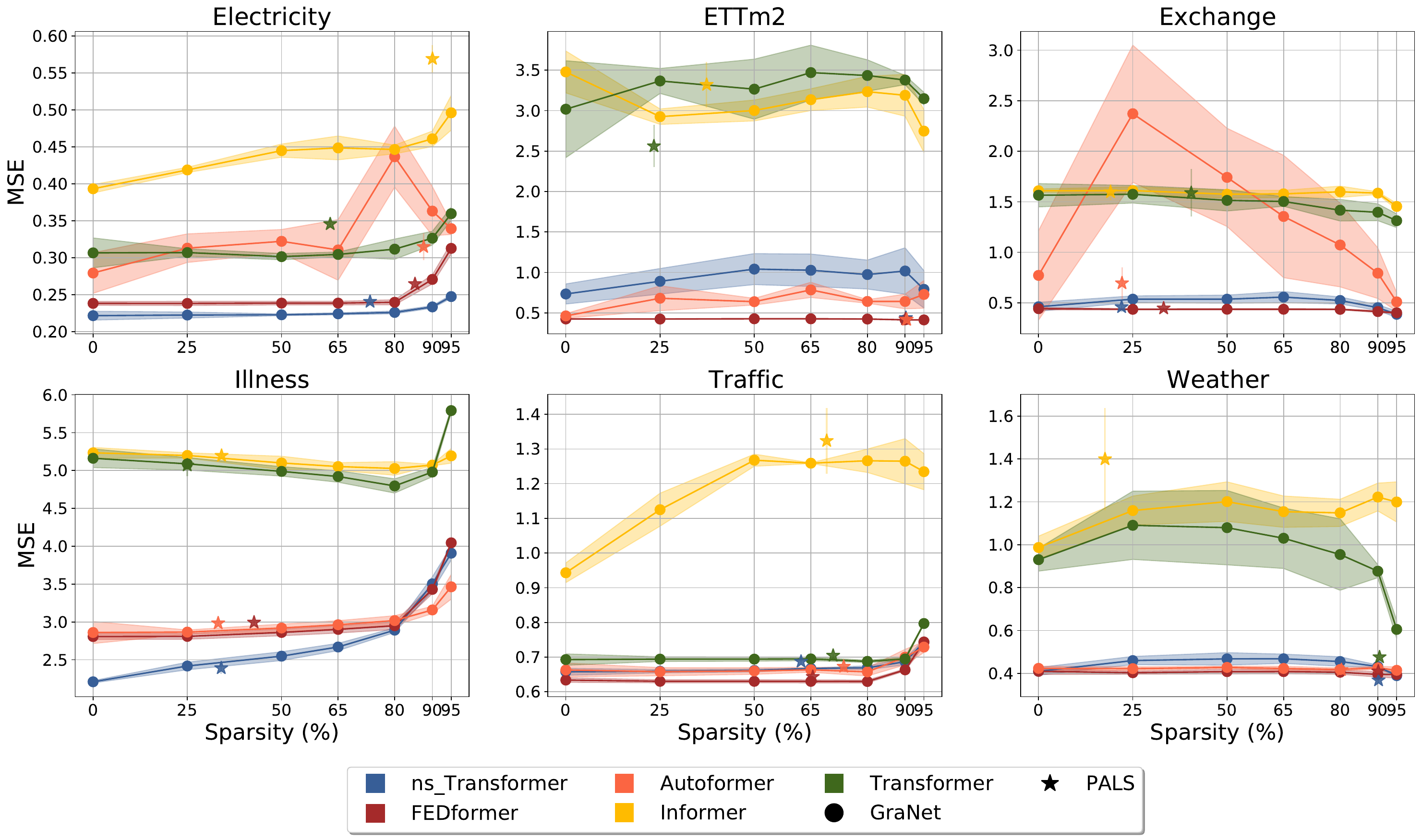}}
\caption{Sparsity effect on the performance of various transformer models for time series forecasting on benchmark datasets in terms of MSE loss for various prediction lengths as indicated in each figure. Each model is sparsified using GraNet \cite{liu2021sparse} to sparsity levels ($\%$) $\in \{25, 50, 65, 80, 90, 95\}$. Sparsity=0 indicates the original dense model.}
\label{fig:granet_sparsity_effect_all}
\end{figure}

GraNet gradually prunes a network  (in this paper, we start from a dense network) during the training to reach a pre-determined sparsity level while allowing for connection regeneration. Therefore, it takes advantage of dense-to-sparse and sparse-to-sparse training by exploring faster the search space and dynamically optimizing the sparse connectivity during training, respectively, to find a decent connectivity pattern efficiently. During training, GraNet executes gradual pruning and zero-cost Neuroregeneration every $\delta t$ iterations. Gradual pruning gradually reduces network density towards a specific sparsity level across multiple pruning iterations. The initial sparsity level can be zero (creating a dense network, resulting in a dense-to-sparse approach) or higher (starting from a random sparse topology, resulting in a sparse-to-sparse approach). At each pruning step, a portion of weights with the lowest magnitudes is pruned, based on a fixed schedule. This stage is similar to gradual magnitude pruning (GMP) \cite{zhu2017prune}. Following each pruning step, a zero-cost Neuroregeneration is executed. This involves dropping a portion of the existing connections with low magnitudes, which are considered damaged, and adding an equal number of new connections back to the network. The new connections are chosen from non-existing connections with the highest gradient value. 

The results for the sparsity effect on the performance of various transformer models are presented in Figure \ref{fig:sparsity_effect_granet} in the paper and Figure \ref{fig:granet_sparsity_effect_all}. The findings are discussed in Section \ref{sec:sparsity_effect}.


\newpage
\section{Comparison Results}\label{app:comp_results}
The detailed results of the experiments performed in Section \ref{sec:results} are presented in Table \ref{tab:results_all}. 
\input{assets/tables/adapt_tuned_detailed}

\section{Univariate results}\label{app:univariate}
The results are presented in Figures \ref{tab:results_all_univariate} and \ref{tab:summary_results|_univariate} and discussed in Section \ref{sec:results} in the paper.
\input{assets/tables/adapt_tuned_detailed_univariate.tex}

\input{assets/tables/summary_comparison_models_univariate.tex}

\newpage
\section{Hyperparameter Sensitivity Analysis}\label{app:hyperparameters}
\looseness=-1
In this Appendix, we present the results for the hyperparameter sensitivity of PALS. We vary the values of $\gamma$ and $\lambda$ in $\{1.05, 1.1, 1.2\}$. The results are presented in Table \ref{tab:hyperparameters}. The results are discussed in Section \ref{ssec:hyperparameter_sensitivity}.

\input{assets/tables/hyperparameter_sensitivity}

\section{Pruning DLinear with PALS}\label{app:dlinear}

In this Appendix, we train PALS with the DLinear \cite{zeng2022transformers} model which is a MLP-based model for time series forecasting and has proven to be effective across various datasets. While our focus in this paper is to reduce the complexity of transformers for time series forecasting, we want to show the generality of our proposed approach to other models. We demonstrate that PALS can be also applied to these models (which are computationally cheaper than transformers) to decrease the model size even more. As an example, we apply PALS to DLinear.

In Table \ref{tab:dlinear}, the results of applying PALS to DLinear in terms of MSE and the sparsity level are presented. In most cases considered, PALS can prune DLinear without compromising loss. On the Electricity, ETTm2, Traffic, and Weather datasets, PALS can prune $\sim 90\%$ of the connections while achieving comparable loss. This shows the effectiveness of PALS when applied to a MLP-based model.

Finally, we want to highlight that our goal in this paper is not to propose a new forecasting model and beat the state-of-the-art for time series forecasting. Instead, we aim to decrease the high computational costs of models for time series forecasting while finding automatically a decent sparsity level and potentially improving their generalization. The reason that we focus on transformers is that they are considered to be computationally expensive while performing well in time series forecasting, and as shown in \cite{wu2022timesnet}, transformer-based models perform generally well in other time series analysis tasks, including, classification, anomaly detection, and imputation compared to the MLP-based models \cite{zeng2022transformers}. Therefore, they can be a promising direction for future time series analysis research. However, PALS is orthogonal to forecasting models and can be applied to any deep learning-based model to reduce its computational costs.

\begin{table}[H]
\caption{Effectiveness of PALS for pruning DLinear model \cite{zeng2022transformers}. Each row presents the results of DLinear before and after applying PALS in terms of MSE, for each prediction length. The achieved sparsity level is shown in parenthesis as \%.}\label{tab:dlinear}
\centering 
\resizebox{\textwidth}{!}{
\begin{tabular}{c|c|cccccc}
\toprule
\textbf{} & \textbf{Model} & \textbf{Electricity} & \textbf{ETTm2} & \textbf{Exchange} & \textbf{Illness} & \textbf{Traffic} & Weather \\
\midrule
\textbf{96/24} & \textbf{DLinear} & 0.140 & 0.172 & 0.094 & 1.997 & 0.413 & 0.175 \\
\textbf{} & \textbf{+PALS} & 0.141 (90.2\%) & 0.181 (90.0\%) & 0.086 (25.5\%) & 1.984 (34.1\%) & 0.412 (90.3\%) & 0.177 (90.2\%) \\\midrule
\textbf{192/36} & \textbf{DLinear} & 0.153 & 0.235 & 0.168 & 2.090 & 0.424 & 0.217 \\
\textbf{} & \textbf{+PALS} & 0.154 (90.2\%) & 0.249 (90.1\%) & 0.173 (20.8\%) & 2.130 (29.8\%) & 0.424 (90.2\%) & 0.218 (90.2\%) \\\midrule
\textbf{336/48} & \textbf{DLinear} & 0.169 & 0.307 & 0.322 & 2.058 & 0.437 & 0.263 \\
\textbf{} & \textbf{+PALS} & 0.170 (90.2\%) & 0.301 (82.7\%) & 0.324 (24.7\%) & 2.124 (32.6\%) & 0.436 (90.1\%) & 0.262 (90.2\%) \\\midrule
\textbf{720/60} & \textbf{DLinear} & 0.204 & 0.390 & 0.959 & 2.375 & 0.467 & 0.328 \\
\textbf{} & \textbf{+PALS} & 0.204 (90.1\%) & 0.433 (90.3\%) & 0.963 (38.0\%) & 2.358 (44.3\%) & 0.467 (90.2\%) & 0.324 (90.2\%)\\
\bottomrule
\end{tabular}}
\end{table}


\section{Efficiency of PALS}\label{app:efficiency}

\input{Supplementary/7-app-efficiency}

\begin{figure}[!b]
\begin{minipage}{0.5\textwidth}
    \centering
    \includegraphics[width=\textwidth]{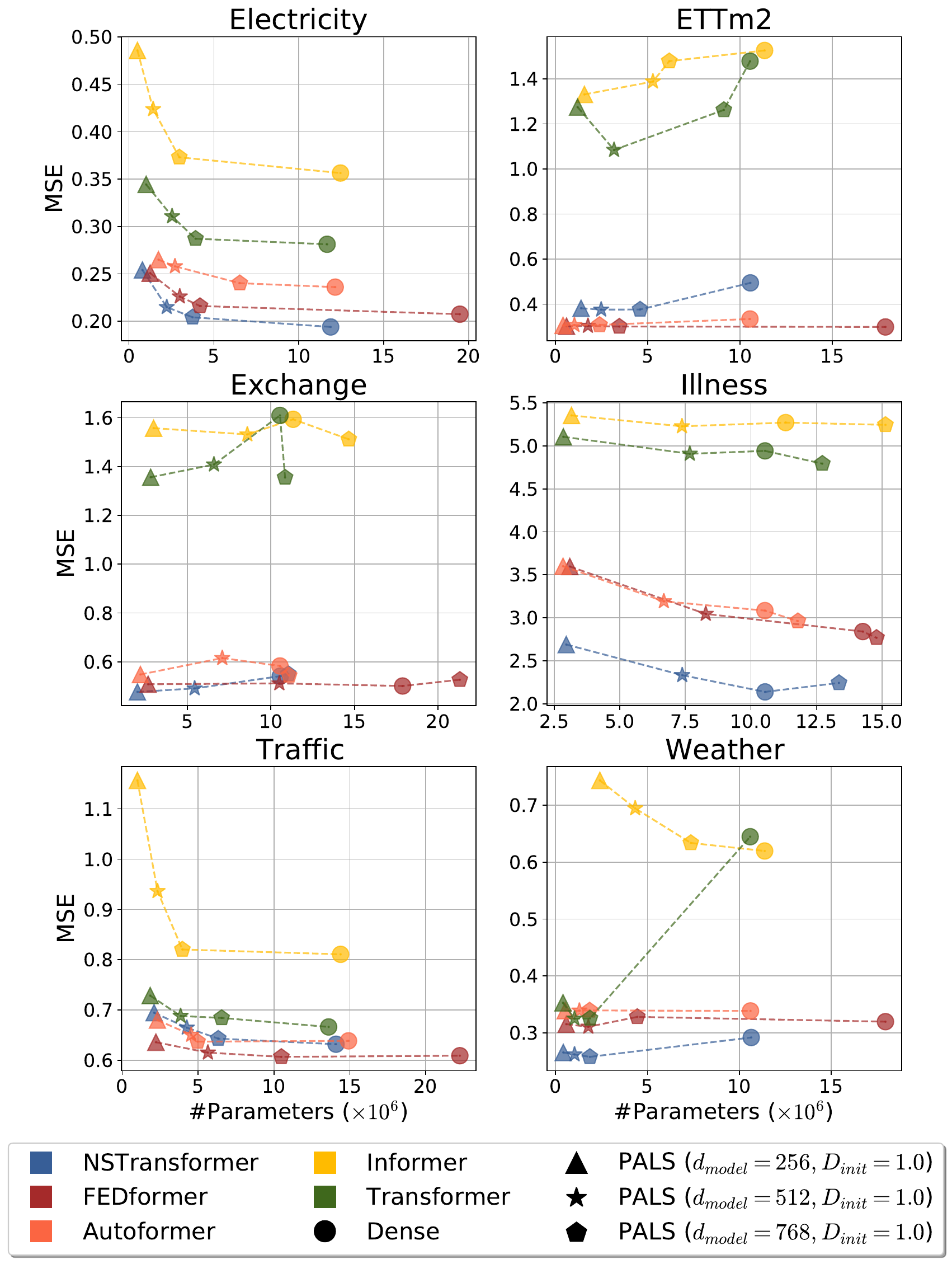}
    \vspace{-1mm}
    \caption{Model size effect by varying $d_{model} \in \{256, 512 (default), 768\}$) on the prediction performance of PALS compared to the original dense model.}
    \label{fig:summary_hidden_size}
\end{minipage}%
\hspace{5 pt}
\begin{minipage}{0.5\textwidth}
    \subfigure[\tiny{Weather}]{\label{fig:weather_Transformer}\includegraphics[width=28mm]{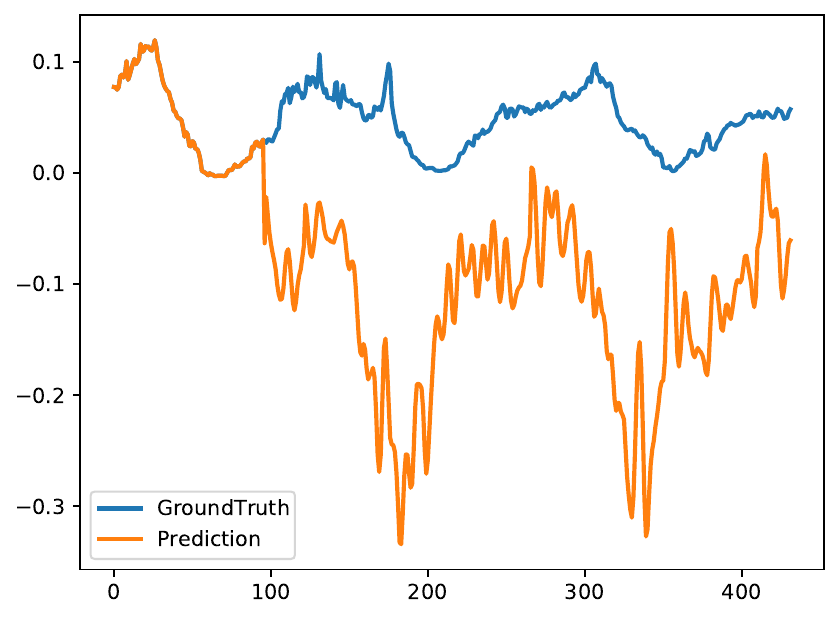}}
    \subfigure[\tiny{Weather (+PALS)}]{\label{fig:weather_Transformer_PALS}\includegraphics[width=28mm]{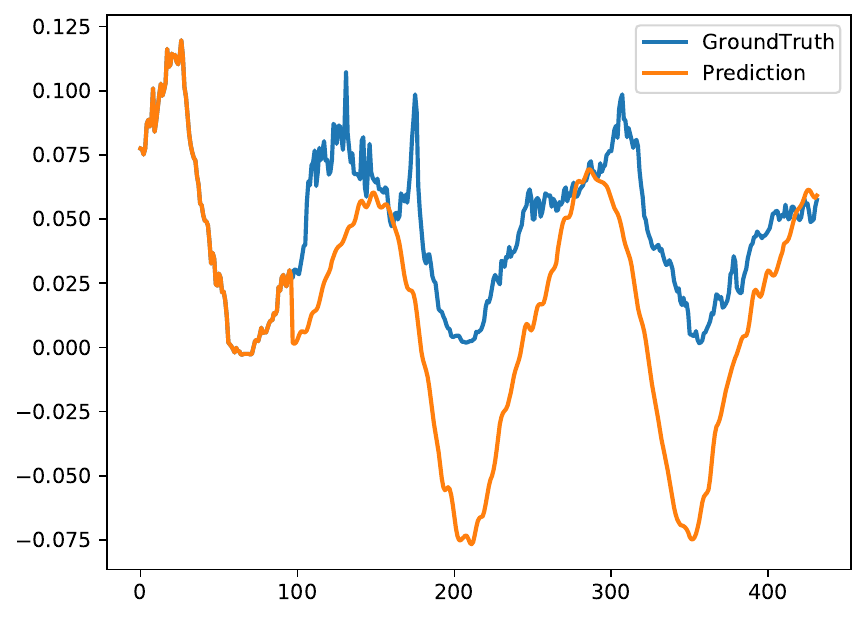}}\\\vspace{-3mm}
    \subfigure[\tiny{ETTm2}]{\label{fig:weather_NSTransformer}\includegraphics[width=28mm]{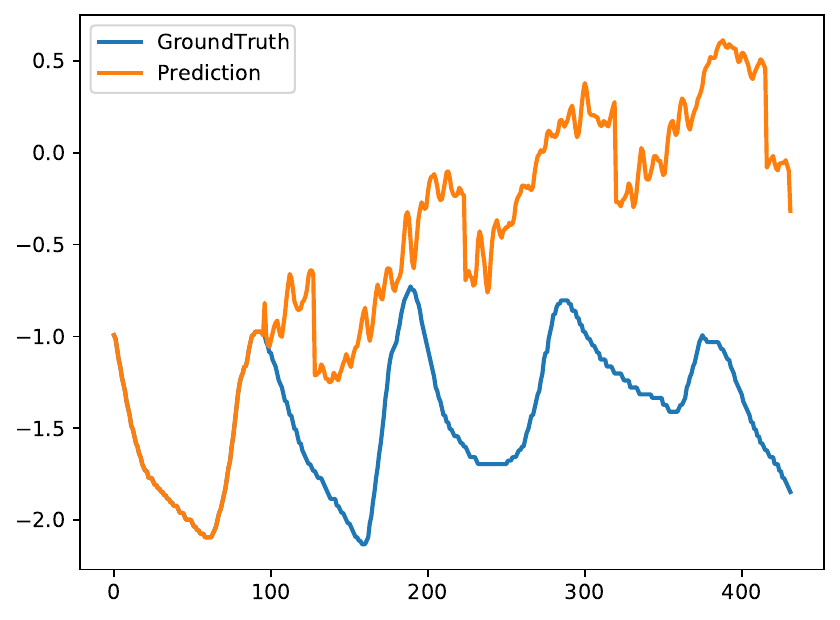}}
    \subfigure[\tiny{ETTm2 (+PALS)}]{\label{fig:weather_NSTransformer_PALS}\includegraphics[width=28mm]{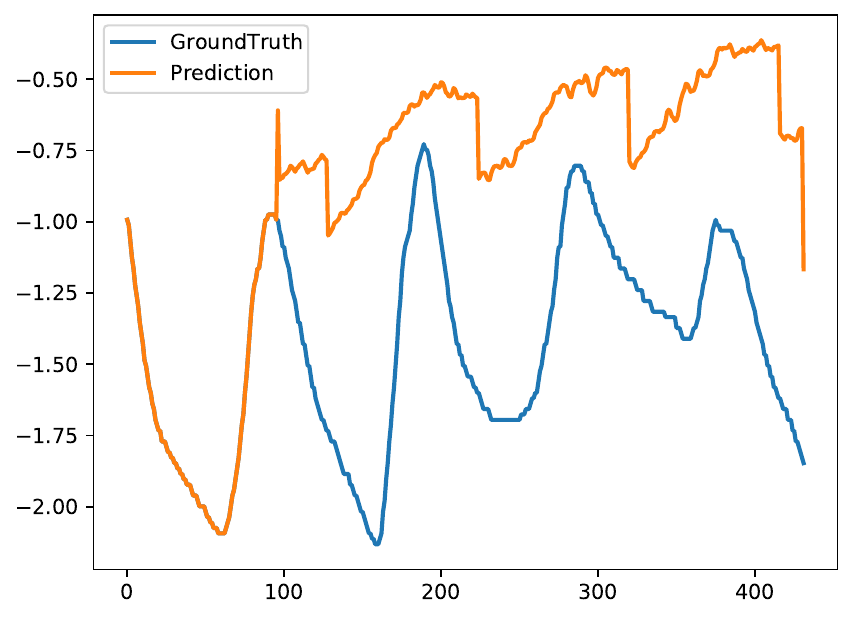}}\\\vspace{-2mm}
    \subfigure[\tiny{Exchange}]{\label{fig:weather_Autoformer}\includegraphics[width=28mm]{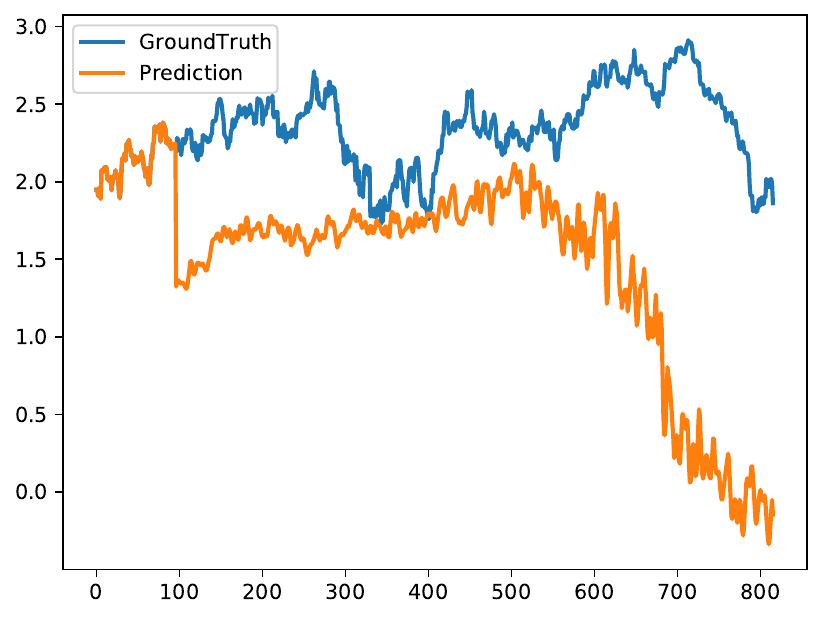}}
    \subfigure[\tiny{Exchange (+PALS)}]{\label{fig:weather_Autoformer_PALS}\includegraphics[width=28mm]{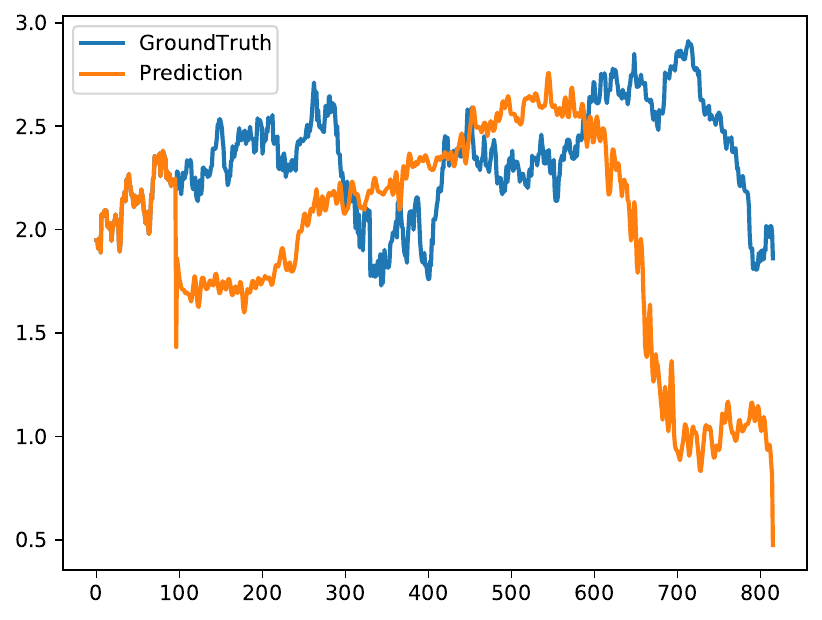}}
    \caption{Forecasting Visualization on Transformer with and without PALS on three datasets.}
    \label{fig:weather_prediction}
\end{minipage}
\end{figure}

\section{Model Size Effect}\label{ssec:model_size_effect}
\looseness=-1
In this Appendix, we study the effect of model size on the performance of PALS by changing the hidden dimension ($d_{model}$) in $\{256, 512 (default), 768\}$ to analyze the trade-off between loss and parameters count. We have also considered the results of the original dense model ($d_{model}=512$). The results for all models in terms of MSE and parameters count (which are the average results over different prediction lengths) are presented in Figure \ref{fig:summary_hidden_size}.

\looseness=-1
In Figure \ref{fig:summary_hidden_size}, we can observe that in most cases considered, the model can be pruned significantly with little or no increase in loss. Similar to our discussion in Section \ref{sec:sparsity_effect}, we see consistent behavior among the Electricity, Illness, and traffic datasets, where a higher number of parameters is mostly in favor of loss reduction. This might come from the inherent complexity of predicting these datasets due to either a high number of variables (such as Electricity and Traffic datasets) or the relatively non-stationary nature of data (such as the illness datasets as shown in \cite{liu2022non}). While being a non-stationary dataset, on the Exchange dataset, a small ($d_{model}=256$) sparse model can perform very closely or better than the original ($d_{model}=512$) dense model on average among various transformer architectures. This might be caused by the abilities of the transformer variants to learn this behavior such that even a pruned small model can substitute this model. On the other datasets (ETTm2 and Weather which are relatively stationary with a low number of variables), sparsity results mostly in a better performance than the dense model.

\looseness=-1
In short, it can be concluded that various time-series datasets do not demonstrate homogenous behavior due to their intrinsic differences. Therefore, we need to take these differences into consideration when choosing the model size or the right sparsity level (Section \ref{sec:sparsity_effect}). 

\section{Prediction Quality}\label{app:prediction_visualization}

\looseness=-1
In this Appendix, we evaluate the prediction of different models and discuss how PALS affects this prediction when pruning them. 

\looseness=-1
The predictions for the transformer on the Weather, ETTm2, and Exchange datasets, where the most significant changes in loss and parameters count were seen, are visualized in Figure \ref{fig:weather_prediction}. It can be observed that PALS significantly improves the prediction of this model on these three datasets. As summarized in Table \ref{tab:results_all} (Appendix \ref{app:comp_results}), PALS is able to gain very close prediction performance on the transformer to the best performer on the Weather dataset (NStransformer) with only pruning $90\%$ of unimportant connections.

\looseness=-1
Next, we visualize the predictions for each model with and without PALS on the Weather, Illness, ETTm2 (periodic dataset), and Exchange (without obvious periodicity) datasets for all transformer variants in Figures \ref{fig:Weather_all_prediction}, \ref{fig:Illness_all_prediction}, \ref{fig:ETTm2_all_prediction}, and \ref{fig:Exchange_all_prediction}, respectively. In most cases considered in these figures, PALS is able to gain very similar or better performance than the dense counterpart model. By removing unimportant connections, the prediction using PALS is mostly smoother than the prediction of the dense model. Therefore, it can be concluded that it is possible to significantly prune transformers for time series forecasting using PALS without compromising performance.

\input{Supplementary/7-app_prediction}

%% file: Supplementary/7-app-settings.tex
\subsection{Datasets}\label{app:datasets}  
The datasets are summarized in Table \ref{tab:datasets}. 1) \textit{Electricity} \footnote{\url{https://archive.ics.uci.edu/ml/datasets/ElectricityLoadDiagrams20112014}} dataset includes the hourly electricity consumption for 321 consumers between 2012 and 2014. 2) \textit{ETT} \cite{zhou2021informer} (Electricity Transformer Temperature) dataset contains load and oil temperature measurements from electricity transformers. 3) \textit{Exchange} \cite{lai2018modeling} dataset consists of daily exchange rates from 8 countries between 1990 and 2016. 4) \textit{Illness} \footnote{\url{https://gis.cdc.gov/grasp/fluview/fluportaldashboard.html}} dataset includes weekly collected data from influenza-like illness patients between 2002 and 2021 reported by Centers for Disease Control and Prevention of the United States. 5) \textit{Traffic} \footnote{\url{https://pems.dot.ca.gov/}} dataset contains road occupancy rates in San Francisco Bay area freeways. 6) \textit{Weather} \footnote{\url{https://www.bgc-jena.mpg.de/wetter/}} consists of measurements of 21 weather indicators collected every 10 minutes in 2020. All datasets are divided in chronological order into the train, validation, and test sets with a split ratio of 7:2:2 (except for the ETT  dataset where we use 6:2:2 split ratio).

\input{assets/tables/datasets}

Datasets in Table \ref{tab:datasets} have different characteristics. Such characteristics include: 1) \textit{Sampling frequency}: while the sampling frequency for some time series datasets is very high (e.g., Electricity (1 hour) and Weather datasets (10 minutes)), it might be very low for some others (Exchange (1 day) and Illness (1 week)) 2) \textit{Periodicity of the variables}: time series datasets can be periodic (ETTm2) or without obvious periodicity (Exchange) 3) \textit{Number of variables}: the number of variables can vary significantly. Some datasets have below 10 variables (ETT, Illness) while others have in order of hundreds (Traffic, Electricity). This characteristic results in different levels of complexity. Therefore, automatically tuning the sparsity can help to tune the complexity of the task at hand that eventually help prevent overfitting in simple tasks (e.g., Weather) and maintaining overparameterization for complex tasks (e.g., Traffic, Electricity). The beauty of our proposed method consists in the fact that it does not have to consider any of these intrinsic differences. We did not make any finetuning for PALS to account for these differences, and it does everything automatically. Of course, finetuning PALS per dataset specificity would improve its final performance, but it would reduce the generality of our proposed work and we prefer not to do it.
 
\subsection{Prediction Quality Evaluation Metrics} 
We use MSE and MAE as the evaluation metrics, which can be computed as below:

\begin{equation}
    MSE(\widetilde{\mX}_{t:t+H}, \mX_{t:t+H}) = \frac{1}{H}\Sigma_{i=0}^{H-1} (\widetilde{\vx}_{t+i} - \vx_{t+i})^2.
\end{equation}

\begin{equation}
    MAE(\widetilde{\mX}_{t:t+H}, \mX_{t:t+H}) = \frac{1}{H}\Sigma_{i=0}^{H-1} |\widetilde{\vx}_{t+i} - \vx_{t+i}|.
\end{equation}

\subsection{Hyperparameters} 
The settings of the transformer models and the hyperparameter values are adopted from the NSTransformer implementation\footnote{\url{https://github.com/thuml/Nonstationary_Transformers}}. Sequence length $L$ was set to $36$ for the Illness dataset and $96$ for the other datasets. Several values for prediction length were tested in the experiments: $H \in \{96, 192, 336, 720\}$ (except for the Illness dataset for which $H \in \{24, 36, 48, 60\}$). The models considered in the experiments are all trained with the ADAM optimizer with a learning rate of $10^{-4}$. The batch size used in the experiments was equal to $32$. A maximum number of $10$ epochs was used for each experiment. Each transformer model consisted of two encoder layers and one decoder layer. Model dimension $d_{model}$ was set to $512$ in the experiments unless stated otherwise. PALS starts from a dense model (Initial density $D_{init} = 1$). The pruning rate $\zeta$ is initialized to $0.5$ and decreased during training with a cosine decay schedule. The values of pruning rate factor $\gamma$ and loss freedom factor $\lambda$ are optimized in $\{1.05, 1.1, 1.2\}$ on the validation set using a single random seed for each experiment. Minimum sparsity $S_{min}$ and  maximum sparsity$S_{max}$ are set to $20$ and $90$, respectively. These values give the flexibility to the user to tune the range of the model sparsity based on the resource availability in their application. The mask update frequency $\Delta t$ was equal to $5$ for the Illness dataset and $20$ for the other datasets. Each experiment was run on three different random seeds and the average measurements are reported for each metric.

%% file: assets/tables/datasets.tex
\begin{table}[!h]
\vspace{0mm}
\centering
\vspace{-3mm}
\caption{Datasets.}
\label{tab:datasets}
\scalebox{0.75}{
    \begin{tabular}{cccc}
    \toprule
    \textbf{Dataset} &\textbf{ \# Variables} &\textbf{Sampling Frequency} &\textbf{ \# Observations}\\\midrule
    Electricity     & 321     & 1 Hour     &26304\\
    ETTm2     & 7     & 15 Miutes     &69680\\
    Exchange     & 8     & 1 Day     &7588\\
    Illness     & 7     & 1 Week     &966\\
    Traffic     & 862     & 1 Hour     &17544\\
    Weather     & 21     & 10 Minutes     &52695\\
    \bottomrule
    \end{tabular}
}
\end{table}

%% file: assets/tables/adapt_tuned_detailed.tex
\begin{table}[!h]

\caption{Comparison on the benchmark Datasets for prediction length $H \in \{96, 192, 336, 720\}$ (except for the Illness dataset for which $H \in \{24, 36, 48, 60\}$). For each model, the performance of the original (Dense) and the pruned model using PALS is presented in terms of MSE and parameter count ($\times 10^{6}$). The difference between the results compared to the dense counterpart is shown in parenthesis as \% (\textcolor{blue}{blue} indicates improvement in the results compared to the dense model). \textbf{Bold} entries are the best performer in each experiment among various models. }\label{tab:results_all}
\centering
\resizebox{1\textwidth}{!}{
\begin{tabular}{@{\hskip 0.0in}c@{\hskip 0.03in}c@{\hskip 0.03in}|@{\hskip 0.03in}c@{\hskip 0.03in}c@{\hskip 0.03in}c@{\hskip 0.03in}c@{\hskip 0.03in}|@{\hskip 0.03in}c@{\hskip 0.03in}c@{\hskip 0.03in}c@{\hskip 0.03in}c@{\hskip 0.03in}|@{\hskip 0.03in}c@{\hskip 0.03in}@{\hskip 0.03in}c@{\hskip 0.03in}c@{\hskip 0.03in}c|c@{\hskip 0.03in}c@{\hskip 0.03in}c@{\hskip 0.03in}c@{\hskip 0.03in}|@{\hskip 0.03in}c@{\hskip 0.03in}c@{\hskip 0.03in}c@{\hskip 0.03in}c}
\toprule
\bt \multirow{3}{*}{\rotatebox[origin=c]{90}{Datasets}}& \bt \multirow{3}{*}{$H$} &\multicolumn{4}{c}{NSTransformer} &\multicolumn{4}{c}{FEDformer} &\multicolumn{4}{c}{Autoformer}&\multicolumn{4}{c}{Informer} &\multicolumn{4}{c}{Transformer}\\

&&\multicolumn{2}{c}{Dense} &\multicolumn{2}{c}{+PALS}       &\multicolumn{2}{c}{Dense} &\multicolumn{2}{c}{+PALS}  &\multicolumn{2}{c}{Dense} &\multicolumn{2}{c}{+PALS}       &\multicolumn{2}{c}{Dense} &\multicolumn{2}{c}{+PALS} &\multicolumn{2}{c}{Dense} &\multicolumn{2}{c}{+PALS}\\

& &MSE &\#Param  &MSE &\#Param       &MSE &\#Param  &MSE &\#Param       &MSE &\#Param  &MSE &\#Param          &MSE &\#Param  &MSE &\#Param      &MSE &\#Param  &MSE &\#Param  \\
\midrule 

\multirow{4}{*}{\rotatebox[origin=c]{90}{Electricity}}& 96& \pmb{0.170}& 12.2& 0.195 (\textcolor{black}{14.3\%} $\uparrow$)& 2.6  (\textcolor{black}{78.2\%} $\downarrow$)& 0.187& 19.5& 0.203 (\textcolor{black}{8.6\%} $\uparrow$)& 4.4  (\textcolor{black}{77.3\%} $\downarrow$)& 0.207& 12.1& 0.206 (\textcolor{blue}{0.4\%} $\downarrow$)& 5.3  (\textcolor{black}{56.2\%} $\downarrow$)& 0.333& 12.5& 0.347 (\textcolor{black}{4.3\%} $\uparrow$)& 1.2  (\textcolor{black}{90.1\%} $\downarrow$)& 0.257& 11.7& 0.282 (\textcolor{black}{9.6\%} $\uparrow$)& 3.0  (\textcolor{black}{73.9\%} $\downarrow$) \\
& 192& \pmb{0.186}& 12.2& 0.204 (\textcolor{black}{10.0\%} $\uparrow$)& 1.9  (\textcolor{black}{84.3\%} $\downarrow$)& 0.196& 19.5& 0.212 (\textcolor{black}{8.3\%} $\uparrow$)& 2.8  (\textcolor{black}{85.9\%} $\downarrow$)& 0.219& 12.1& 0.241 (\textcolor{black}{9.9\%} $\uparrow$)& 2.8  (\textcolor{black}{76.7\%} $\downarrow$)& 0.347& 12.5& 0.373 (\textcolor{black}{7.4\%} $\uparrow$)& 2.0  (\textcolor{black}{84.2\%} $\downarrow$)& 0.270& 11.7& 0.300 (\textcolor{black}{11.3\%} $\uparrow$)& 1.4  (\textcolor{black}{87.8\%} $\downarrow$) \\
& 336& \pmb{0.198}& 11.9& 0.220 (\textcolor{black}{11.1\%} $\uparrow$)& 1.2  (\textcolor{black}{90.0\%} $\downarrow$)& 0.209& 19.5& 0.225 (\textcolor{black}{7.7\%} $\uparrow$)& 1.9  (\textcolor{black}{90.1\%} $\downarrow$)& 0.239& 12.1& 0.270 (\textcolor{black}{13.2\%} $\uparrow$)& 1.2  (\textcolor{black}{90.2\%} $\downarrow$)& 0.352& 12.5& 0.405 (\textcolor{black}{15.2\%} $\uparrow$)& 1.2  (\textcolor{black}{90.1\%} $\downarrow$)& 0.291& 11.7& 0.315 (\textcolor{black}{8.2\%} $\uparrow$)& 1.3  (\textcolor{black}{88.5\%} $\downarrow$) \\
& 720& \pmb{0.222}& 11.9& 0.241 (\textcolor{black}{8.6\%} $\uparrow$)& 3.2  (\textcolor{black}{73.5\%} $\downarrow$)& 0.238& 19.5& 0.265 (\textcolor{black}{11.0\%} $\uparrow$)& 2.8  (\textcolor{black}{85.4\%} $\downarrow$)& 0.279& 12.1& 0.315 (\textcolor{black}{12.7\%} $\uparrow$)& 1.5  (\textcolor{black}{87.7\%} $\downarrow$)& 0.393& 12.5& 0.569 (\textcolor{black}{44.8\%} $\uparrow$)& 1.2  (\textcolor{black}{90.0\%} $\downarrow$)& 0.307& 11.7& 0.346 (\textcolor{black}{12.7\%} $\uparrow$)& 4.3  (\textcolor{black}{62.9\%} $\downarrow$) \\
\midrule
\multirow{4}{*}{\rotatebox[origin=c]{90}{ETTm2}}& 96& 0.241& 10.6& 0.244 (\textcolor{black}{1.0\%} $\uparrow$)& 4.1  (\textcolor{black}{61.0\%} $\downarrow$)& \pmb{0.190}& 17.9& 0.201 (\textcolor{black}{5.7\%} $\uparrow$)& 1.8  (\textcolor{black}{90.2\%} $\downarrow$)& 0.233& 10.5& 0.214 (\textcolor{blue}{7.9\%} $\downarrow$)& 1.0  (\textcolor{black}{90.1\%} $\downarrow$)& 0.485& 11.3& 0.371 (\textcolor{blue}{23.3\%} $\downarrow$)& 8.7  (\textcolor{black}{22.9\%} $\downarrow$)& 0.418& 10.5& 0.268 (\textcolor{blue}{35.8\%} $\downarrow$)& 1.0  (\textcolor{black}{90.8\%} $\downarrow$) \\
& 192& 0.387& 10.7& 0.480 (\textcolor{black}{24.1\%} $\uparrow$)& 3.7  (\textcolor{black}{65.6\%} $\downarrow$)& \pmb{0.259}& 17.9& 0.263 (\textcolor{black}{1.7\%} $\uparrow$)& 1.8  (\textcolor{black}{90.2\%} $\downarrow$)& 0.302& 10.5& 0.269 (\textcolor{blue}{10.8\%} $\downarrow$)& 1.0  (\textcolor{black}{90.4\%} $\downarrow$)& 0.742& 11.3& 0.777 (\textcolor{black}{4.7\%} $\uparrow$)& 4.1  (\textcolor{black}{63.6\%} $\downarrow$)& 1.093& 10.5& 0.527 (\textcolor{blue}{51.8\%} $\downarrow$)& 1.0  (\textcolor{black}{90.3\%} $\downarrow$) \\
& 336& 0.615& 10.6& 0.339 (\textcolor{blue}{44.9\%} $\downarrow$)& 1.0  (\textcolor{black}{90.2\%} $\downarrow$)& \pmb{0.320}& 17.9& 0.325 (\textcolor{black}{1.4\%} $\uparrow$)& 1.8  (\textcolor{black}{90.2\%} $\downarrow$)& 0.338& 10.5& 0.326 (\textcolor{blue}{3.4\%} $\downarrow$)& 1.0  (\textcolor{black}{90.2\%} $\downarrow$)& 1.400& 11.3& 1.084 (\textcolor{blue}{22.5\%} $\downarrow$)& 1.2  (\textcolor{black}{89.8\%} $\downarrow$)& 1.387& 10.5& 0.982 (\textcolor{blue}{29.2\%} $\downarrow$)& 2.6  (\textcolor{black}{75.2\%} $\downarrow$) \\
& 720& 0.734& 10.6& 0.439 (\textcolor{blue}{40.1\%} $\downarrow$)& 1.0  (\textcolor{black}{90.2\%} $\downarrow$)& 0.425& 17.9& 0.423 (\textcolor{blue}{0.4\%} $\downarrow$)& 1.8  (\textcolor{black}{90.1\%} $\downarrow$)& 0.463& 10.5& \pmb{0.417} (\textcolor{blue}{9.9\%} $\downarrow$)& 1.0  (\textcolor{black}{90.4\%} $\downarrow$)& 3.479& 11.3& 3.320 (\textcolor{blue}{4.5\%} $\downarrow$)& 7.1  (\textcolor{black}{37.4\%} $\downarrow$)& 3.018& 10.5& 2.561 (\textcolor{blue}{15.1\%} $\downarrow$)& 8.1  (\textcolor{black}{23.4\%} $\downarrow$) \\
\midrule
\multirow{4}{*}{\rotatebox[origin=c]{90}{Exchange}}& 96& 0.124& 10.5& \pmb{0.116} (\textcolor{blue}{6.8\%} $\downarrow$)& 1.0  (\textcolor{black}{90.4\%} $\downarrow$)& 0.146& 17.9& 0.155 (\textcolor{black}{5.7\%} $\uparrow$)& 6.7  (\textcolor{black}{62.7\%} $\downarrow$)& 0.147& 10.5& 0.148 (\textcolor{black}{0.4\%} $\uparrow$)& 5.9  (\textcolor{black}{44.2\%} $\downarrow$)& 0.941& 11.3& 0.945 (\textcolor{black}{0.5\%} $\uparrow$)& 7.7  (\textcolor{black}{32.1\%} $\downarrow$)& 0.679& 10.5& 0.651 (\textcolor{blue}{4.2\%} $\downarrow$)& 7.2  (\textcolor{black}{31.6\%} $\downarrow$) \\
& 192& 0.243& 10.5& \pmb{0.233} (\textcolor{blue}{4.4\%} $\downarrow$)& 4.2  (\textcolor{black}{60.2\%} $\downarrow$)& 0.271& 17.9& 0.276 (\textcolor{black}{1.6\%} $\uparrow$)& 10.6  (\textcolor{black}{40.5\%} $\downarrow$)& 0.285& 10.5& 0.412 (\textcolor{black}{44.3\%} $\uparrow$)& 6.2  (\textcolor{black}{41.6\%} $\downarrow$)& 1.077& 11.3& 1.087 (\textcolor{black}{1.0\%} $\uparrow$)& 7.7  (\textcolor{black}{32.5\%} $\downarrow$)& 1.255& 10.5& 1.229 (\textcolor{blue}{2.0\%} $\downarrow$)& 7.0  (\textcolor{black}{33.5\%} $\downarrow$) \\
& 336& 0.462& 10.6& 0.459 (\textcolor{blue}{0.5\%} $\downarrow$)& 8.2  (\textcolor{black}{22.0\%} $\downarrow$)& \pmb{0.443}& 17.9& 0.447 (\textcolor{black}{0.9\%} $\uparrow$)& 11.9  (\textcolor{black}{33.2\%} $\downarrow$)& 0.772& 10.5& 0.695 (\textcolor{blue}{9.9\%} $\downarrow$)& 8.2  (\textcolor{black}{22.1\%} $\downarrow$)& 1.609& 11.3& 1.596 (\textcolor{blue}{0.8\%} $\downarrow$)& 9.2  (\textcolor{black}{19.1\%} $\downarrow$)& 1.565& 10.5& 1.588 (\textcolor{black}{1.5\%} $\uparrow$)& 6.3  (\textcolor{black}{40.5\%} $\downarrow$) \\
& 720& 1.336& 10.6& 1.156 (\textcolor{blue}{13.5\%} $\downarrow$)& 8.3  (\textcolor{black}{21.3\%} $\downarrow$)& 1.143& 17.9& 1.169 (\textcolor{black}{2.3\%} $\uparrow$)& 12.8  (\textcolor{black}{28.6\%} $\downarrow$)& \pmb{1.128}& 10.5& 1.206 (\textcolor{black}{6.9\%} $\uparrow$)& 8.1  (\textcolor{black}{23.0\%} $\downarrow$)& 2.747& 11.3& 2.497 (\textcolor{blue}{9.1\%} $\downarrow$)& 9.8  (\textcolor{black}{13.1\%} $\downarrow$)& 2.938& 10.5& 2.164 (\textcolor{blue}{26.3\%} $\downarrow$)& 5.9  (\textcolor{black}{44.4\%} $\downarrow$) \\
\midrule
\multirow{4}{*}{\rotatebox[origin=c]{90}{Illness}}& 24& \pmb{2.555}& 10.5& 2.641 (\textcolor{black}{3.3\%} $\uparrow$)& 6.9  (\textcolor{black}{35.0\%} $\downarrow$)& 3.285& 13.1& 3.590 (\textcolor{black}{9.3\%} $\uparrow$)& 7.6  (\textcolor{black}{42.1\%} $\downarrow$)& 3.686& 10.5& 3.648 (\textcolor{blue}{1.0\%} $\downarrow$)& 7.5  (\textcolor{black}{29.2\%} $\downarrow$)& 5.361& 11.3& 5.412 (\textcolor{black}{1.0\%} $\uparrow$)& 7.9  (\textcolor{black}{29.9\%} $\downarrow$)& 4.740& 10.5& 4.739 (\textcolor{blue}{0.0\%} $\downarrow$)& 6.9  (\textcolor{black}{34.3\%} $\downarrow$) \\
& 36& \pmb{1.913}& 10.5& 2.111 (\textcolor{black}{10.4\%} $\uparrow$)& 7.9  (\textcolor{black}{25.4\%} $\downarrow$)& 2.682& 13.5& 2.835 (\textcolor{black}{5.7\%} $\uparrow$)& 8.5  (\textcolor{black}{36.7\%} $\downarrow$)& 2.799& 10.5& 3.118 (\textcolor{black}{11.4\%} $\uparrow$)& 6.9  (\textcolor{black}{34.8\%} $\downarrow$)& 5.304& 11.3& 5.239 (\textcolor{blue}{1.2\%} $\downarrow$)& 6.6  (\textcolor{black}{41.7\%} $\downarrow$)& 4.776& 10.5& 4.911 (\textcolor{black}{2.8\%} $\uparrow$)& 8.1  (\textcolor{black}{23.2\%} $\downarrow$) \\
& 48& \pmb{1.873}& 10.5& 2.186 (\textcolor{black}{16.7\%} $\uparrow$)& 7.8  (\textcolor{black}{25.6\%} $\downarrow$)& 2.585& 13.9& 2.762 (\textcolor{black}{6.9\%} $\uparrow$)& 8.8  (\textcolor{black}{36.4\%} $\downarrow$)& 2.990& 10.5& 3.017 (\textcolor{black}{0.9\%} $\uparrow$)& 5.3  (\textcolor{black}{49.3\%} $\downarrow$)& 5.187& 11.3& 5.064 (\textcolor{blue}{2.4\%} $\downarrow$)& 7.5  (\textcolor{black}{34.0\%} $\downarrow$)& 5.090& 10.5& 4.914 (\textcolor{blue}{3.5\%} $\downarrow$)& 7.7  (\textcolor{black}{26.5\%} $\downarrow$) \\
& 60& \pmb{2.209}& 10.5& 2.391 (\textcolor{black}{8.2\%} $\uparrow$)& 7.0  (\textcolor{black}{34.0\%} $\downarrow$)& 2.807& 14.3& 2.994 (\textcolor{black}{6.7\%} $\uparrow$)& 8.2  (\textcolor{black}{42.7\%} $\downarrow$)& 2.860& 10.5& 2.984 (\textcolor{black}{4.3\%} $\uparrow$)& 7.0  (\textcolor{black}{33.2\%} $\downarrow$)& 5.234& 11.3& 5.197 (\textcolor{blue}{0.7\%} $\downarrow$)& 7.5  (\textcolor{black}{34.1\%} $\downarrow$)& 5.162& 10.5& 5.070 (\textcolor{blue}{1.8\%} $\downarrow$)& 7.9  (\textcolor{black}{25.0\%} $\downarrow$) \\
\midrule
\multirow{4}{*}{\rotatebox[origin=c]{90}{Traffic}}& 96& 0.608& 14.1& 0.641 (\textcolor{black}{5.4\%} $\uparrow$)& 4.3  (\textcolor{black}{69.6\%} $\downarrow$)& \pmb{0.581}& 22.3& 0.585 (\textcolor{black}{0.8\%} $\uparrow$)& 5.4  (\textcolor{black}{75.6\%} $\downarrow$)& 0.649& 14.9& 0.629 (\textcolor{blue}{3.1\%} $\downarrow$)& 5.2  (\textcolor{black}{65.2\%} $\downarrow$)& 0.724& 14.4& 0.744 (\textcolor{black}{2.7\%} $\uparrow$)& 1.5  (\textcolor{black}{89.3\%} $\downarrow$)& 0.646& 13.6& 0.677 (\textcolor{black}{4.8\%} $\uparrow$)& 3.7  (\textcolor{black}{72.8\%} $\downarrow$) \\
& 192& 0.623& 14.1& 0.660 (\textcolor{black}{5.9\%} $\uparrow$)& 4.3  (\textcolor{black}{69.4\%} $\downarrow$)& \pmb{0.605}& 22.3& 0.608 (\textcolor{black}{0.5\%} $\uparrow$)& 3.9  (\textcolor{black}{82.4\%} $\downarrow$)& 0.630& 14.9& 0.653 (\textcolor{black}{3.6\%} $\uparrow$)& 5.6  (\textcolor{black}{62.4\%} $\downarrow$)& 0.746& 14.4& 0.771 (\textcolor{black}{3.3\%} $\uparrow$)& 1.6  (\textcolor{black}{88.9\%} $\downarrow$)& 0.660& 13.6& 0.685 (\textcolor{black}{3.9\%} $\uparrow$)& 4.5  (\textcolor{black}{66.7\%} $\downarrow$) \\
& 336& 0.640& 14.6& 0.672 (\textcolor{black}{5.0\%} $\uparrow$)& 3.1  (\textcolor{black}{78.7\%} $\downarrow$)& 0.618& 22.3& 0.625 (\textcolor{black}{1.1\%} $\uparrow$)& 5.6  (\textcolor{black}{75.0\%} $\downarrow$)& \pmb{0.614}& 14.9& 0.649 (\textcolor{black}{5.7\%} $\uparrow$)& 3.5  (\textcolor{black}{76.8\%} $\downarrow$)& 0.830& 14.4& 0.907 (\textcolor{black}{9.3\%} $\uparrow$)& 1.7  (\textcolor{black}{88.1\%} $\downarrow$)& 0.668& 13.6& 0.687 (\textcolor{black}{2.8\%} $\uparrow$)& 3.2  (\textcolor{black}{76.5\%} $\downarrow$) \\
& 720& 0.658& 14.1& 0.687 (\textcolor{black}{4.5\%} $\uparrow$)& 5.3  (\textcolor{black}{62.4\%} $\downarrow$)& \pmb{0.634}& 22.3& 0.643 (\textcolor{black}{1.5\%} $\uparrow$)& 7.7  (\textcolor{black}{65.5\%} $\downarrow$)& 0.663& 14.9& 0.673 (\textcolor{black}{1.5\%} $\uparrow$)& 3.9  (\textcolor{black}{73.7\%} $\downarrow$)& 0.943& 14.4& 1.324 (\textcolor{black}{40.3\%} $\uparrow$)& 4.4  (\textcolor{black}{69.2\%} $\downarrow$)& 0.693& 13.6& 0.705 (\textcolor{black}{1.7\%} $\uparrow$)& 4.0  (\textcolor{black}{70.9\%} $\downarrow$) \\
\midrule
\multirow{4}{*}{\rotatebox[origin=c]{90}{Weather}}& 96& 0.182& 10.8& \pmb{0.167} (\textcolor{blue}{8.0\%} $\downarrow$)& 1.0  (\textcolor{black}{90.3\%} $\downarrow$)& 0.232& 17.9& 0.215 (\textcolor{blue}{7.2\%} $\downarrow$)& 1.8  (\textcolor{black}{90.0\%} $\downarrow$)& 0.266& 10.6& 0.258 (\textcolor{blue}{2.9\%} $\downarrow$)& 1.1  (\textcolor{black}{90.0\%} $\downarrow$)& 0.375& 11.4& 0.349 (\textcolor{blue}{7.0\%} $\downarrow$)& 1.1  (\textcolor{black}{90.2\%} $\downarrow$)& 0.380& 10.6& 0.198 (\textcolor{blue}{47.8\%} $\downarrow$)& 1.0  (\textcolor{black}{90.1\%} $\downarrow$) \\
& 192& 0.247& 10.6& \pmb{0.220} (\textcolor{blue}{11.0\%} $\downarrow$)& 1.0  (\textcolor{black}{90.5\%} $\downarrow$)& 0.281& 17.9& 0.280 (\textcolor{blue}{0.4\%} $\downarrow$)& 1.8  (\textcolor{black}{90.0\%} $\downarrow$)& 0.304& 10.6& 0.306 (\textcolor{black}{0.7\%} $\uparrow$)& 1.0  (\textcolor{black}{90.4\%} $\downarrow$)& 0.512& 11.4& 0.469 (\textcolor{blue}{8.4\%} $\downarrow$)& 2.6  (\textcolor{black}{76.8\%} $\downarrow$)& 0.584& 10.6& 0.282 (\textcolor{blue}{51.6\%} $\downarrow$)& 1.0  (\textcolor{black}{90.1\%} $\downarrow$) \\
& 336& 0.329& 10.6& \pmb{0.293} (\textcolor{blue}{10.9\%} $\downarrow$)& 1.0  (\textcolor{black}{90.1\%} $\downarrow$)& 0.355& 17.9& 0.337 (\textcolor{blue}{5.1\%} $\downarrow$)& 1.8  (\textcolor{black}{90.0\%} $\downarrow$)& 0.360& 10.6& 0.375 (\textcolor{black}{4.1\%} $\uparrow$)& 1.0  (\textcolor{black}{90.1\%} $\downarrow$)& 0.604& 11.4& 0.560 (\textcolor{blue}{7.3\%} $\downarrow$)& 4.2  (\textcolor{black}{63.0\%} $\downarrow$)& 0.684& 10.6& 0.343 (\textcolor{blue}{49.9\%} $\downarrow$)& 1.0  (\textcolor{black}{90.1\%} $\downarrow$) \\
& 720& 0.410& 10.6& \pmb{0.368} (\textcolor{blue}{10.1\%} $\downarrow$)& 1.0  (\textcolor{black}{90.2\%} $\downarrow$)& 0.410& 17.9& 0.410 (\textcolor{black}{0.0\%} $\uparrow$)& 1.8  (\textcolor{black}{90.1\%} $\downarrow$)& 0.423& 10.6& 0.416 (\textcolor{blue}{1.7\%} $\downarrow$)& 2.1  (\textcolor{black}{80.3\%} $\downarrow$)& 0.987& 11.4& 1.400 (\textcolor{black}{41.8\%} $\uparrow$)& 9.4  (\textcolor{black}{17.6\%} $\downarrow$)& 0.930& 10.6& 0.476 (\textcolor{blue}{48.9\%} $\downarrow$)& 1.0  (\textcolor{black}{90.4\%} $\downarrow$) \\

\bottomrule

\end{tabular}}

\end{table}

%% file: assets/tables/adapt_tuned_detailed_univariate.tex
\begin{table}[!h]

\caption{Univariate prediction comparison on the ETTm2 and Exchange datasets for prediction length $H \in \{96, 192, 336, 720\}$. For each model, the performance of the original (Dense) and the pruned model using PALS is presented in terms of MSE and parameter count. The difference between the results compared to the dense counterpart is shown in parenthesis as \% (\textcolor{blue}{blue} indicates improvement in the results compared to the dense model). \textbf{Bold} entries are the best performer in each experiment among various models. }\label{tab:results_all_univariate}
\centering
\resizebox{1\textwidth}{!}{
\begin{tabular}{@{\hskip 0.0in}c@{\hskip 0.03in}c@{\hskip 0.03in}|@{\hskip 0.03in}c@{\hskip 0.03in}c@{\hskip 0.03in}c@{\hskip 0.03in}c@{\hskip 0.03in}|@{\hskip 0.03in}c@{\hskip 0.03in}c@{\hskip 0.03in}c@{\hskip 0.03in}c@{\hskip 0.03in}|@{\hskip 0.03in}c@{\hskip 0.03in}@{\hskip 0.03in}c@{\hskip 0.03in}c@{\hskip 0.03in}c|c@{\hskip 0.03in}c@{\hskip 0.03in}c@{\hskip 0.03in}c@{\hskip 0.03in}|@{\hskip 0.03in}c@{\hskip 0.03in}c@{\hskip 0.03in}c@{\hskip 0.03in}c}
\toprule
\bt \multirow{3}{*}{\rotatebox[origin=c]{90}{Datasets}}& \bt \multirow{3}{*}{$H$} &\multicolumn{4}{c}{NSTransformer} &\multicolumn{4}{c}{FEDformer} &\multicolumn{4}{c}{Autoformer}&\multicolumn{4}{c}{Informer} &\multicolumn{4}{c}{Transformer}\\

&&\multicolumn{2}{c}{Dense} &\multicolumn{2}{c}{+PALS}       &\multicolumn{2}{c}{Dense} &\multicolumn{2}{c}{+PALS}  &\multicolumn{2}{c}{Dense} &\multicolumn{2}{c}{+PALS}       &\multicolumn{2}{c}{Dense} &\multicolumn{2}{c}{+PALS} &\multicolumn{2}{c}{Dense} &\multicolumn{2}{c}{+PALS}\\

& &MSE &\#Param  &MSE &\#Param       &MSE &\#Param  &MSE &\#Param       &MSE &\#Param  &MSE &\#Param          &MSE &\#Param  &MSE &\#Param      &MSE &\#Param  &MSE &\#Param  \\
\midrule 

\multirow{4}{*}{\rotatebox[origin=c]{90}{ETTm2}}& 96& 0.074& 10.6& 0.068 (\textcolor{blue}{7.6\%} $\downarrow$)& 1.1  (\textcolor{black}{90.0\%} $\downarrow$)& 0.069& 17.8& \pmb{0.065} (\textcolor{blue}{5.9\%} $\downarrow$)& 1.7  (\textcolor{black}{90.2\%} $\downarrow$)& 0.125& 10.5& 0.127 (\textcolor{black}{1.7\%} $\uparrow$)& 2.5  (\textcolor{black}{76.6\%} $\downarrow$)& 0.092& 11.3& 0.092 (\textcolor{black}{0.2\%} $\uparrow$)& 2.7  (\textcolor{black}{76.5\%} $\downarrow$)& 0.079& 10.5& 0.070 (\textcolor{blue}{12.0\%} $\downarrow$)& 1.1  (\textcolor{black}{89.3\%} $\downarrow$) \\
& 192& 0.128& 10.7& 0.107 (\textcolor{blue}{16.3\%} $\downarrow$)& 1.1  (\textcolor{black}{90.2\%} $\downarrow$)& \pmb{0.100}& 17.8& 0.101 (\textcolor{black}{0.6\%} $\uparrow$)& 1.7  (\textcolor{black}{90.2\%} $\downarrow$)& 0.141& 10.5& 0.144 (\textcolor{black}{2.2\%} $\uparrow$)& 1.0  (\textcolor{black}{90.4\%} $\downarrow$)& 0.137& 11.3& 0.129 (\textcolor{blue}{6.0\%} $\downarrow$)& 1.1  (\textcolor{black}{90.1\%} $\downarrow$)& 0.119& 10.5& 0.117 (\textcolor{blue}{1.3\%} $\downarrow$)& 7.0  (\textcolor{black}{33.8\%} $\downarrow$) \\
& 336& 0.146& 10.5& 0.153 (\textcolor{black}{5.4\%} $\uparrow$)& 1.0  (\textcolor{black}{90.1\%} $\downarrow$)& 0.133& 17.8& \pmb{0.131} (\textcolor{blue}{1.6\%} $\downarrow$)& 2.6  (\textcolor{black}{85.7\%} $\downarrow$)& 0.146& 10.5& 0.143 (\textcolor{blue}{2.4\%} $\downarrow$)& 1.0  (\textcolor{black}{90.4\%} $\downarrow$)& 0.174& 11.3& 0.170 (\textcolor{blue}{2.5\%} $\downarrow$)& 5.2  (\textcolor{black}{54.3\%} $\downarrow$)& 0.171& 10.5& 0.137 (\textcolor{blue}{19.5\%} $\downarrow$)& 1.0  (\textcolor{black}{90.1\%} $\downarrow$) \\
& 720& 0.225& 10.5& 0.230 (\textcolor{black}{2.0\%} $\uparrow$)& 3.7  (\textcolor{black}{65.3\%} $\downarrow$)& 0.185& 17.8& 0.186 (\textcolor{black}{0.2\%} $\uparrow$)& 2.6  (\textcolor{black}{85.4\%} $\downarrow$)& 0.195& 10.5& \pmb{0.181} (\textcolor{blue}{7.3\%} $\downarrow$)& 1.0  (\textcolor{black}{90.3\%} $\downarrow$)& 0.211& 11.3& 0.213 (\textcolor{black}{1.0\%} $\uparrow$)& 6.3  (\textcolor{black}{44.6\%} $\downarrow$)& 0.192& 10.5& 0.189 (\textcolor{blue}{1.7\%} $\downarrow$)& 6.0  (\textcolor{black}{42.6\%} $\downarrow$) \\
\midrule
\multirow{4}{*}{\rotatebox[origin=c]{90}{Exchange}}& 96& 0.161& 10.5& 0.150 (\textcolor{blue}{6.8\%} $\downarrow$)& 8.4  (\textcolor{black}{20.0\%} $\downarrow$)& \pmb{0.122}& 17.8& 0.131 (\textcolor{black}{7.4\%} $\uparrow$)& 12.2  (\textcolor{black}{31.4\%} $\downarrow$)& 0.161& 10.5& 0.159 (\textcolor{blue}{1.2\%} $\downarrow$)& 8.2  (\textcolor{black}{21.8\%} $\downarrow$)& 0.374& 11.3& 0.316 (\textcolor{blue}{15.6\%} $\downarrow$)& 6.9  (\textcolor{black}{39.2\%} $\downarrow$)& 0.329& 10.5& 0.268 (\textcolor{blue}{18.4\%} $\downarrow$)& 8.6  (\textcolor{black}{18.5\%} $\downarrow$) \\
& 192& \pmb{0.222}& 10.5& 0.252 (\textcolor{black}{13.7\%} $\uparrow$)& 8.5  (\textcolor{black}{19.1\%} $\downarrow$)& 0.257& 17.8& 0.276 (\textcolor{black}{7.3\%} $\uparrow$)& 9.2  (\textcolor{black}{48.4\%} $\downarrow$)& 0.304& 10.5& 0.345 (\textcolor{black}{13.5\%} $\uparrow$)& 8.1  (\textcolor{black}{23.2\%} $\downarrow$)& 1.180& 11.3& 1.064 (\textcolor{blue}{9.8\%} $\downarrow$)& 8.4  (\textcolor{black}{25.3\%} $\downarrow$)& 1.549& 10.5& 1.339 (\textcolor{blue}{13.5\%} $\downarrow$)& 8.0  (\textcolor{black}{23.8\%} $\downarrow$) \\
& 336& 0.395& 10.5& \pmb{0.327} (\textcolor{blue}{17.3\%} $\downarrow$)& 8.6  (\textcolor{black}{18.3\%} $\downarrow$)& 0.499& 17.8& 0.525 (\textcolor{black}{5.2\%} $\uparrow$)& 13.8  (\textcolor{black}{22.9\%} $\downarrow$)& 0.669& 10.5& 0.616 (\textcolor{blue}{8.0\%} $\downarrow$)& 7.4  (\textcolor{black}{29.8\%} $\downarrow$)& 1.771& 11.3& 1.747 (\textcolor{blue}{1.4\%} $\downarrow$)& 9.2  (\textcolor{black}{18.3\%} $\downarrow$)& 2.822& 10.5& 2.115 (\textcolor{blue}{25.1\%} $\downarrow$)& 8.2  (\textcolor{black}{21.8\%} $\downarrow$) \\
& 720& 0.981& 10.5& \pmb{0.973} (\textcolor{blue}{0.8\%} $\downarrow$)& 8.2  (\textcolor{black}{22.1\%} $\downarrow$)& 1.258& 17.8& 1.295 (\textcolor{black}{2.9\%} $\uparrow$)& 13.7  (\textcolor{black}{23.0\%} $\downarrow$)& 1.284& 10.5& 1.311 (\textcolor{black}{2.1\%} $\uparrow$)& 7.0  (\textcolor{black}{33.7\%} $\downarrow$)& 1.497& 11.3& 1.764 (\textcolor{black}{17.9\%} $\uparrow$)& 9.1  (\textcolor{black}{19.9\%} $\downarrow$)& 2.091& 10.5& 2.226 (\textcolor{black}{6.4\%} $\uparrow$)& 8.4  (\textcolor{black}{20.5\%} $\downarrow$) \\

\bottomrule
\end{tabular}}

\end{table}

%% file: assets/tables/summary_comparison_models_univariate.tex
\begin{table}[!h]

\caption{Summary of the results on the ETTm2 and Exchange datasets in Table \ref{tab:results_all_univariate}. For each experiment on a transformer model and dataset, the average MSE, MAE, and number of parameters ($\times 10^{6}$) for various prediction lengths are reported before and after applying PALS. The difference between these results is shown in \% where the \textcolor{blue}{blue} color means improvement of PALS compared to the corresponding dense model.}\label{tab:summary_results|_univariate}
\centering
\resizebox{.4\textwidth}{!}{
\begin{tabular}{c|ccc|ccc}
\toprule
\bt Model   & \multicolumn{3}{c}{ETTm2-uni}  & \multicolumn{3}{c}{Exchange-uni} \\

& MSE &MAE &\#Params & MSE &MAE &\#Params\\
\midrule

NSTransformer& 0.143& 0.285& 10.6& 0.440& 0.485& 10.5 \\
+PALS& 0.140& 0.277& 1.7& \textbf{0.425}& \textbf{0.470}& 8.4 \\
Difference& \textcolor{blue}{2.4\% $\downarrow$}& \textcolor{blue}{2.6\% $\downarrow$}& \textcolor{blue}{83.9\% $\downarrow$}& \textcolor{blue}{3.3\% $\downarrow$}& \textcolor{blue}{3.1\% $\downarrow$}& \textcolor{blue}{19.8\% $\downarrow$} \\
\midrule

FEDformer& 0.122& 0.265& 17.8& 0.534& 0.520& 17.8 \\
+PALS& \textbf{0.121}& \textbf{0.262}& 2.2& 0.557& 0.531& 12.2 \\
Difference& \textcolor{blue}{1.1\% $\downarrow$}& \textcolor{blue}{1.0\% $\downarrow$}& \textcolor{blue}{87.9\% $\downarrow$}& 4.2\% $\uparrow$& 2.2\% $\uparrow$& \textcolor{blue}{31.5\% $\downarrow$} \\
\midrule

Autoformer& 0.152& 0.300& 10.5& 0.605& 0.561& 10.5 \\
+PALS& 0.149& 0.297& \textbf{1.4}& 0.608& 0.561& \textbf{7.7} \\
Difference& \textcolor{blue}{2.1\% $\downarrow$}& \textcolor{blue}{0.9\% $\downarrow$}& \textcolor{blue}{86.9\% $\downarrow$}& 0.5\% $\uparrow$& \textcolor{blue}{0.1\% $\downarrow$}& \textcolor{blue}{27.1\% $\downarrow$} \\
\midrule

Informer& 0.153& 0.304& 11.3& 1.206& 0.852& 11.3 \\
+PALS& 0.151& 0.303& 3.8& 1.223& 0.866& 8.4 \\
Difference& \textcolor{blue}{1.7\% $\downarrow$}& \textcolor{blue}{0.4\% $\downarrow$}& \textcolor{blue}{66.4\% $\downarrow$}& 1.4\% $\uparrow$& 1.6\% $\uparrow$& \textcolor{blue}{25.7\% $\downarrow$} \\
\midrule

Transformer& 0.140& 0.286& 10.5& 1.698& 0.928& 10.5 \\
+PALS& 0.128& 0.274& 3.8& 1.487& 0.887& 8.3 \\
Difference& \textcolor{blue}{8.5\% $\downarrow$}& \textcolor{blue}{4.2\% $\downarrow$}& \textcolor{blue}{63.9\% $\downarrow$}& \textcolor{blue}{12.4\% $\downarrow$}& \textcolor{blue}{4.4\% $\downarrow$}& \textcolor{blue}{21.2\% $\downarrow$} \\

\bottomrule
\end{tabular}}

\end{table}

%% file: assets/tables/hyperparameter_sensitivity.tex
\begin{table}[!h]
\begin{scriptsize}
\caption{Hyperparameter Sensitivity of PALS. The prediction MSE and model's parameter count ($\times 10^{6}$) when applying PALS on NSTransformers is measured when changing the values of $\gamma$ and $\lambda$ in $\{1.05, 1.1, 1.2\}$. \textcolor{blue}{Blue} indicates improvement in the results compared to the dense model. \textbf{Bold} entries are the best performer in each row.}\label{tab:hyperparameters}
\begin{center}
\scalebox{0.8}{
\begin{tabular}{r@{\hskip 0.08in}|@{\hskip 0.08in}c|c@{\hskip 0.08in}|@{\hskip 0.08in}c@{\hskip 0.08in}c@{\hskip 0.08in}c@{\hskip 0.08in}|@{\hskip 0.08in}c@{\hskip 0.08in}c@{\hskip 0.08in}c@{\hskip 0.08in}|@{\hskip 0.08in}c@{\hskip 0.08in}c@{\hskip 0.08in}c@{\hskip 0.08in}}
\toprule
& & &\multicolumn{3}{c}{ \bt  $\gamma = 1.05$} & \multicolumn{3}{c}{ \bt $\gamma = 1.1$} & \multicolumn{3}{c}{ \bt  $\gamma = 1.2$}  \\
&\bt $H$ & Dense&\bt $\lambda=1.05$ &\bt$\lambda=1.1$ &\bt$\lambda=1.2$ &\bt $\lambda=1.05$ &\bt$\lambda=1.1$ &\bt$\lambda=1.2$   &\bt $\lambda=1.05$ &\bt$\lambda=1.1$ &\bt$\lambda=1.2$  \\ \midrule

\multirow{4}{*}{\rotatebox[origin=c]{90}{Electricity}}& 96& \pmb{0.170}& 0.195 (78.2\%)& 0.193 (74.8\%)& 0.193 (74.8\%)& 0.200 (90.3\%)& 0.198 (90.1\%)& 0.199 (90.1\%)& 0.206 (90.6\%)& 0.206 (90.5\%)& 0.206 (90.2\%) \\
& 192& \pmb{0.186}& 0.204 (84.3\%)& 0.211 (78.0\%)& 0.209 (78.6\%)& 0.215 (90.1\%)& 0.214 (90.1\%)& 0.214 (90.1\%)& 0.258 (90.3\%)& 0.260 (90.6\%)& 0.260 (90.3\%) \\
& 336& \pmb{0.198}& 0.215 (90.0\%)& 0.218 (90.0\%)& 0.219 (90.0\%)& 0.243 (90.1\%)& 0.246 (90.1\%)& 0.244 (90.1\%)& 0.271 (90.6\%)& 0.271 (90.4\%)& 0.270 (90.2\%) \\
& 720& \pmb{0.222}& 0.240 (78.1\%)& 0.242 (73.5\%)& 0.242 (73.5\%)& 0.250 (90.1\%)& 0.251 (90.1\%)& 0.252 (90.0\%)& 0.297 (90.4\%)& 0.295 (90.3\%)& 0.300 (90.1\%) \\
\midrule
\multirow{4}{*}{\rotatebox[origin=c]{90}{ETTm2}}& 96& 0.241& 0.272 (25.7\%)& 0.263 (30.3\%)& 0.268 (64.6\%)& 0.244 (40.0\%)& 0.248 (61.0\%)& \textcolor{blue}{0.202} (90.4\%)& \textcolor{blue}{\pmb{0.193}} (90.3\%)& \textcolor{blue}{0.193} (90.3\%)& \textcolor{blue}{0.195} (90.1\%) \\
& 192& \pmb{0.387}& 0.644 (21.4\%)& 0.626 (21.5\%)& 0.650 (25.9\%)& 0.574 (27.4\%)& 0.547 (41.1\%)& 0.498 (54.5\%)& 0.541 (40.4\%)& 0.393 (65.4\%)& 0.484 (65.3\%) \\
& 336& 0.615& \textcolor{blue}{0.529} (20.5\%)& \textcolor{blue}{0.529} (21.2\%)& \textcolor{blue}{0.480} (27.9\%)& 0.753 (21.9\%)& 0.725 (35.7\%)& \textcolor{blue}{0.563} (67.5\%)& \textcolor{blue}{0.473} (73.7\%)& \textcolor{blue}{\pmb{0.340}} (90.2\%)& \textcolor{blue}{0.340} (90.1\%) \\
& 720& 0.734& 0.751 (19.7\%)& 0.780 (27.7\%)& 0.793 (42.1\%)& 0.777 (30.9\%)& 0.804 (34.9\%)& \textcolor{blue}{0.641} (78.5\%)& \textcolor{blue}{0.678} (45.5\%)& \textcolor{blue}{0.600} (79.2\%)& \textcolor{blue}{\pmb{0.441}} (90.2\%) \\
\midrule
\multirow{4}{*}{\rotatebox[origin=c]{90}{Exchange}}& 96& 0.124& 0.139 (21.1\%)& 0.138 (22.3\%)& 0.135 (32.5\%)& 0.139 (22.3\%)& 0.143 (22.5\%)& 0.139 (64.2\%)& 0.137 (28.9\%)& \textcolor{blue}{\pmb{0.113}} (79.5\%)& 0.125 (87.0\%) \\
& 192& 0.243& 0.269 (16.3\%)& 0.269 (16.3\%)& 0.269 (16.3\%)& 0.266 (22.9\%)& 0.264 (27.7\%)& 0.263 (30.1\%)& 0.249 (34.6\%)& \textcolor{blue}{0.241} (38.1\%)& \textcolor{blue}{\pmb{0.239}} (60.9\%) \\
& 336& 0.462& 0.486 (17.7\%)& 0.486 (17.7\%)& 0.486 (17.7\%)& \textcolor{blue}{0.459} (22.0\%)& \textcolor{blue}{0.459} (22.0\%)& \textcolor{blue}{0.459} (22.0\%)& \textcolor{blue}{\pmb{0.429}} (13.1\%)& 0.478 (18.3\%)& \textcolor{blue}{0.460} (28.7\%) \\
& 720& 1.336& \textcolor{blue}{1.263} (17.0\%)& \textcolor{blue}{1.263} (17.0\%)& \textcolor{blue}{1.263} (17.0\%)& \textcolor{blue}{\pmb{1.134}} (22.9\%)& \textcolor{blue}{\pmb{1.134}} (22.9\%)& \textcolor{blue}{\pmb{1.134}} (22.9\%)& \textcolor{blue}{1.236} (15.6\%)& \textcolor{blue}{1.236} (15.6\%)& \textcolor{blue}{1.236} (15.6\%) \\
\midrule
\multirow{4}{*}{\rotatebox[origin=c]{90}{Illness}}& 24& \pmb{2.555}& 2.593 (27.5\%)& 2.638 (34.3\%)& 2.603 (31.6\%)& 2.668 (44.5\%)& 2.620 (47.9\%)& 2.613 (53.0\%)& 2.849 (65.6\%)& 2.784 (70.9\%)& 2.884 (75.7\%) \\
& 36& \pmb{1.913}& 2.112 (24.8\%)& 2.118 (27.5\%)& 2.125 (30.8\%)& 2.193 (34.3\%)& 2.220 (41.3\%)& 2.246 (51.6\%)& 2.433 (53.0\%)& 2.496 (59.7\%)& 2.644 (70.9\%) \\
& 48& \pmb{1.873}& 2.186 (25.6\%)& 2.207 (31.3\%)& 2.214 (34.0\%)& 2.246 (39.6\%)& 2.277 (48.8\%)& 2.332 (53.1\%)& 2.492 (60.4\%)& 2.650 (65.4\%)& 2.632 (63.5\%) \\
& 60& \pmb{2.209}& 2.413 (30.7\%)& 2.425 (34.5\%)& 2.425 (34.5\%)& 2.523 (50.9\%)& 2.549 (56.4\%)& 2.577 (48.0\%)& 2.772 (57.8\%)& 2.895 (64.7\%)& 3.019 (58.6\%) \\
\midrule
\multirow{4}{*}{\rotatebox[origin=c]{90}{Traffic}}& 96& \pmb{0.608}& 0.641 (69.6\%)& 0.647 (68.9\%)& 0.647 (68.9\%)& 0.676 (70.6\%)& 0.666 (90.1\%)& 0.681 (80.3\%)& 0.671 (90.4\%)& 0.669 (90.2\%)& 0.674 (90.6\%) \\
& 192& \pmb{0.623}& 0.658 (75.7\%)& 0.661 (69.6\%)& 0.661 (59.9\%)& 0.688 (76.4\%)& 0.694 (80.2\%)& 0.720 (60.2\%)& 0.711 (90.1\%)& 0.697 (90.1\%)& 0.693 (90.4\%) \\
& 336& \pmb{0.640}& 0.672 (78.7\%)& 0.671 (76.7\%)& 0.675 (72.9\%)& 0.712 (76.9\%)& 0.688 (90.3\%)& 0.690 (90.3\%)& 0.715 (90.4\%)& 0.720 (90.1\%)& 0.718 (90.6\%) \\
& 720& \pmb{0.658}& 0.687 (62.4\%)& 0.687 (62.4\%)& 0.687 (62.4\%)& 0.730 (76.4\%)& 0.735 (68.8\%)& 0.735 (68.8\%)& 0.728 (90.3\%)& 0.721 (90.3\%)& 0.724 (90.9\%) \\
\midrule
\multirow{4}{*}{\rotatebox[origin=c]{90}{Weather}}& 96& 0.182& \textcolor{blue}{\pmb{0.166}} (78.2\%)& \textcolor{blue}{0.167} (80.8\%)& \textcolor{blue}{0.167} (80.8\%)& \textcolor{blue}{0.167} (90.3\%)& \textcolor{blue}{0.167} (90.5\%)& \textcolor{blue}{0.167} (90.5\%)& \textcolor{blue}{0.172} (90.2\%)& \textcolor{blue}{0.172} (90.1\%)& \textcolor{blue}{0.172} (90.1\%) \\
& 192& 0.247& \textcolor{blue}{0.225} (83.7\%)& \textcolor{blue}{0.224} (85.4\%)& \textcolor{blue}{0.224} (85.4\%)& \textcolor{blue}{0.222} (90.5\%)& \textcolor{blue}{0.222} (90.5\%)& \textcolor{blue}{0.222} (90.5\%)& \textcolor{blue}{\pmb{0.220}} (90.1\%)& \textcolor{blue}{\pmb{0.220}} (90.1\%)& \textcolor{blue}{\pmb{0.220}} (90.1\%) \\
& 336& 0.329& \textcolor{blue}{0.303} (80.0\%)& \textcolor{blue}{0.301} (80.4\%)& \textcolor{blue}{0.301} (80.4\%)& \textcolor{blue}{\pmb{0.294}} (90.3\%)& \textcolor{blue}{0.295} (90.5\%)& \textcolor{blue}{0.295} (90.5\%)& \textcolor{blue}{0.297} (90.2\%)& \textcolor{blue}{0.298} (90.1\%)& \textcolor{blue}{0.298} (90.1\%) \\
& 720& 0.410& 0.413 (68.9\%)& \textcolor{blue}{0.392} (85.2\%)& \textcolor{blue}{0.392} (85.2\%)& \textcolor{blue}{0.372} (90.3\%)& \textcolor{blue}{0.382} (90.5\%)& \textcolor{blue}{0.382} (90.5\%)& \textcolor{blue}{\pmb{0.364}} (90.3\%)& \textcolor{blue}{0.375} (90.1\%)& \textcolor{blue}{0.375} (90.1\%) \\

\bottomrule
\end{tabular}}
\end{center}
\end{scriptsize}
\end{table}

%% file: Supplementary/7-app-efficiency.tex
In this appendix, we discuss the efficiency of PALS in terms of computational costs from various perspectives. 

\subsection{Training FLOPs}\label{app:train_flops}
In this section, we present a comprehensive analysis of the training computational efficiency, as outlined in Table \ref{tab:summary_results_train_flops}. By looking into the inference FLOPs in Table \ref{tab:summary_results} in the paper, we can observe that the efficiency gain during training is even higher than inference. This observation underscores the significance of optimizing computational resources for training, a critical aspect in the deployment of machine learning algorithms.

\input{assets/tables/summary_comparison_models_train_FLOPs}

\subsection{Pruning Capabilities}
Based on the observations in Section \ref{ssec:comp_pruning}, PALS achieves the highest average sparsity level among the considered pruning and sparse training methods. More importantly, it finds the optimal sparsity level automatically without requiring any prior information, while most pruning and sparse training algorithms need to receive the sparsity level as an input of the algorithm. In short, PALS can find a network with higher sparsity than the competitors (GMP, GraNet, and RigL), where the sparsity level is found automatically.

\subsection{Convergence Speed}\label{app:convergence_speed}
Another factor that we consider regarding the efficiency of the methods is the convergence speed. If a method converges faster than the others, it can be considered to be more efficient in terms of resource usage.

To compare the convergence speed of each model, we compare the training epochs. As we use early stopping during training, each training round might not need the full training epochs (which is set to 10). In Table \ref{tab:convergence}, we report the average number of training epochs for various datasets. The results are an average for different prediction lengths and three random seeds. While GraNet needs almost full training time due to using a fixed pruning schedule which is determined based on the number of epochs, PALS can automate the pruning speed based on the loss. On four out of six datasets, PALS converges faster than GraNet and the dense model. On the Weather dataset, PALS requires longer training time than the dense model. As will be explained in Section \ref{app-ssec:sparsity_during_training}, PALS performs most part of this training at a very high sparsity level ($\sim 90\%$), thus being resource-efficient. Overall, PALS is able to achieve a higher or comparable speed to the dense model in most cases considered.

\begin{table}[!h]
    \caption{Comparison of convergence speed in terms of the number of training epochs on the NSTransformer model. The results are average over various prediction lengths of $H \in \{96, 192, 336, 720\}$ (except for the Illness dataset for which $H \in \{24, 36, 48, 60\}$ ).}\label{tab:convergence}
    \begin{center}
        \resizebox{0.35\textwidth}{!}{
        \begin{tabular}{c|ccc}
            \toprule
            Dataset & PALS & GraNet & Dense \\\midrule
            Electricity & \textbf{8.83} & 9.75 & 8.92 \\
            ETTm2 & \textbf{4.58} & 9.00 & 4.92 \\
            Exchange & 5.83 & 9.42 & \textbf{4.33} \\
            Illness & \textbf{7.97} & 9.58 & 8.92 \\
            Traffic & \textbf{8.83} & 9.50 & 9.5 \\
            Weather & 7.00 & 9.08 & \textbf{4.08}\\
            \bottomrule
        \end{tabular}}
    \end{center}
\end{table}

\subsection{Sparsity During Training}\label{app-ssec:sparsity_during_training}
Another factor affecting the efficiency of PALS is the sparsity during training. As in our experiment, PALS starts from a dense network, we analyze when it reaches the final sparsity and how the sparsity changes during training. Then, would be able to analyze whether the forward pass is mostly performed sparsely or not.

\begin{figure}[!h]
\centering  
\subfigure[NSTransformer]{\label{fig:sparsities_ns_Transformer}\includegraphics[width=0.49\columnwidth]{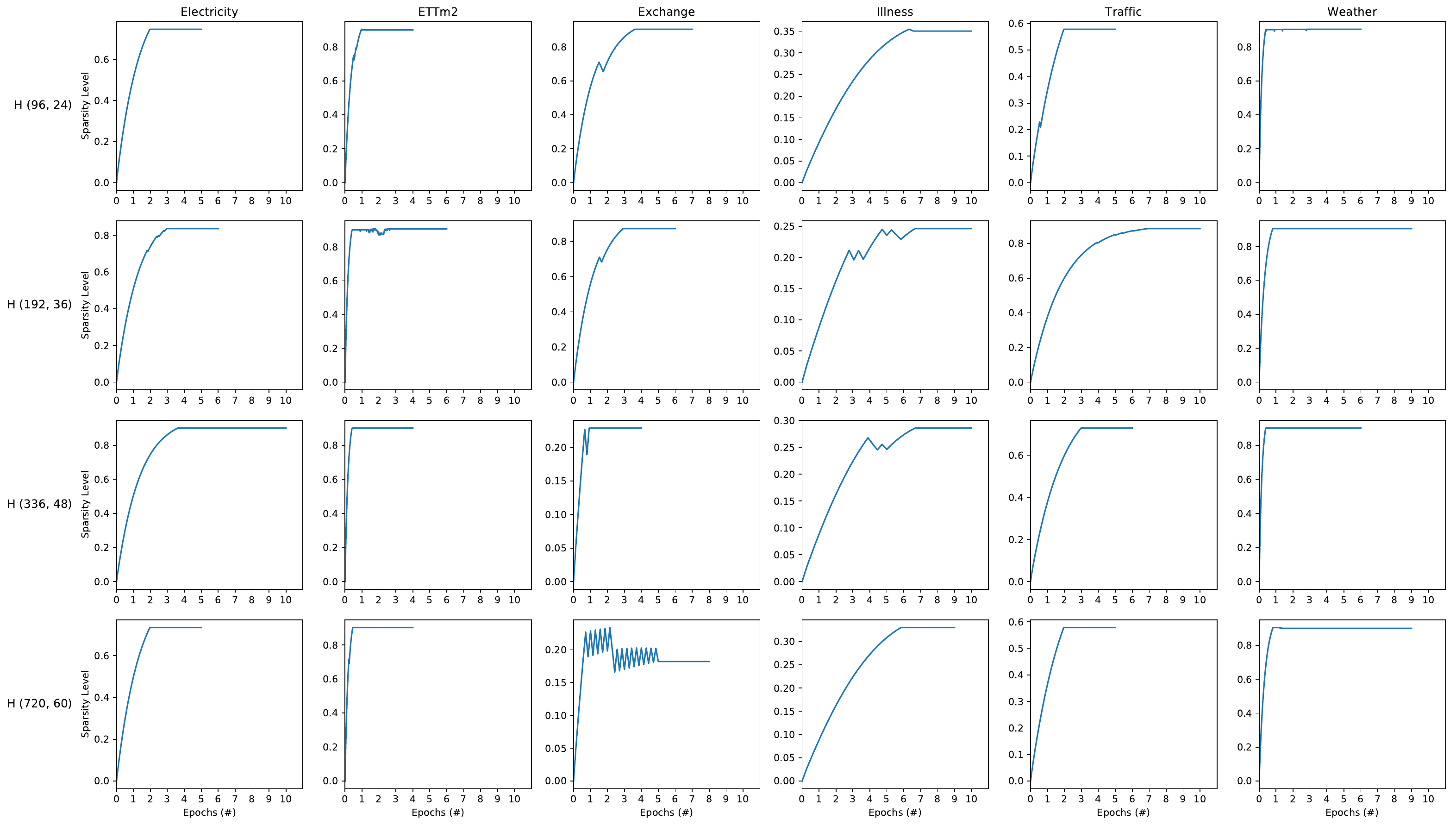}}
\subfigure[Autoformer]{\label{fig:sparsities_Autoformer}\includegraphics[width=0.49\columnwidth]{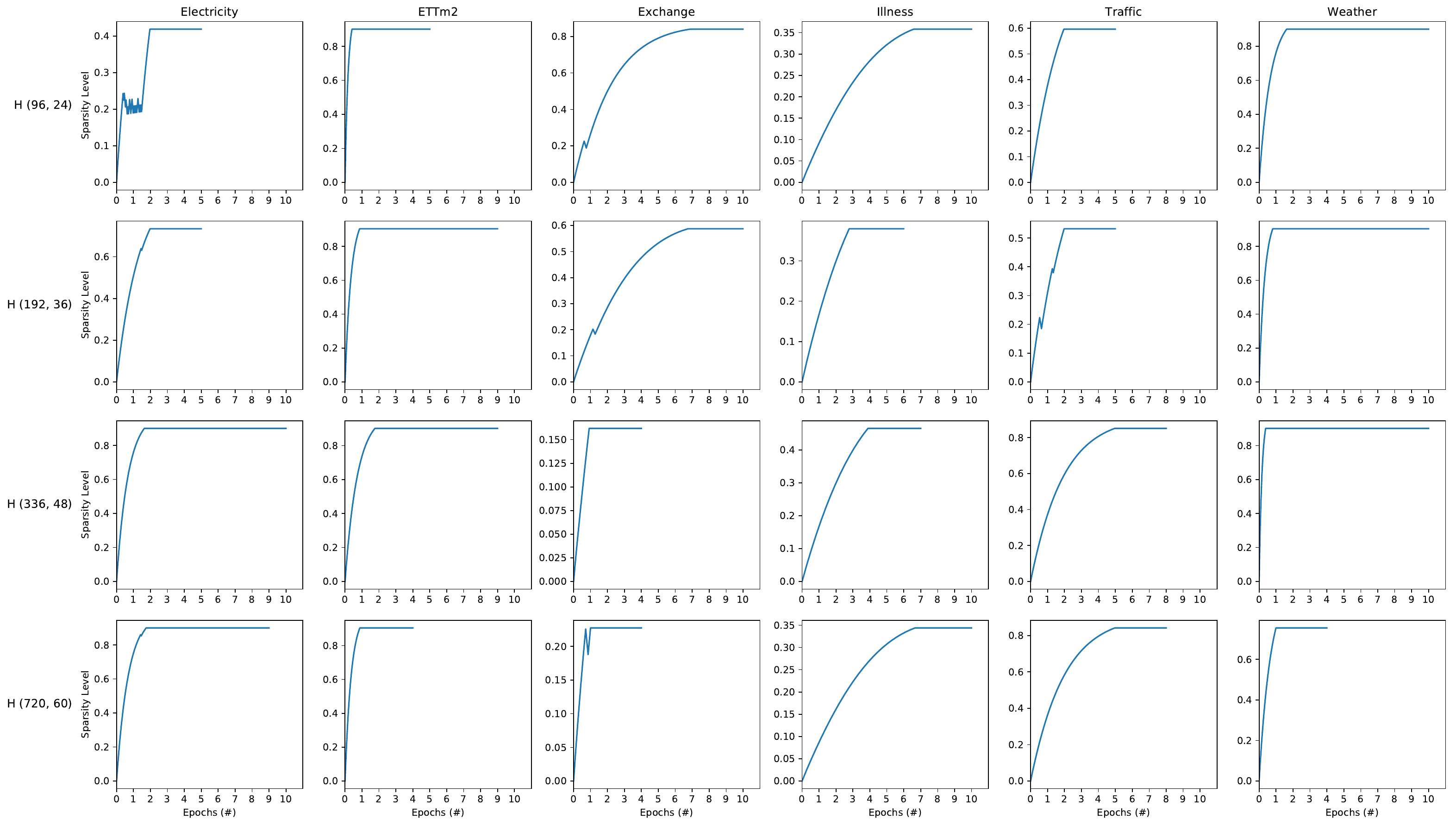}}
\subfigure[FEDformer]{\label{fig:sparsities_FEDformer}\includegraphics[width=0.49\columnwidth]{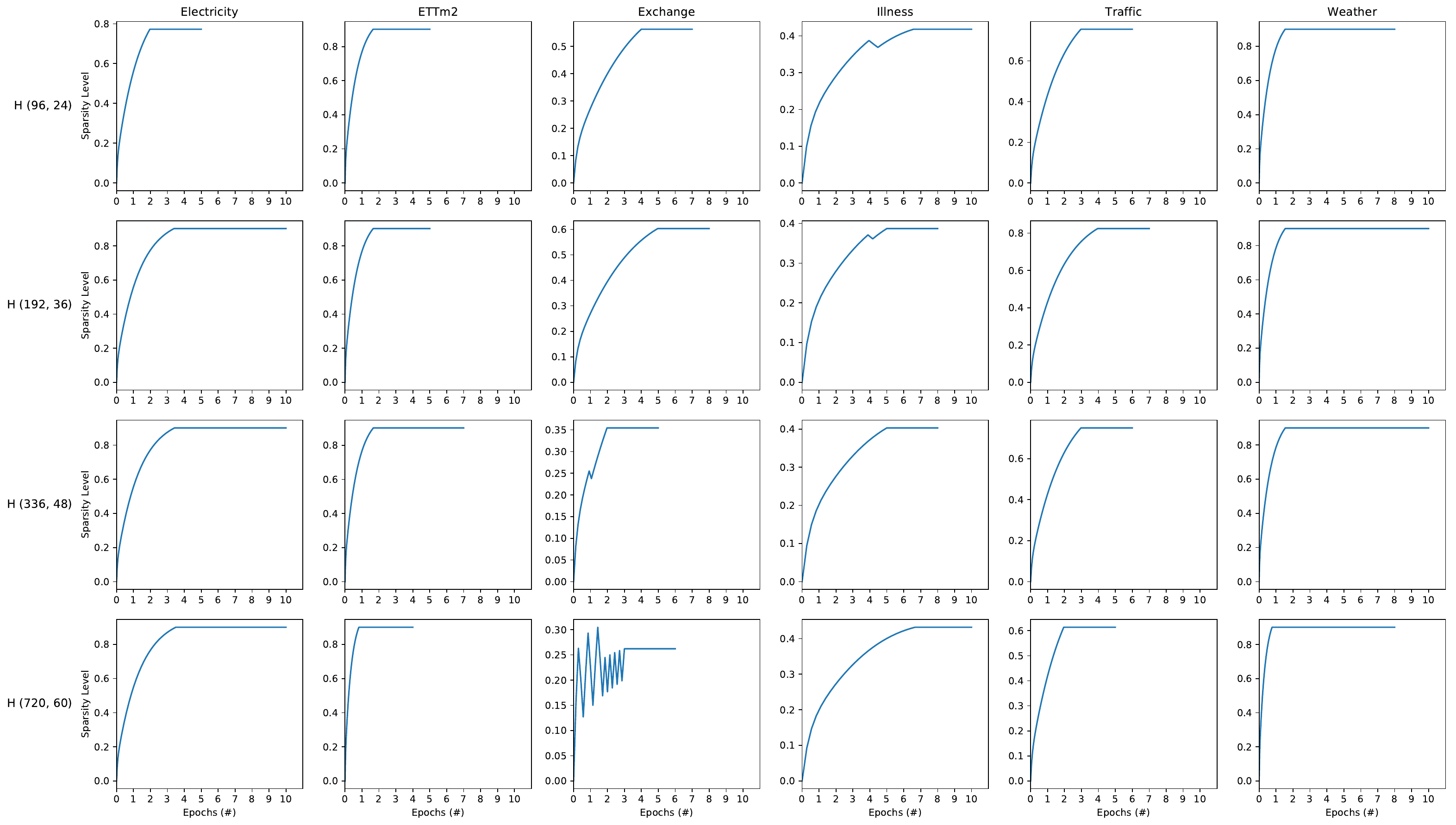}}
\subfigure[Informer]{\label{fig:sparsities_Informer}\includegraphics[width=0.49\columnwidth]{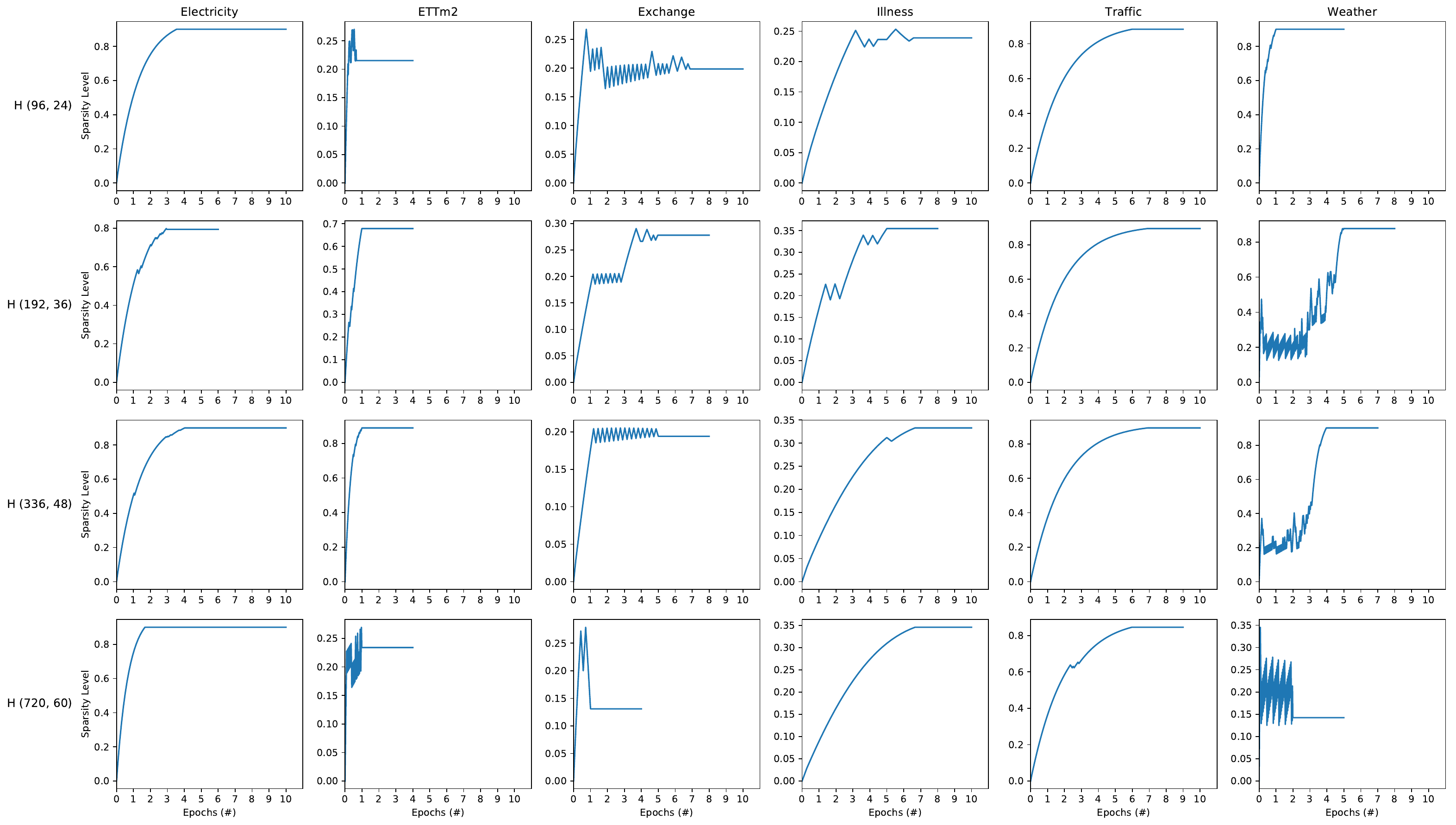}}
\subfigure[Transformer]{\label{fig:sparsities_Transformer}\includegraphics[width=0.49\columnwidth]{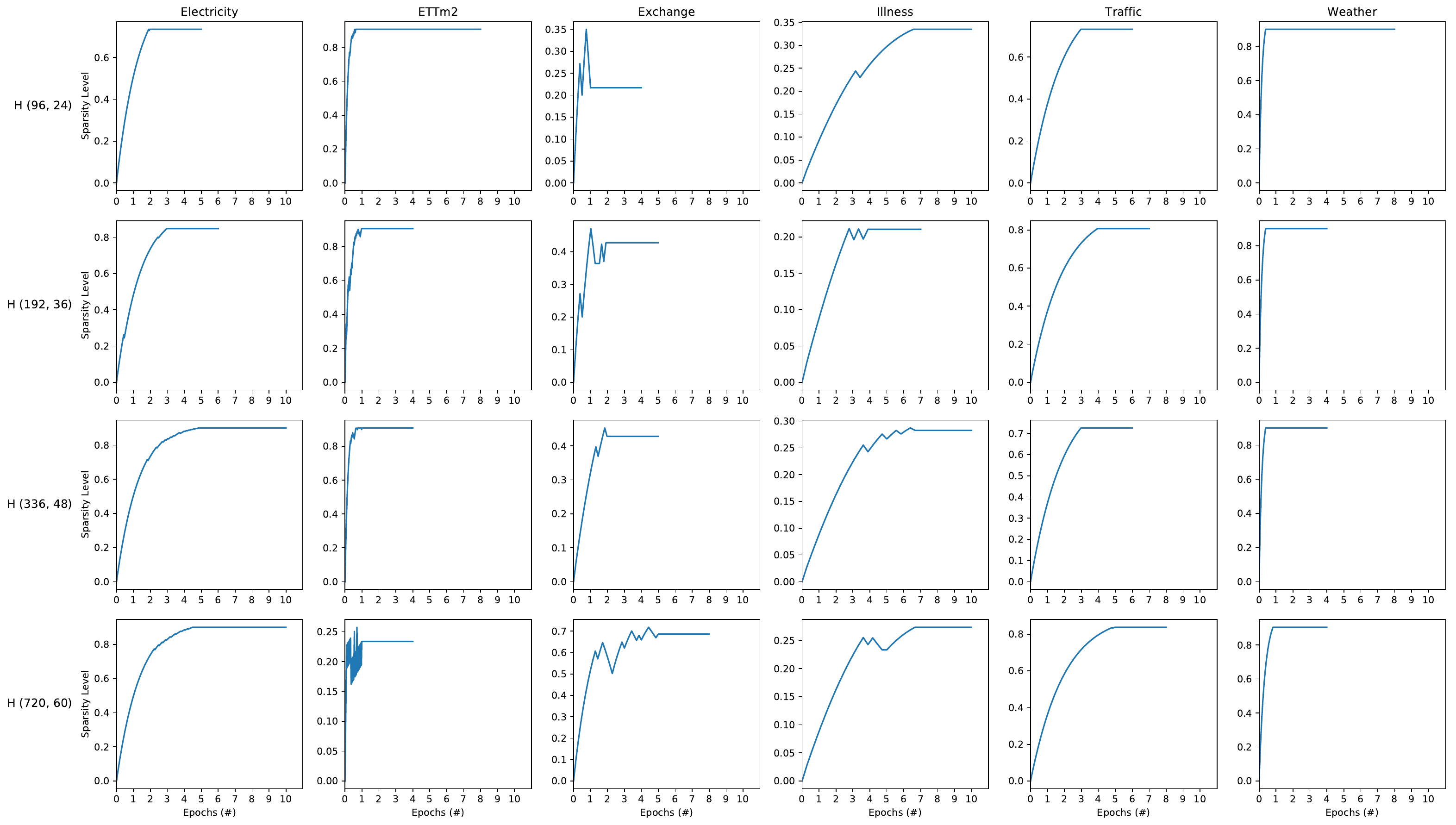}}
\caption{Sparsity level of each network during training of PALS. In most cases, the final sparsity is achieved within a few epochs after the training starts.  Therefore, the forward pass during training is performed sparsely for a large fraction of the training process.}
\label{fig:sparsities}
\end{figure}

To achieve this, we plot the sparsity level during the training of PALS for each transformer model and dataset for various prediction lengths. The sparsity levels are measured after each pruning iteration which is repeated every $5$ batches on the Illness dataset and every $20$ batches on the other datasets. These values are measured till the last pruning iteration before saving the model. The results for all models are presented in Figure \ref{fig:sparsities}.

In Figure \ref{fig:sparsities}, due to different convergence speeds on each dataset, each model requires a different number of training epochs. As the pruning is tuned based on the loss value, the pruning speed varies for each dataset and model. For example, on the Weather, Electricity, and Ettm2 datasets, for most of the considered transformer models and prediction lengths, the models reach the final sparsity level within a few training epochs. However, for the Exchange, Illness, and Traffic datasets, the convergence speed can be different among different models and prediction lengths. Also, in these datasets, the final sparsity is reached slower than the earlier category (Weather, Electricity, and Ettm2 datasets). However, in most cases considered, the final sparsity is reached only within half of the training period.

Overall, it can be observed that in most cases PALS reaches the final sparsity level in a few epochs. Therefore, the forward pass during training is performed sparsely for a large fraction of the training process.

\subsection{Sparse Implementation.}\label{app:ssec:sparse_implementation}
Another aspect regarding efficiency is concerned with the true sparse implementation. Further steps are required to achieve the full advantage of GPU speedup. However, this is a challenging research question that is out of the scope of the current work. Unstructured sparsity (e.g., pruning connections of a network as we use in this paper) still suffers from hardware support \cite{liu2023ten}, while proven to be more effective than structured sparsity (e.g., pruning neurons or other units) \cite{hoefler2021sparsity}. Although there have been some works that have implemented truly sparse neural networks efficiently on CPUs \cite{liu2021sparse,curci2021truly,atashgahi2022quick} and just for MLPs and Autoencoders, proper and efficient GPU support for sparse matrix computation is yet to be developed. In this paper, we focus on the algorithmic aspect of developing resource-efficient algorithms, while getting full advantage of sparsity is a large amount of research that we cannot address with our current human resources. We hope that works such as ours can light an awareness signal, bringing more researchers from the community together to start addressing seriously the “elephant in the room”.

%% file: assets/tables/summary_comparison_models_train_FLOPs.tex
\begin{table*}[!h]

\vspace{-3mm}
\caption{Summary of the results on the benchmark Datasets in Table \ref{tab:results_all}. For each experiment on a transformer model and dataset, the average MSE, MAE, number of parameters ($\times 10^{6}$), and the training FLOPs count ($\times 10^{12}$) for various prediction lengths are reported before and after applying PALS. The difference between these results is shown in \% where the \textcolor{blue}{blue} color means improvement of PALS compared to the corresponding dense model.}\label{tab:summary_results_train_flops}

\centering
\scalebox{0.4}{
\begin{tabular}{c| c@{\hskip 0.04in}c@{\hskip 0.04in}c@{\hskip 0.04in}c|c@{\hskip 0.04in}c@{\hskip 0.04in}c@{\hskip 0.04in}c|c@{\hskip 0.04in}c@{\hskip 0.04in}c@{\hskip 0.04in}c|c@{\hskip 0.04in}c@{\hskip 0.04in}c@{\hskip 0.04in}c|c@{\hskip 0.04in}c@{\hskip 0.04in}c@{\hskip 0.04in}c|c@{\hskip 0.04in}c@{\hskip 0.04in}c@{\hskip 0.04in}c@{\hskip 0.04in}c}

\toprule
\bt Model  &\multicolumn{4}{c}{Electricity} & \multicolumn{4}{c}{ETTm2}  & \multicolumn{4}{c}{Exchange} & \multicolumn{4}{c}{Illness}  & \multicolumn{4}{c}{Traffic}  & \multicolumn{4}{c}{Weather}\\

& MSE &MAE &\#Params &\#FLOPs & MSE &MAE &\#Params &\#FLOPs& MSE &MAE &\#Params &\#FLOPs& MSE &MAE  &\#Params &\#FLOPs & MSE &MAE &\#Params &\#FLOPs & MSE &MAE  &\#Params &\#FLOPs \\
\midrule

NSTransformer& \textbf{0.19}& \textbf{0.30}& 12.0& 898.95& 0.49& 0.43& 10.6& 854.73& 0.54& 0.49& 10.6& 113.93& \textbf{2.14}& \textbf{0.92}& 10.5& 6.21& 0.63& \textbf{0.34}& 14.2& 721.70& 0.29& 0.31& 10.7& 810.94 \\
+PALS& 0.21& 0.32& 2.2& 27.11& 0.38& 0.39& 2.5& 41.99& \textbf{0.49}& \textbf{0.47}& \textbf{5.4}& 51.50& 2.33& 0.97& 7.4& 1.44& 0.67& 0.37& 4.3& 35.29& \textbf{0.26}& \textbf{0.29}& 1.0& \textbf{21.24} \\
Difference& 10.8\% $\uparrow$& 7.3\% $\uparrow$& \textcolor{blue}{81.5\% $\downarrow$}& \textcolor{blue}{97.0\% $\downarrow$}& \textcolor{blue}{24.0\% $\downarrow$}& \textcolor{blue}{11.2\% $\downarrow$}& \textcolor{blue}{76.7\% $\downarrow$}& \textcolor{blue}{95.1\% $\downarrow$}& \textcolor{blue}{9.3\% $\downarrow$}& \textcolor{blue}{3.6\% $\downarrow$}& \textcolor{blue}{48.5\% $\downarrow$}& \textcolor{blue}{54.8\% $\downarrow$}& 9.1\% $\uparrow$& 5.1\% $\uparrow$& \textcolor{blue}{30.0\% $\downarrow$}& \textcolor{blue}{76.8\% $\downarrow$}& 5.2\% $\uparrow$& 9.1\% $\uparrow$& \textcolor{blue}{70.1\% $\downarrow$}& \textcolor{blue}{95.1\% $\downarrow$}& \textcolor{blue}{10.2\% $\downarrow$}& \textcolor{blue}{6.9\% $\downarrow$}& \textcolor{blue}{90.3\% $\downarrow$}& \textcolor{blue}{97.4\% $\downarrow$} \\
\midrule

FEDformer& 0.21& 0.32& 19.5& 965.68& \textbf{0.30}& 0.35& 17.9& 1094.02& 0.50& 0.49& 17.9& 151.60& 2.84& 1.14& 13.7& 6.64& \textbf{0.61}& 0.38& 22.3& 618.66& 0.32& 0.37& 17.9& 1027.26 \\
+PALS& 0.23& 0.34& 3.0& 21.09& 0.30& \textbf{0.35}& 1.8& 20.92& 0.51& 0.50& 10.5& \textbf{32.44}& 3.05& 1.19& 8.3& \textbf{0.99}& 0.62& 0.38& 5.6& 30.39& 0.31& 0.36& 1.8& 22.73 \\
Difference& 9.0\% $\uparrow$& 4.9\% $\uparrow$& \textcolor{blue}{84.7\% $\downarrow$}& \textcolor{blue}{97.8\% $\downarrow$}& 1.5\% $\uparrow$& \textcolor{blue}{0.5\% $\downarrow$}& \textcolor{blue}{90.2\% $\downarrow$}& \textcolor{blue}{98.1\% $\downarrow$}& 2.1\% $\uparrow$& 1.1\% $\uparrow$& \textcolor{blue}{41.2\% $\downarrow$}& \textcolor{blue}{78.6\% $\downarrow$}& 7.2\% $\uparrow$& 5.0\% $\uparrow$& \textcolor{blue}{39.5\% $\downarrow$}& \textcolor{blue}{85.1\% $\downarrow$}& 1.0\% $\uparrow$& 1.1\% $\uparrow$& \textcolor{blue}{74.6\% $\downarrow$}& \textcolor{blue}{95.1\% $\downarrow$}& \textcolor{blue}{2.8\% $\downarrow$}& \textcolor{blue}{3.2\% $\downarrow$}& \textcolor{blue}{90.0\% $\downarrow$}& \textcolor{blue}{97.8\% $\downarrow$} \\
\midrule

Autoformer& 0.24& 0.34& 12.1& 751.01& 0.33& 0.37& 10.5& 867.75& 0.58& 0.53& 10.5& 111.87& 3.08& 1.18& 10.5& 5.89& 0.64& 0.40& 14.9& 669.03& 0.34& 0.38& 10.6& 1175.29 \\
+PALS& 0.26& 0.36& 2.7& 27.92& 0.31& 0.35& \textbf{1.0}& \textbf{20.23}& 0.62& 0.55& 7.1& 40.54& 3.19& 1.22& \textbf{6.7}& 1.17& 0.65& 0.41& 4.5& 31.83& 0.34& 0.38& 1.3& 32.32 \\
Difference& 9.3\% $\uparrow$& 4.6\% $\uparrow$& \textcolor{blue}{77.7\% $\downarrow$}& \textcolor{blue}{96.3\% $\downarrow$}& \textcolor{blue}{8.1\% $\downarrow$}& \textcolor{blue}{5.6\% $\downarrow$}& \textcolor{blue}{90.3\% $\downarrow$}& \textcolor{blue}{97.7\% $\downarrow$}& 5.5\% $\uparrow$& 4.4\% $\uparrow$& \textcolor{blue}{32.7\% $\downarrow$}& \textcolor{blue}{63.8\% $\downarrow$}& 3.5\% $\uparrow$& 3.2\% $\uparrow$& \textcolor{blue}{36.6\% $\downarrow$}& \textcolor{blue}{80.2\% $\downarrow$}& 1.9\% $\uparrow$& 2.1\% $\uparrow$& \textcolor{blue}{69.5\% $\downarrow$}& \textcolor{blue}{95.2\% $\downarrow$}& 0.1\% $\uparrow$& \textcolor{blue}{1.3\% $\downarrow$}& \textcolor{blue}{87.7\% $\downarrow$}& \textcolor{blue}{97.3\% $\downarrow$} \\
\midrule

Informer& 0.36& 0.43& 12.5& 917.53& 1.53& 0.88& 11.3& 694.06& 1.59& 1.00& 11.3& 135.66& 5.27& 1.58& 11.3& 5.23& 0.81& 0.46& 14.4& 683.57& 0.62& 0.55& 11.4& 1135.86 \\
+PALS& 0.42& 0.48& \textbf{1.4}& \textbf{14.93}& 1.39& 0.83& 5.3& 187.17& 1.53& 0.98& 8.6& 59.27& 5.23& 1.57& 7.4& 1.15& 0.94& 0.53& \textbf{2.3}& \textbf{19.08}& 0.69& 0.56& 4.3& 187.78 \\
Difference& 18.9\% $\uparrow$& 11.3\% $\uparrow$& \textcolor{blue}{88.6\% $\downarrow$}& \textcolor{blue}{98.4\% $\downarrow$}& \textcolor{blue}{9.0\% $\downarrow$}& \textcolor{blue}{5.8\% $\downarrow$}& \textcolor{blue}{53.4\% $\downarrow$}& \textcolor{blue}{73.0\% $\downarrow$}& \textcolor{blue}{3.9\% $\downarrow$}& \textcolor{blue}{1.6\% $\downarrow$}& \textcolor{blue}{24.2\% $\downarrow$}& \textcolor{blue}{56.3\% $\downarrow$}& \textcolor{blue}{0.8\% $\downarrow$}& \textcolor{blue}{1.0\% $\downarrow$}& \textcolor{blue}{34.9\% $\downarrow$}& \textcolor{blue}{78.0\% $\downarrow$}& 15.5\% $\uparrow$& 15.3\% $\uparrow$& \textcolor{blue}{83.9\% $\downarrow$}& \textcolor{blue}{97.2\% $\downarrow$}& 12.1\% $\uparrow$& 2.0\% $\uparrow$& \textcolor{blue}{61.9\% $\downarrow$}& \textcolor{blue}{83.5\% $\downarrow$} \\
\midrule

Transformer& 0.28& 0.38& 11.7& 868.54& 1.48& 0.86& 10.5& 922.01& 1.61& 0.97& 10.5& 134.00& 4.94& 1.49& 10.5& 6.43& 0.67& 0.36& 13.6& 630.91& 0.64& 0.56& 10.6& 874.91 \\
+PALS& 0.31& 0.40& 2.5& 31.28& 1.08& 0.75& 3.2& 205.35& 1.41& 0.91& 6.6& 34.74& 4.91& 1.48& 7.7& 1.49& 0.69& 0.38& 3.8& 29.67& 0.32& 0.38& \textbf{1.0}& 21.53 \\
Difference& 10.5\% $\uparrow$& 5.7\% $\uparrow$& \textcolor{blue}{78.3\% $\downarrow$}& \textcolor{blue}{96.4\% $\downarrow$}& \textcolor{blue}{26.7\% $\downarrow$}& \textcolor{blue}{13.3\% $\downarrow$}& \textcolor{blue}{69.9\% $\downarrow$}& \textcolor{blue}{77.7\% $\downarrow$}& \textcolor{blue}{12.5\% $\downarrow$}& \textcolor{blue}{6.1\% $\downarrow$}& \textcolor{blue}{37.5\% $\downarrow$}& \textcolor{blue}{74.1\% $\downarrow$}& \textcolor{blue}{0.7\% $\downarrow$}& \textcolor{blue}{0.9\% $\downarrow$}& \textcolor{blue}{27.2\% $\downarrow$}& \textcolor{blue}{76.8\% $\downarrow$}& 3.3\% $\uparrow$& 5.4\% $\uparrow$& \textcolor{blue}{71.7\% $\downarrow$}& \textcolor{blue}{95.3\% $\downarrow$}& \textcolor{blue}{49.6\% $\downarrow$}& \textcolor{blue}{32.5\% $\downarrow$}& \textcolor{blue}{90.2\% $\downarrow$}& \textcolor{blue}{97.5\% $\downarrow$} \\

\bottomrule

\end{tabular}}
\vspace{-4mm}
\end{table*}

%% file: Supplementary/7-app_prediction.tex
\begin{figure}[!h]
\centering  
\subfigure[\scriptsize{NStransformer}]{\label{fig:Weather_NStransformer}\includegraphics[width=23mm]{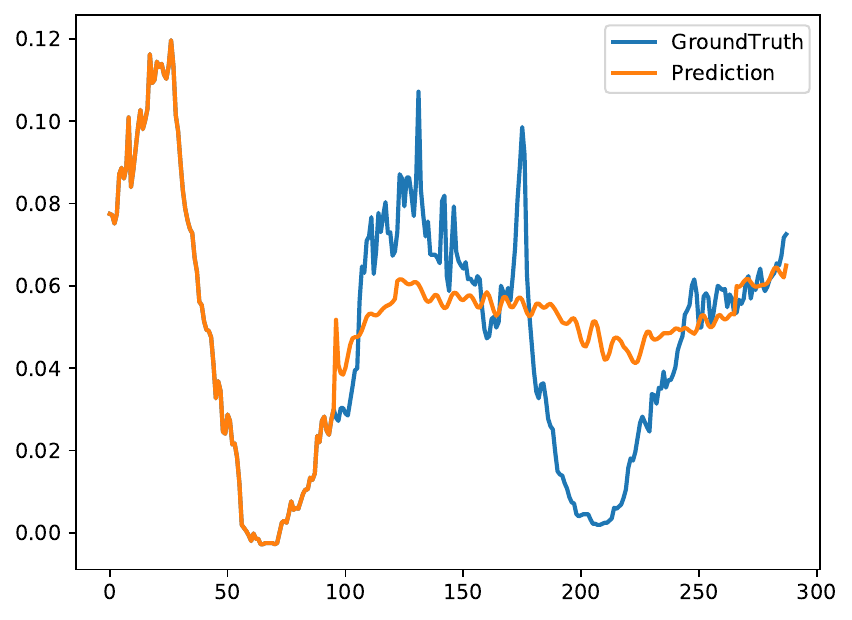}}
\subfigure[\scriptsize{Fedformer}]{\label{fig:Weather_Fedformer}\includegraphics[width=23mm]{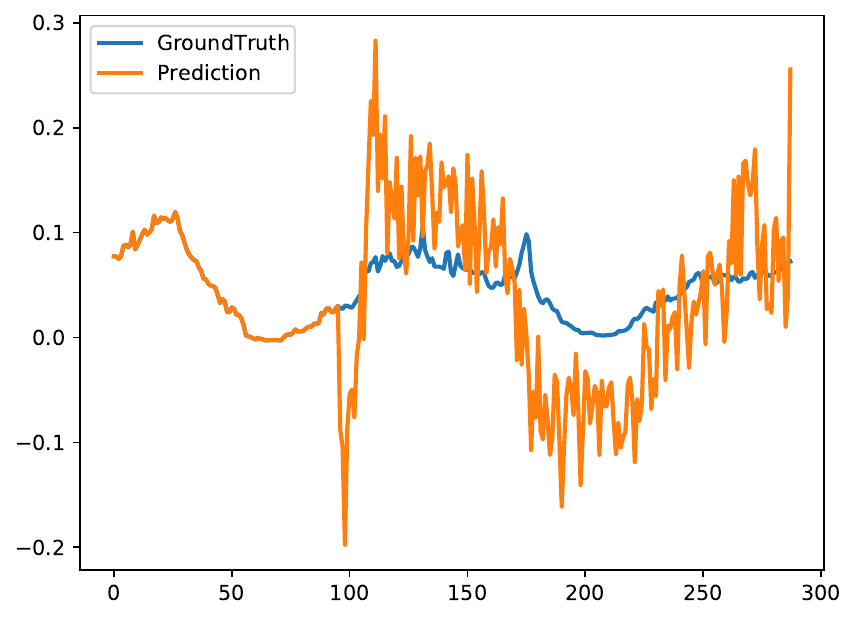}}
\subfigure[\scriptsize{Autoformer}]{\label{fig:Weather_Autoformer}\includegraphics[width=23mm]{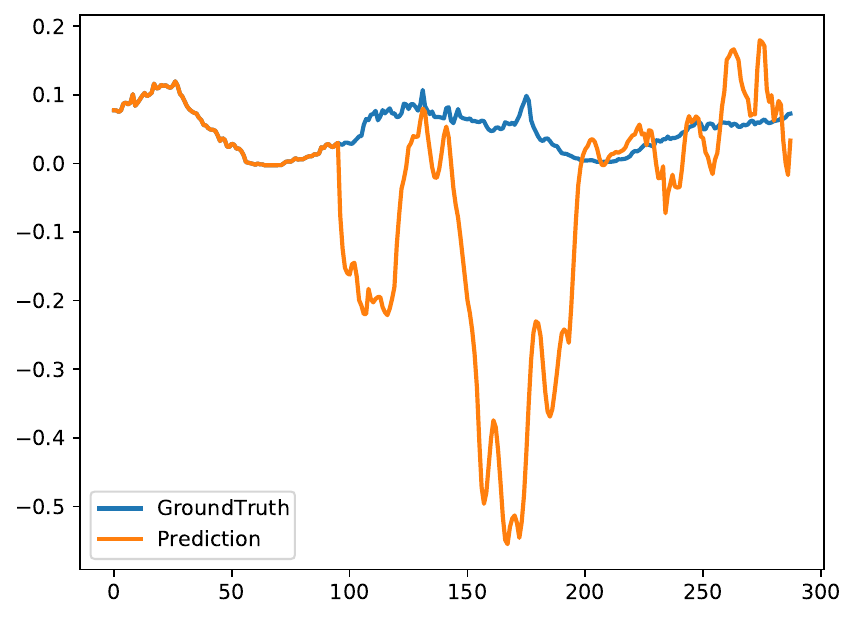}}
\subfigure[\scriptsize{Informer}]{\label{fig:Weather_Informer}\includegraphics[width=23mm]{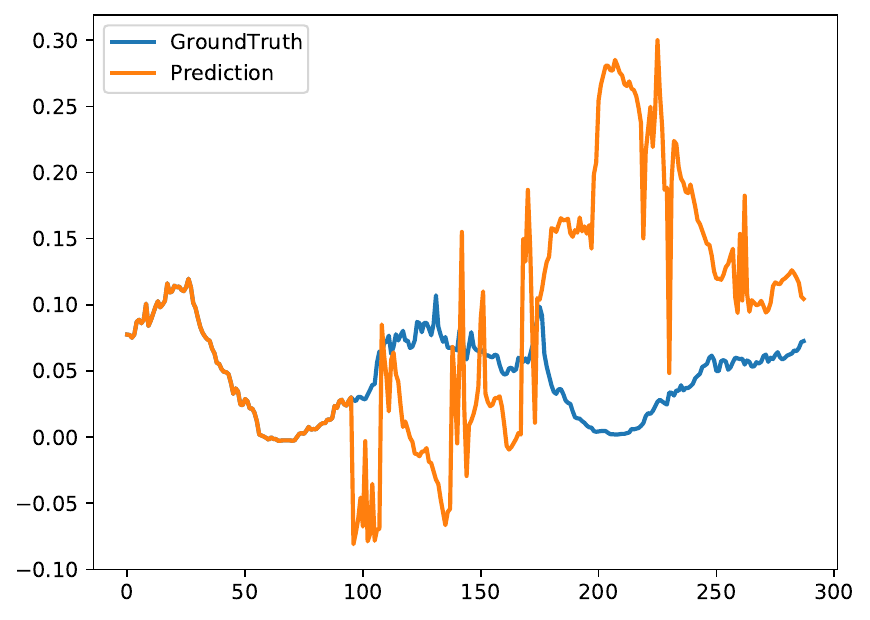}}
\subfigure[\scriptsize{Transformer}]{\label{fig:Weather_Transformer}\includegraphics[width=23mm]{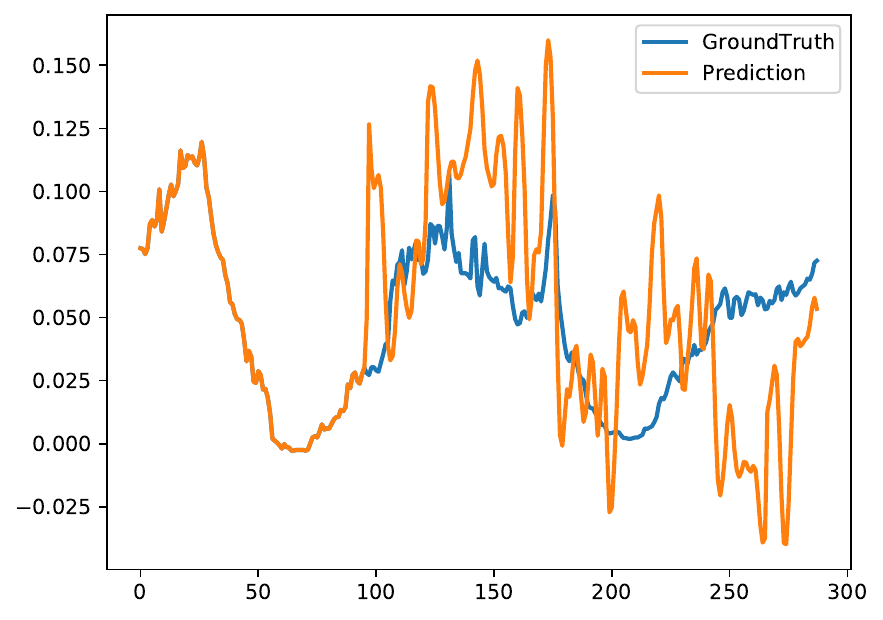}}
\subfigure[\scriptsize{NStransformer + PALS}]{\label{fig:Weather_NStransformer_PALS}\includegraphics[width=23mm]{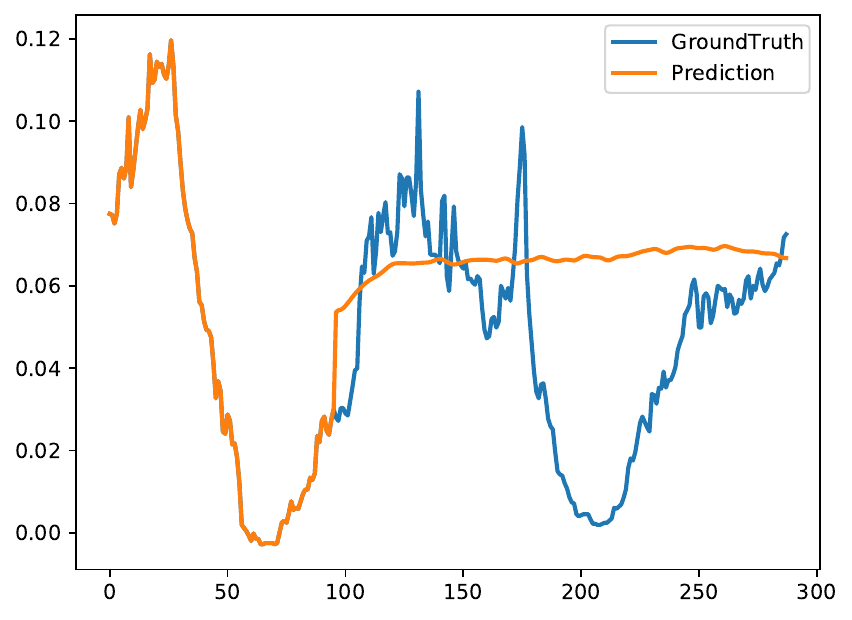}}
\subfigure[\scriptsize{Fedformer + PALS}]{\label{fig:Weather_Fedformer_PALS}\includegraphics[width=23mm]{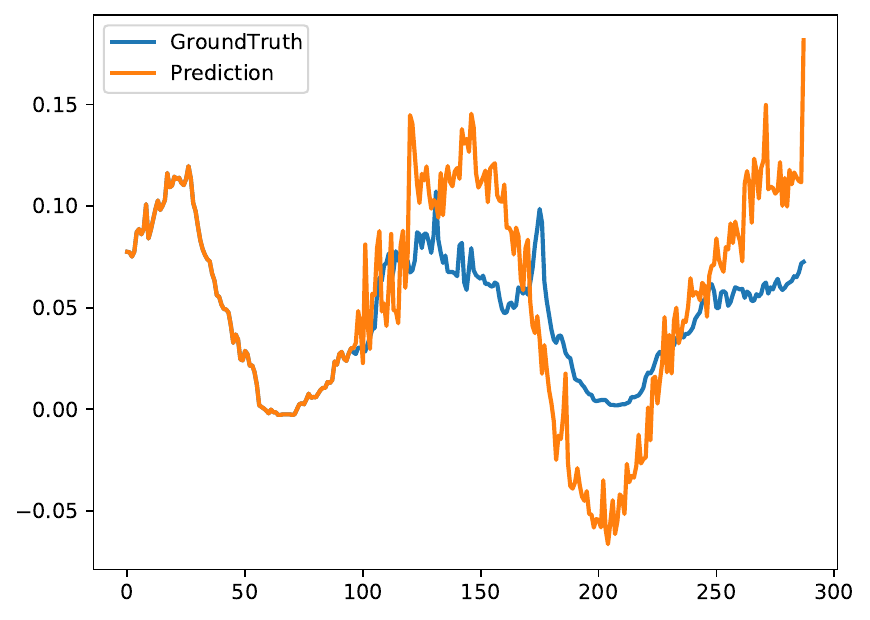}}
\subfigure[\scriptsize{Autoformer + PALS}]{\label{fig:Weather_Autoformer_PALS}\includegraphics[width=23mm]{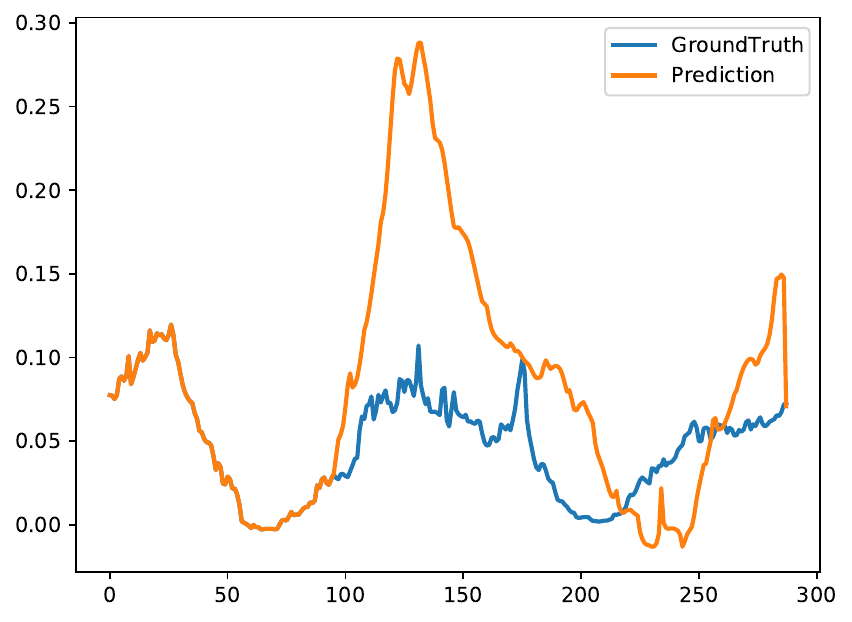}}
\subfigure[\scriptsize{Informer + PALS}]{\label{fig:Weather_Informer_PALS}\includegraphics[width=23mm]{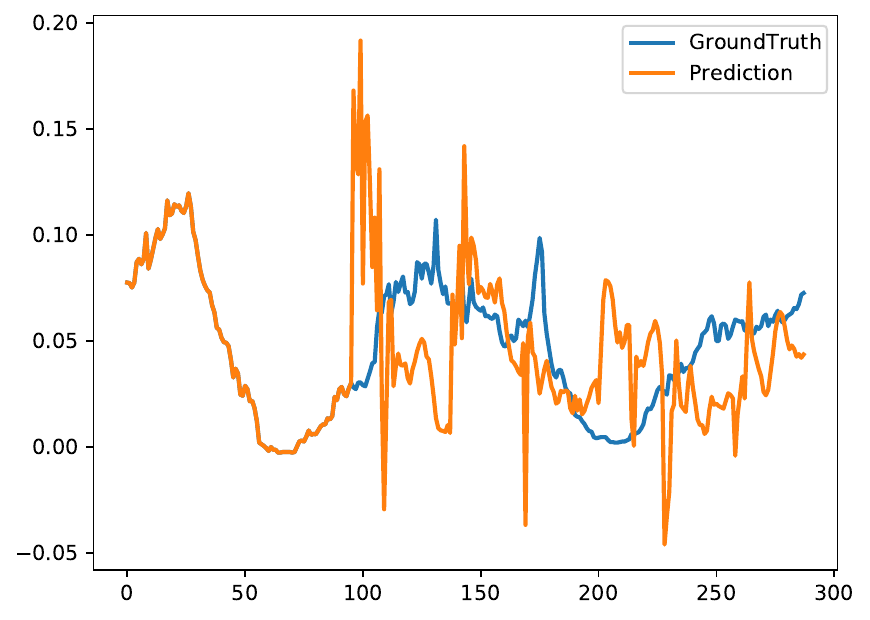}}
\subfigure[\scriptsize{Transformer + PALS}]{\label{fig:Weather_Transformer_PALS}\includegraphics[width=23mm]{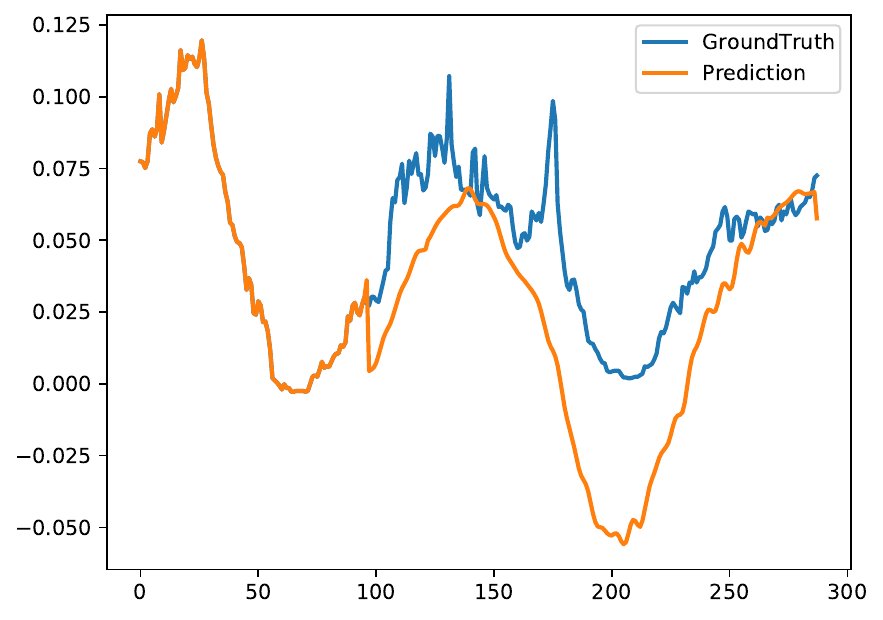}}
\caption{Forecasting Visualization on transformers with and without PALS on the Weather dataset ($H=192$).}
\label{fig:Weather_all_prediction}
\end{figure}

\begin{figure}[!h]
\centering  
\subfigure[\scriptsize{NStransformer}]{\label{fig:Illness_NStransformer}\includegraphics[width=23mm]{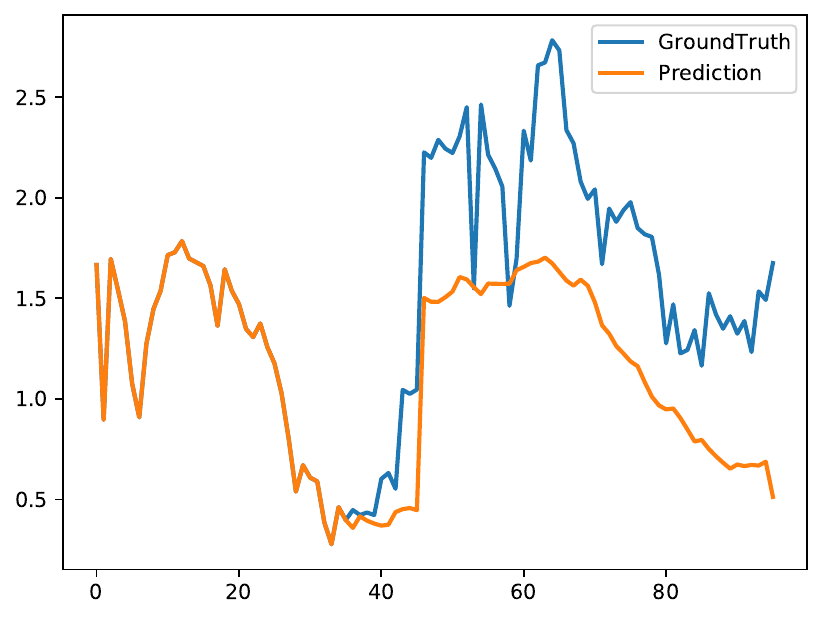}}
\subfigure[\scriptsize{Fedformer}]{\label{fig:Illness_Fedformer}\includegraphics[width=23mm]{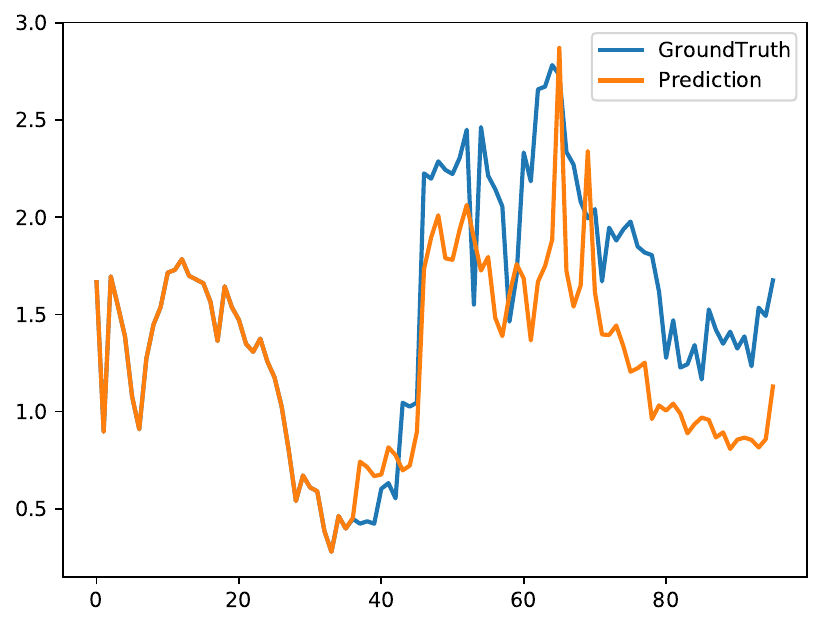}}
\subfigure[\scriptsize{Autoformer}]{\label{fig:Illness_Autoformer}\includegraphics[width=23mm]{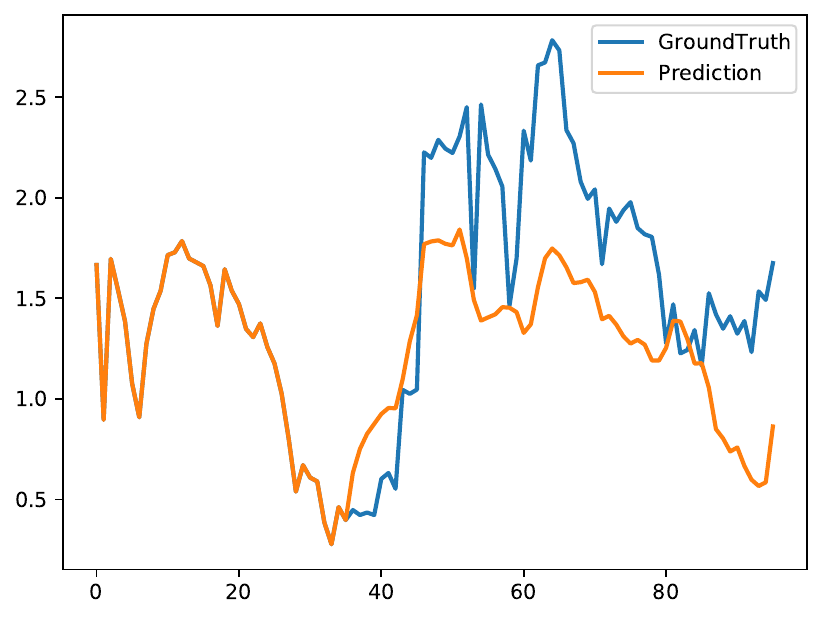}}
\subfigure[\scriptsize{Informer}]{\label{fig:Illness_Informer}\includegraphics[width=23mm]{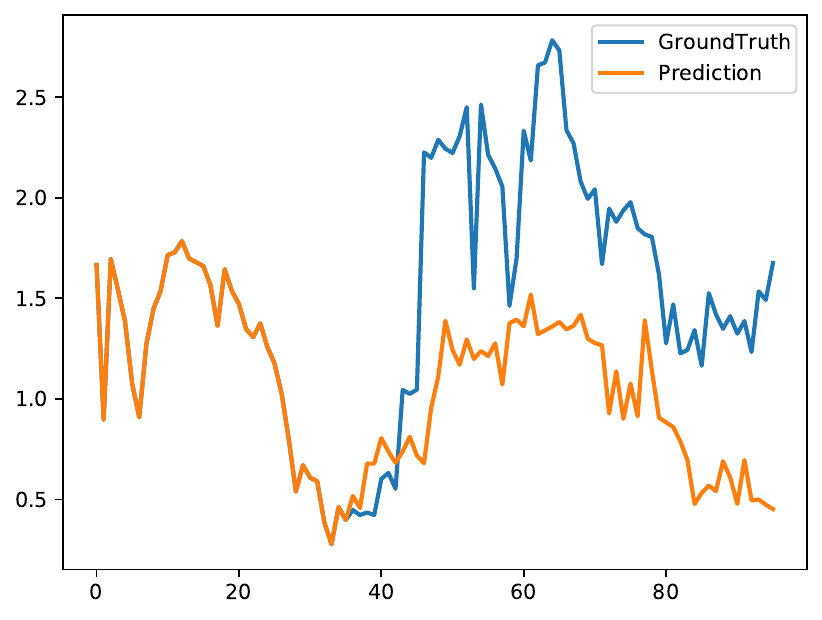}}
\subfigure[\scriptsize{Transformer}]{\label{fig:Illness_Transformer}\includegraphics[width=23mm]{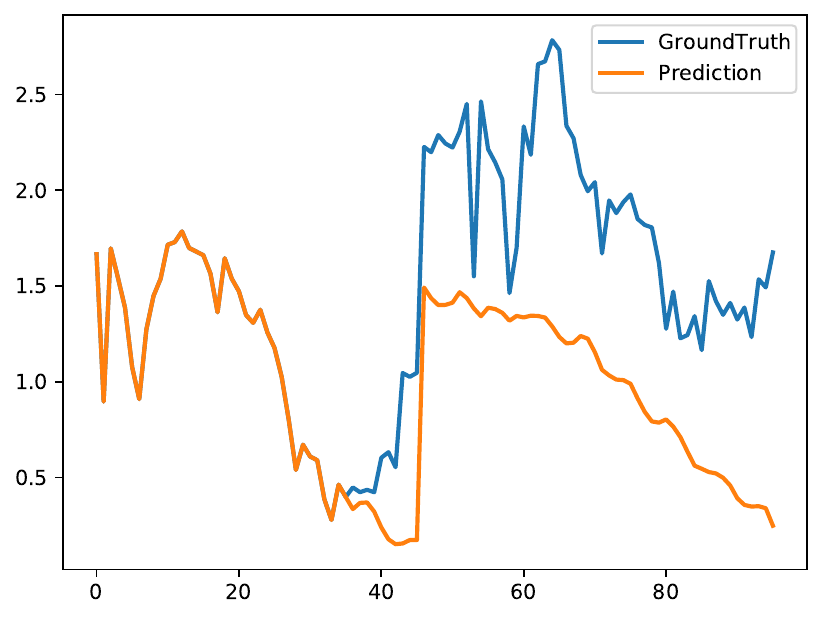}}
\subfigure[\scriptsize{NStransforme + PALS}]{\label{fig:Illness_NStransformer_PALS}\includegraphics[width=23mm]{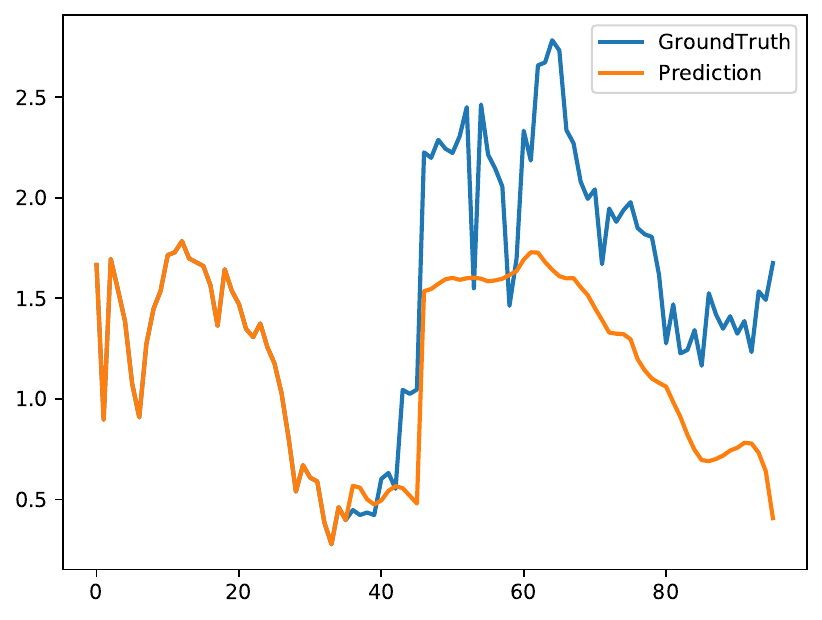}}
\subfigure[\scriptsize{Fedformer + PALS}]{\label{fig:Illness_Fedformer_PALS}\includegraphics[width=23mm]{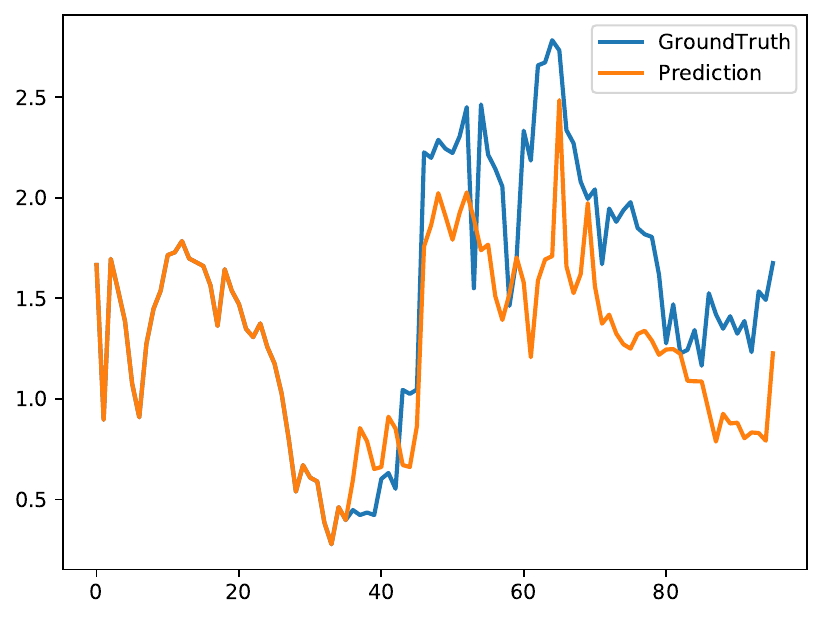}}
\subfigure[\scriptsize{Autoformer + PALS}]{\label{fig:Illness_Autoformer_PALS}\includegraphics[width=23mm]{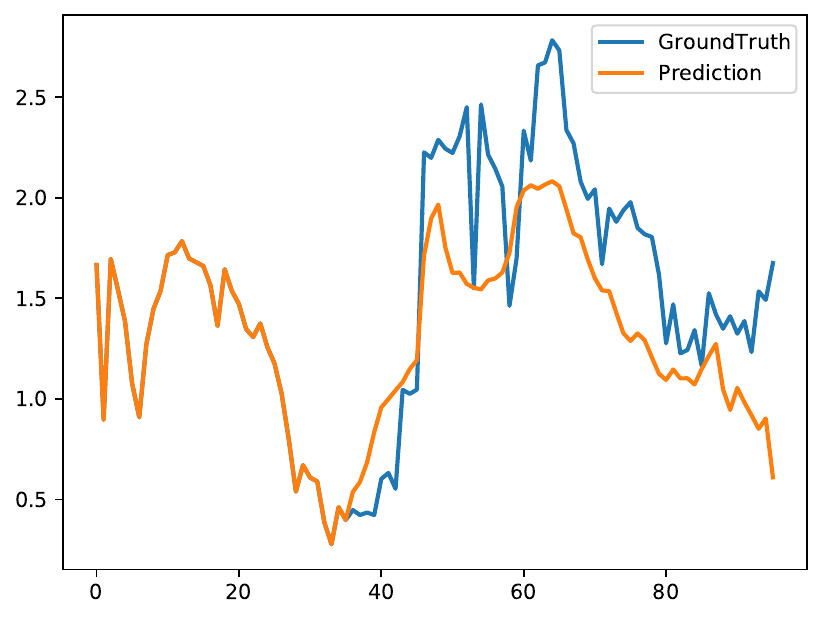}}
\subfigure[\scriptsize{Informer + PALS}]{\label{fig:Illness_Informer_PALS}\includegraphics[width=23mm]{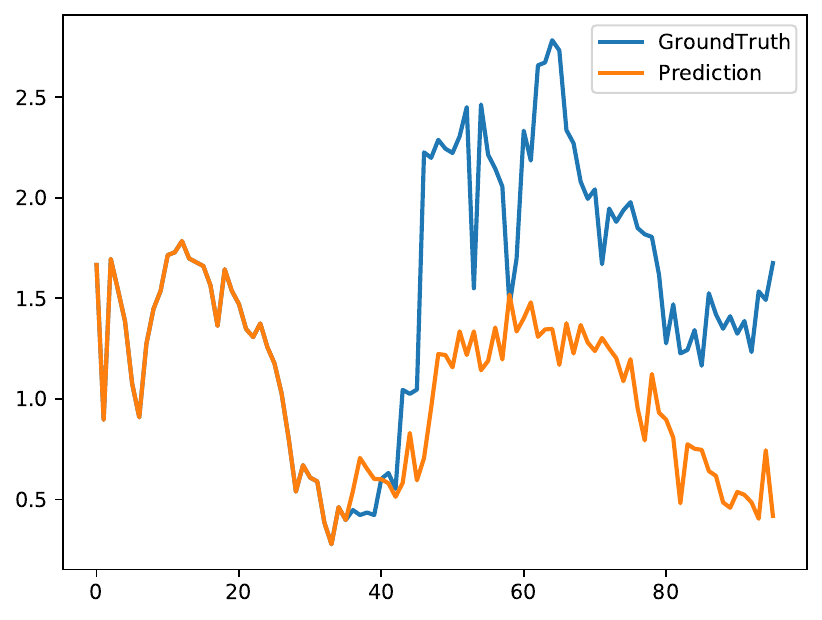}}
\subfigure[\scriptsize{Transformer + PALS}]{\label{fig:Illness_Transformer_PALS}\includegraphics[width=23mm]{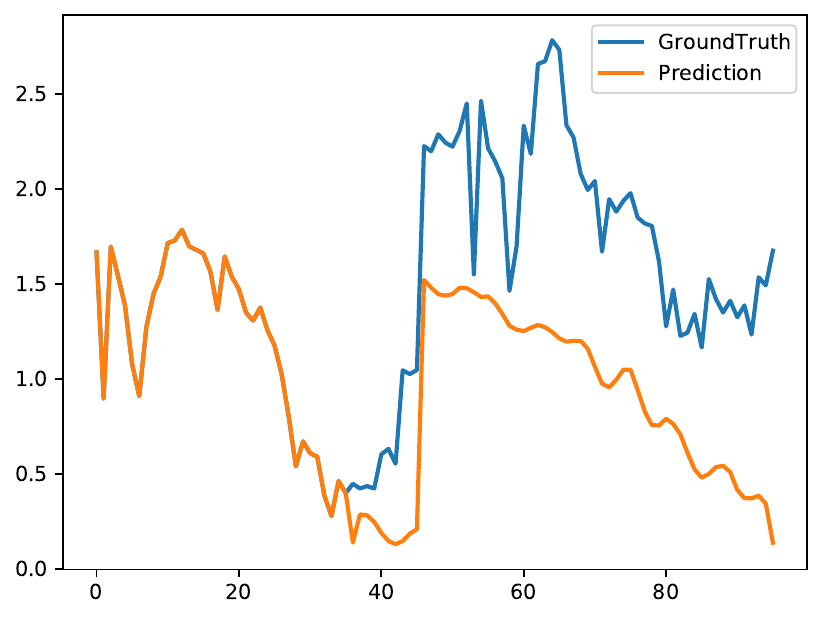}}
\caption{Forecasting Visualization on transformers with and without PALS on the illness dataset ($H=60$).}
\label{fig:Illness_all_prediction}
\end{figure}

\begin{figure}[!h]
\centering  
\subfigure[\scriptsize{NStransformer}]{\label{fig:ETTm2_NStransformer}\includegraphics[width=23mm]{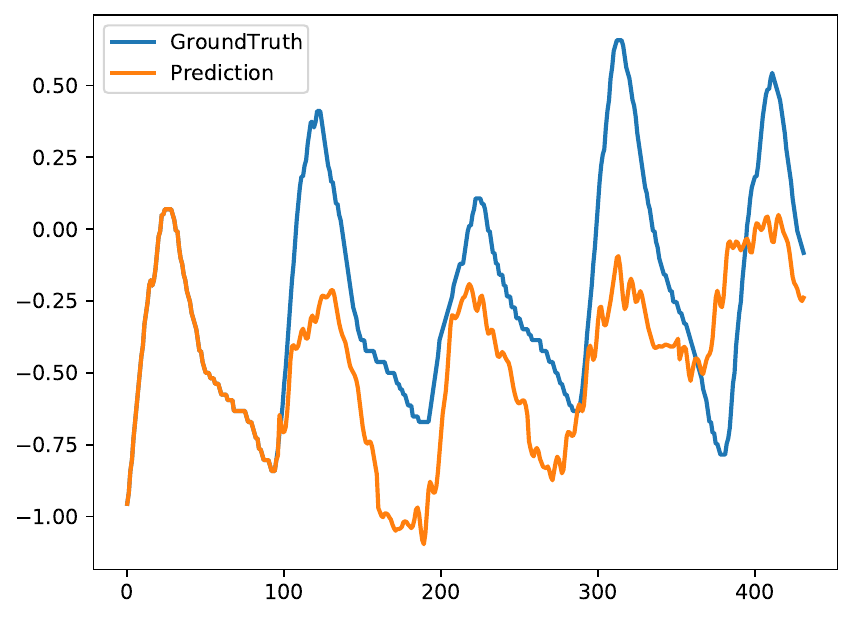}}
\subfigure[\scriptsize{Fedformer}]{\label{fig:ETTm2_Fedformer}\includegraphics[width=23mm]{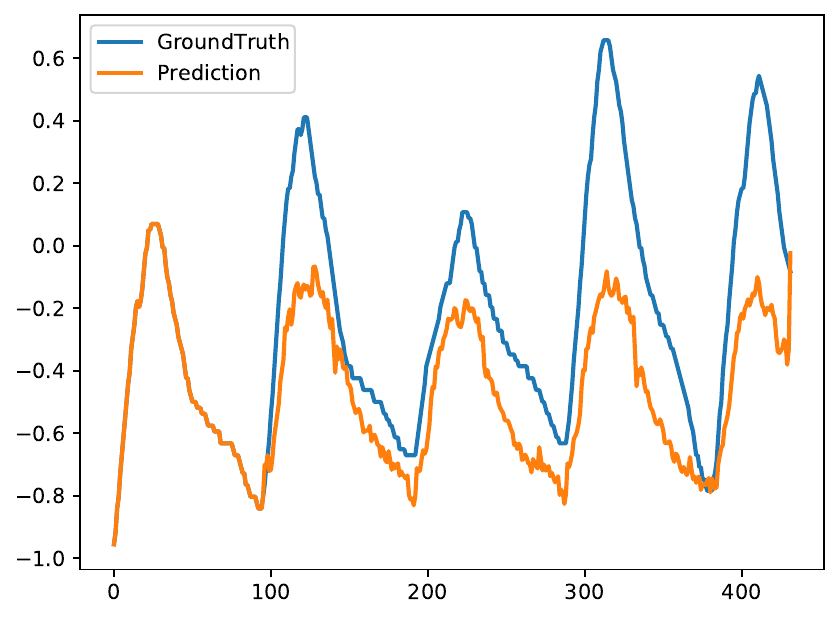}}
\subfigure[\scriptsize{Autoformer}]{\label{fig:ETTm2_Autoformer}\includegraphics[width=23mm]{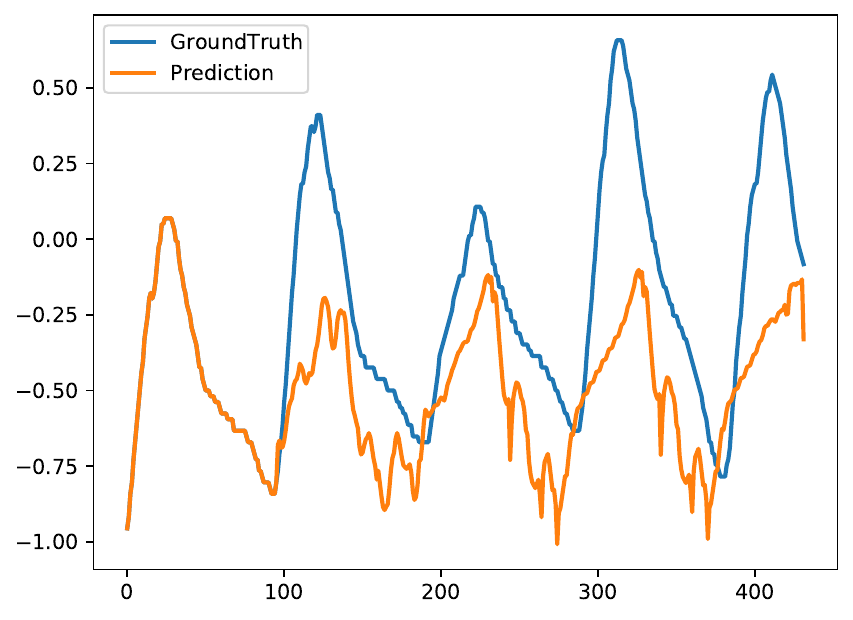}}
\subfigure[\scriptsize{Informer}]{\label{fig:ETTm2_Informer}\includegraphics[width=23mm]{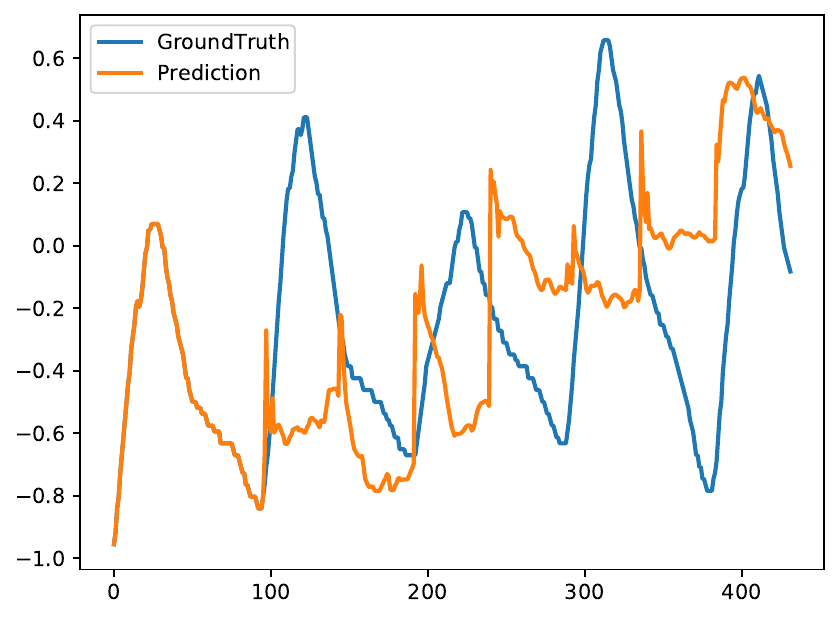}}
\subfigure[\scriptsize{Transformer}]{\label{fig:ETTm2_Transformer}\includegraphics[width=23mm]{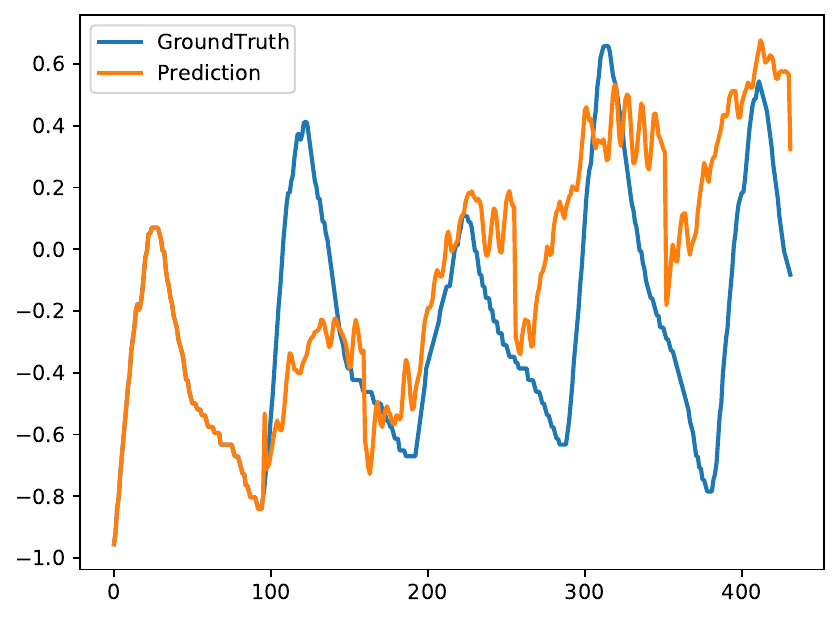}}
\subfigure[\scriptsize{NStransforme + PALS}]{\label{fig:ETTm2_NStransformer_PALS}\includegraphics[width=23mm]{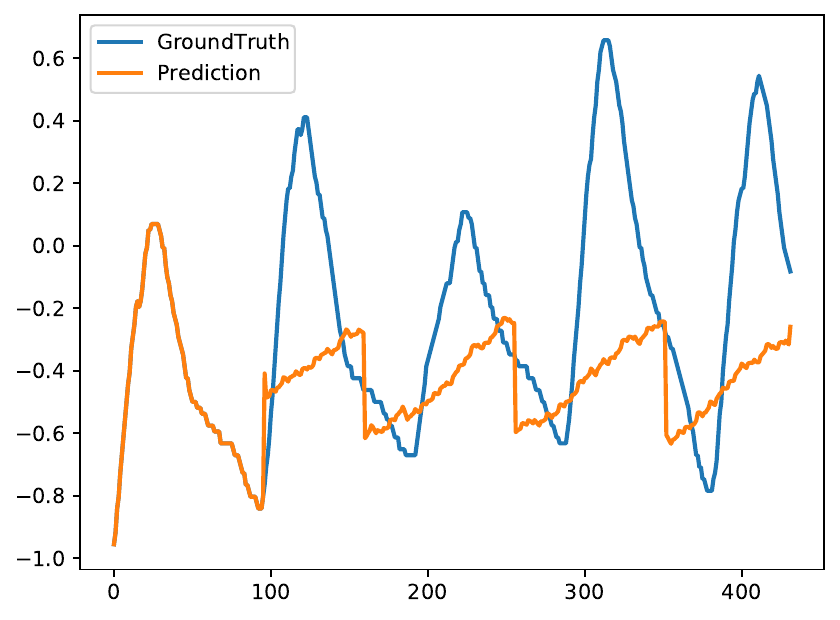}}
\subfigure[\scriptsize{Fedformer + PALS}]{\label{fig:ETTm2_Fedformer_PALS}\includegraphics[width=23mm]{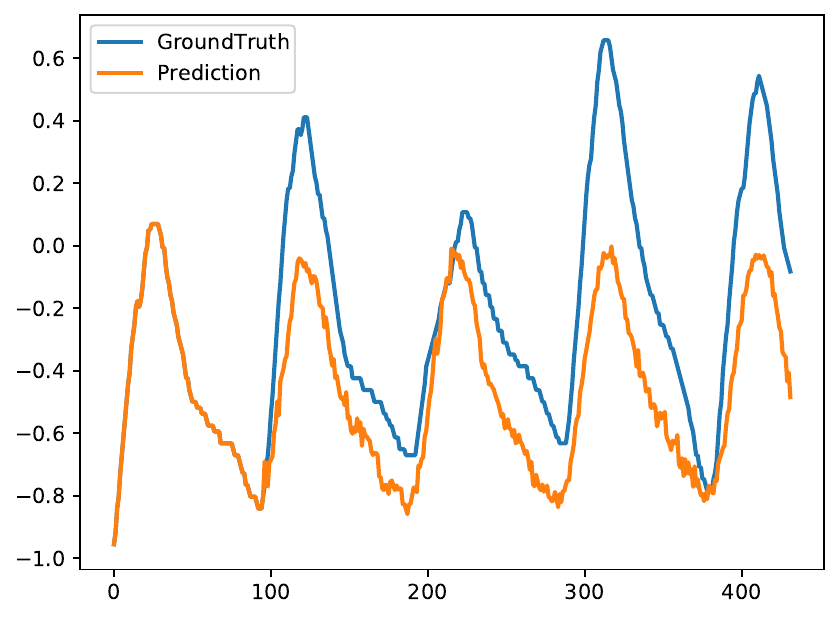}}
\subfigure[\scriptsize{Autoformer + PALS}]{\label{fig:ETTm2_Autoformer_PALS}\includegraphics[width=23mm]{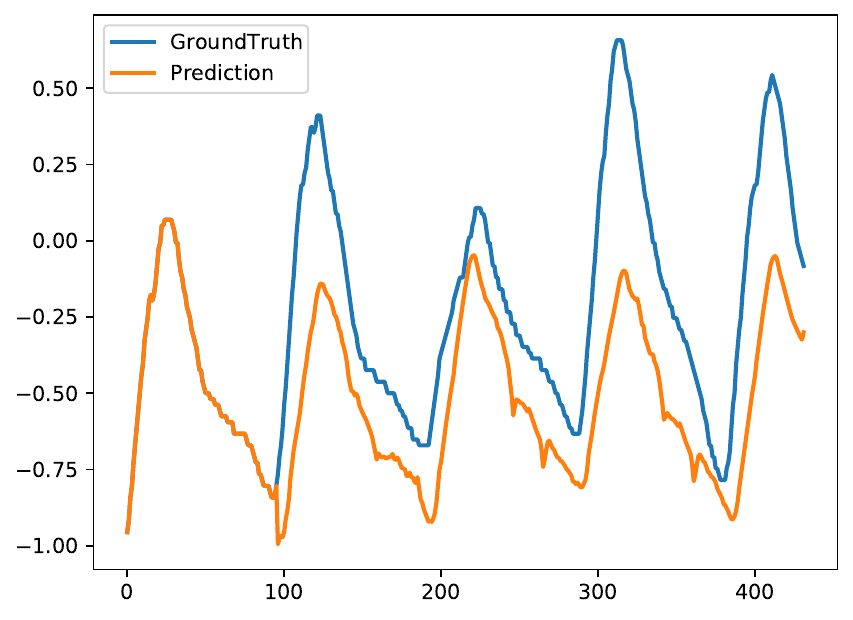}}
\subfigure[\scriptsize{Informer + PALS}]{\label{fig:ETTm2_Informer_PALS}\includegraphics[width=23mm]{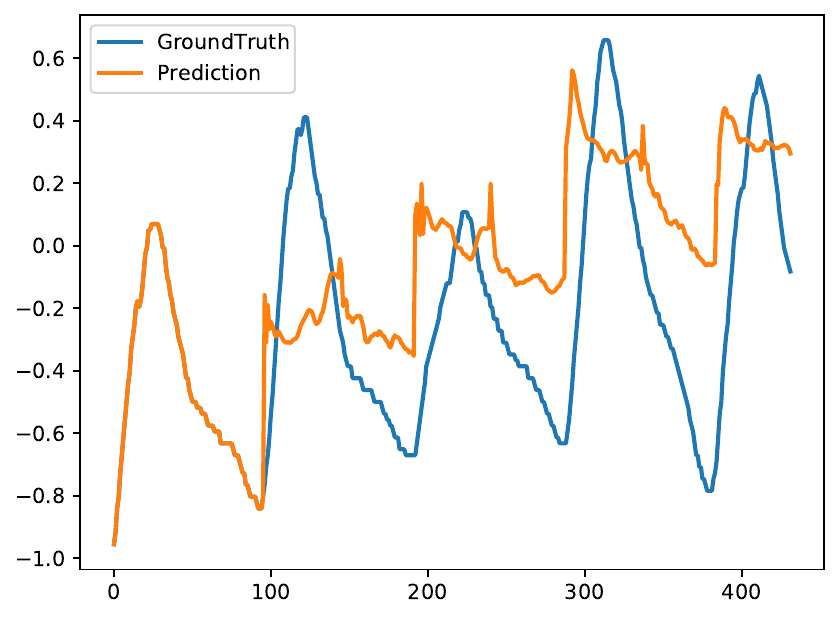}}
\subfigure[\scriptsize{Transformer + PALS}]{\label{fig:ETTm2_Transformer_PALS}\includegraphics[width=23mm]{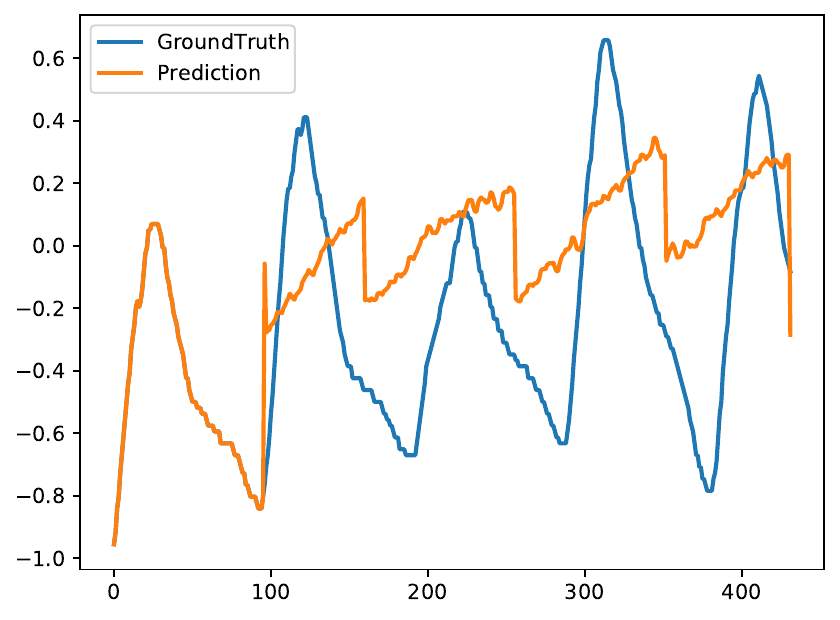}}
\caption{Forecasting Visualization on transformers with and without PALS on the ETTm2 dataset ($H=336$).}
\label{fig:ETTm2_all_prediction}
\end{figure}

\begin{figure}[!h]
\centering  
\subfigure[\scriptsize{NStransformer}]{\label{fig:Exchange_NStransformer}\includegraphics[width=23mm]{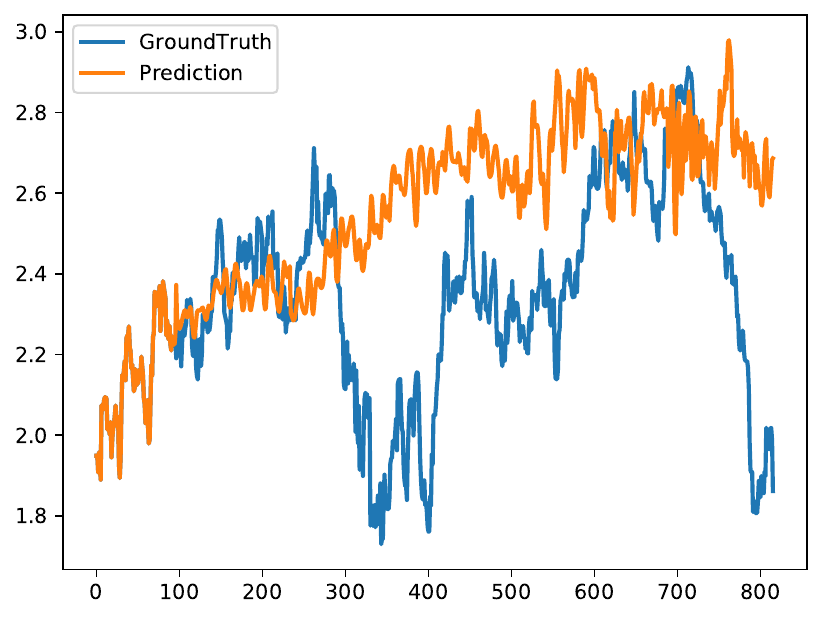}}
\subfigure[\scriptsize{Fedformer}]{\label{fig:Exchange_Fedformer}\includegraphics[width=23mm]{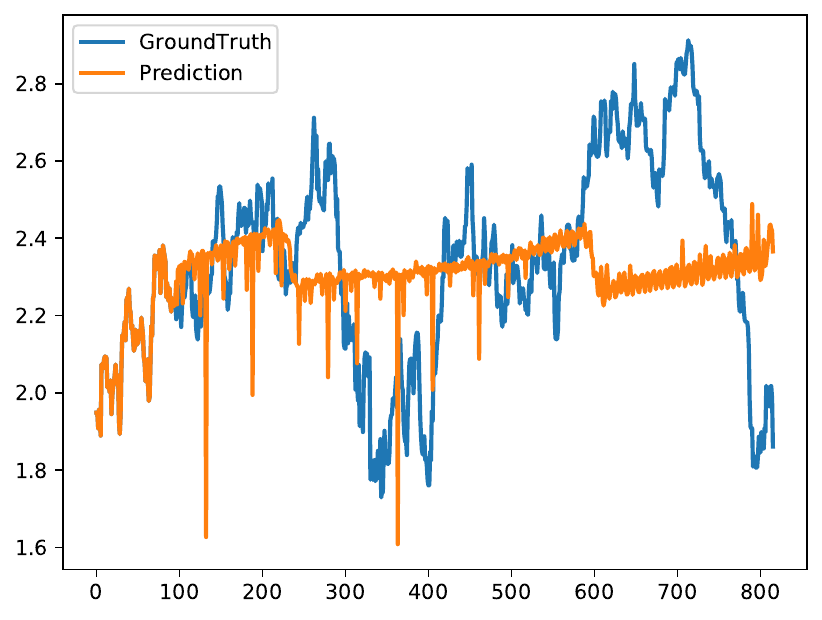}}
\subfigure[\scriptsize{Autoformer}]{\label{fig:Exchange_Autoformer}\includegraphics[width=23mm]{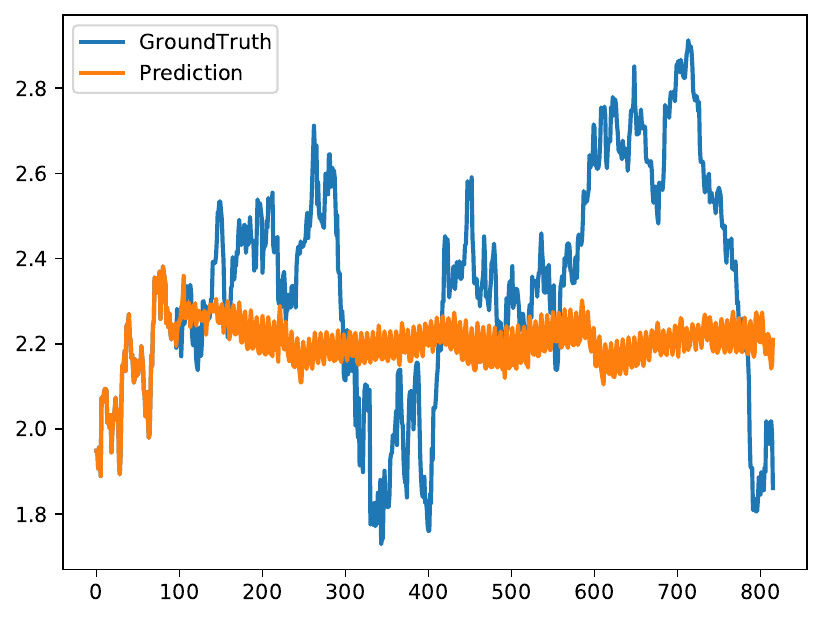}}
\subfigure[\scriptsize{Informer}]{\label{fig:Exchange_Informer}\includegraphics[width=23mm]{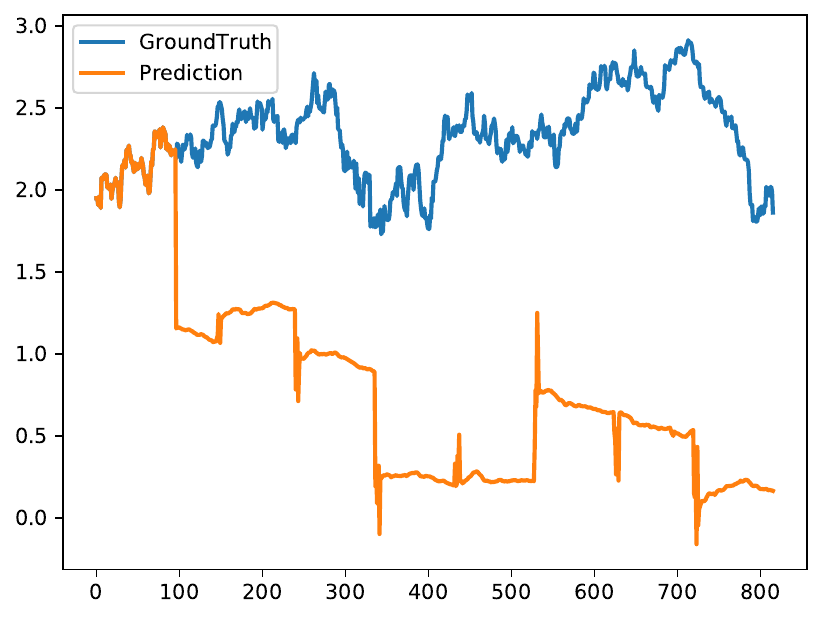}}
\subfigure[\scriptsize{Transformer}]{\label{fig:Exchange_Transformer}\includegraphics[width=23mm]{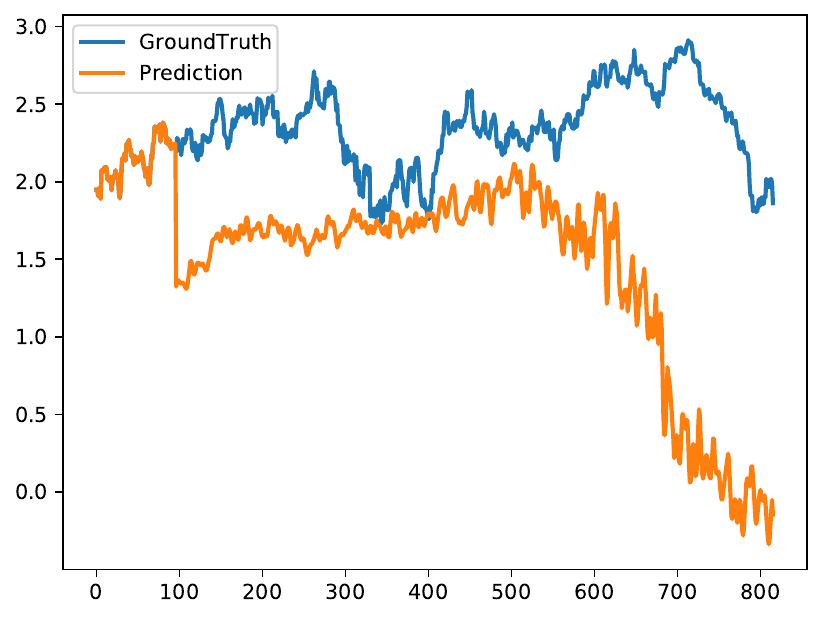}}
\subfigure[\scriptsize{NStransforme + PALS}]{\label{fig:Exchange_NStransformer_PALS}\includegraphics[width=23mm]{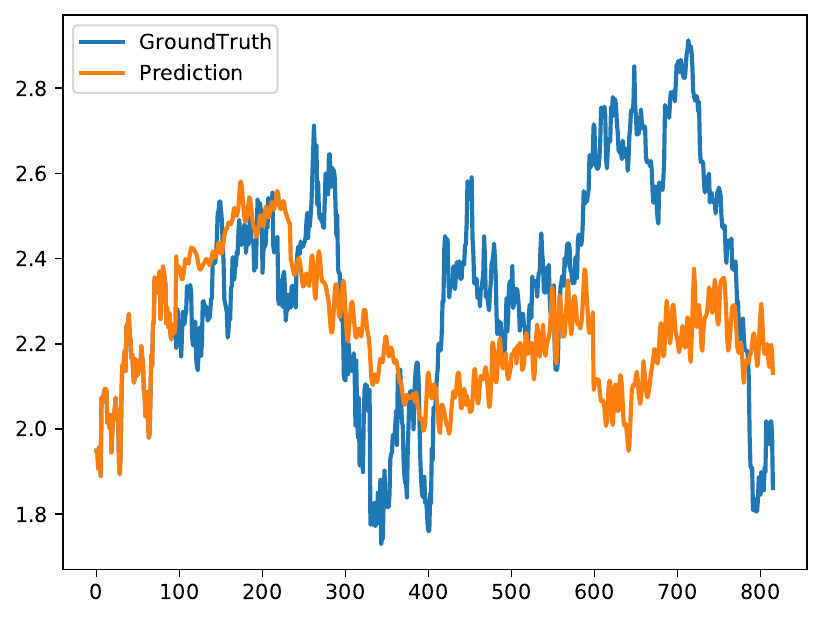}}
\subfigure[\scriptsize{Fedformer + PALS}]{\label{fig:Exchange_Fedformer_PALS}\includegraphics[width=23mm]{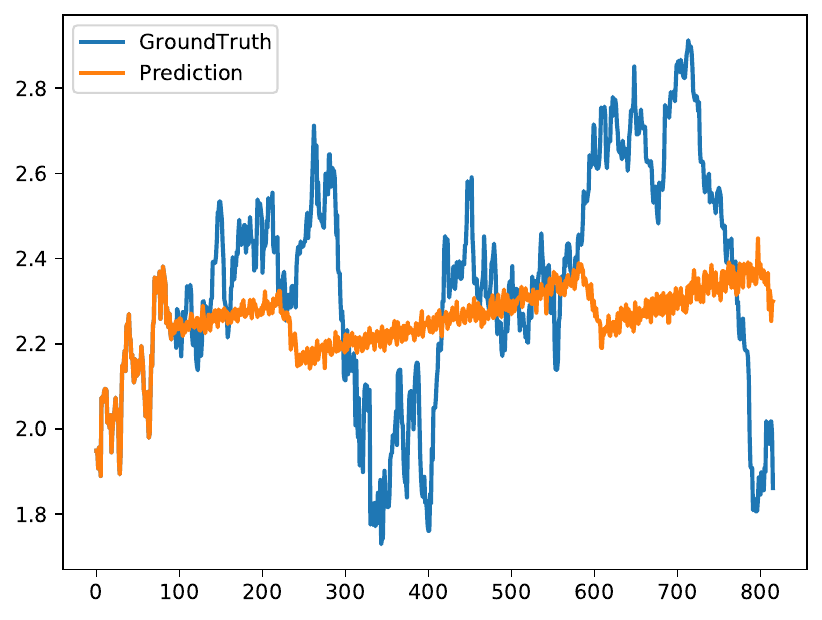}}
\subfigure[\scriptsize{Autoformer + PALS}]{\label{fig:Exchange_Autoformer_PALS}\includegraphics[width=23mm]{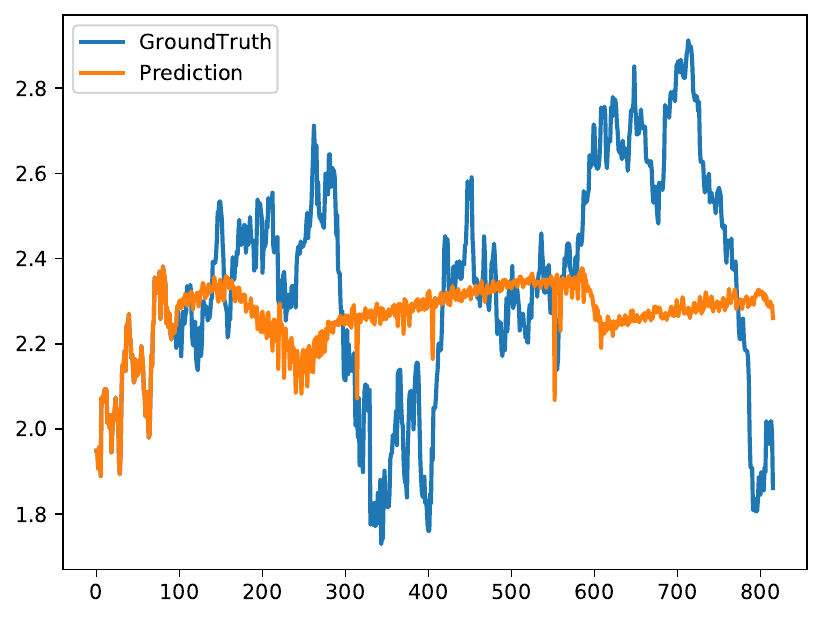}}
\subfigure[\scriptsize{Informer + PALS}]{\label{fig:Exchange_Informer_PALS}\includegraphics[width=23mm]{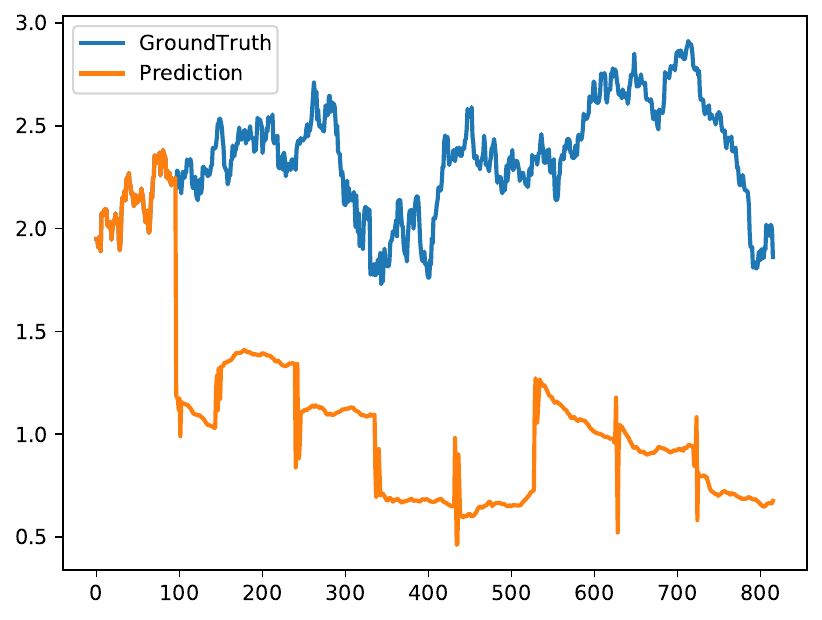}}
\subfigure[\scriptsize{Transformer + PALS}]{\label{fig:Exchange_Transformer_PALS}\includegraphics[width=23mm]{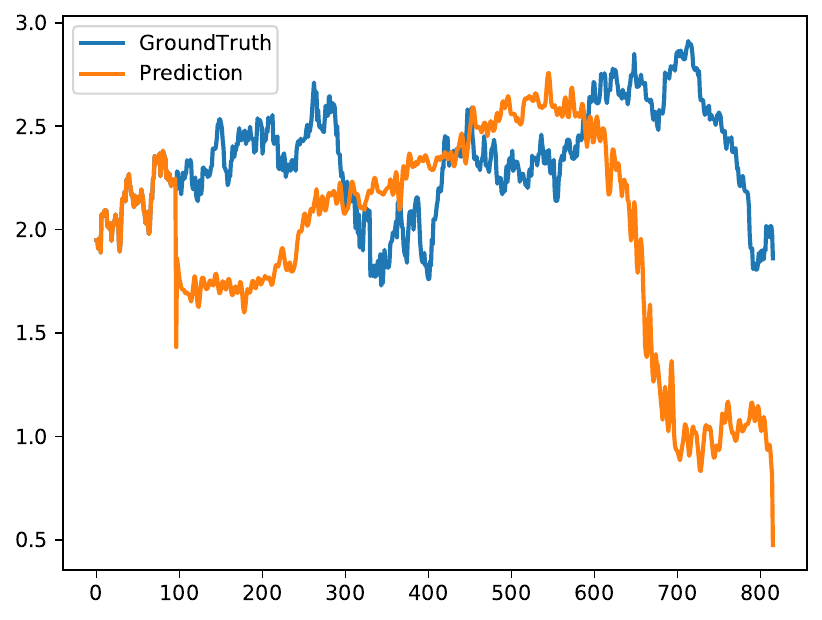}}
\caption{Forecasting Visualization on transformers with and without PALS on the Exchange dataset ($H=720$).}
\label{fig:Exchange_all_prediction}
\end{figure}